\documentclass[11pt, reqno]{amsart}

\usepackage{amsmath}
\usepackage{amsthm}
\usepackage{comment}
\usepackage{mathtools}
\usepackage{xfrac}
\newtheorem{theorem}{Theorem}

\usepackage[round,authoryear]{natbib}
\usepackage{bm}
\usepackage{amssymb}
\usepackage{bbding}
\usepackage{pifont}
\usepackage{rotating}
\usepackage{caption}
\usepackage{cuted}
\usepackage{subcaption} 

\usepackage[utf8]{inputenc} 
\usepackage[T1]{fontenc}    
\usepackage{hyperref}       
\usepackage{url}            
\usepackage{booktabs}       
\usepackage{amsfonts}       
\usepackage{nicefrac}       
\usepackage{microtype}      
\usepackage[table]{xcolor}
\usepackage{algorithm}
\usepackage{algpseudocode}
\usepackage{mathtools}
\usepackage{xfrac}
\usepackage{adjustbox}
\newtheorem{definition}{Definition}

\algdef{SE}[REPEATN]{RepeatN}{End}[1]{\algorithmicrepeat\ #1 \textbf{times (or until stabilization of $\bm{\alpha}$)}}{\algorithmicend}
\newcommand{\bcdot}{\boldsymbol{\cdot}}

\definecolor{darkred}{rgb}{0.5,0,0}
\newcommand{\TODO}[1]{\textcolor{red}{#1}}

\newcommand{\plusterms}[1]{\makebox[10pt][l]{\scalebox{0.65}{\textbf{\textcolor{darkred}{+#1}}}}}
\newcommand*{\thead}[1]{\multicolumn{1}{|c|}{\bfseries #1}}

\title[Unrolled-SINDy]{Unrolled-SINDy: A Stable Explicit Method for Non linear PDE Discovery from Sparsely Sampled Data}

\author[F. ALI BANNA]{Fayad ALI BANNA$^1$}
\author[A. CARADOT]{Antoine CARADOT$^1$}
\author[E. Brandao]{Eduardo BRANDAO$^1$}
\author[J.P. COLOMBIER]{Jean-Philippe COLOMBIER$^1$}
\author[R. EMONET]{R\'emi Emonet$^1$}
\author[M. SEBBAN]{Marc Sebban$^1$}

\address{$^1$Jean Monnet University Saint-\'Etienne, CNRS, Institut d'Optique Graduate School, Inria, Hubert Curien Laboratory UMR 5516, F-42023, SAINT-\'ETIENNE, France}

\begin{document}

%

%

\begin{abstract}
Identifying from observation data the governing differential equations of a physical dynamics is a key challenge in machine learning. Although approaches based on SINDy have shown great promise in this area,  they still fail to address a whole class of real world problems where the data is sparsely sampled in time. In this article, we introduce Unrolled-SINDy, a simple methodology that leverages an unrolling scheme to improve the stability of explicit methods for PDE discovery. By decorrelating the numerical time step size from the sampling rate of the available data, our approach enables the recovery of equation parameters that would not be the minimizers of the original SINDy optimization problem due to large local truncation errors. Our method can be exploited either through an iterative closed-form approach or by a gradient descent scheme. Experiments show the versatility of our method. On both traditional SINDy and state-of-the-art noise-robust iNeuralSINDy, with different numerical schemes (Euler, RK4), our proposed unrolling scheme allows to tackle problems not accessible to non-unrolled methods.
\end{abstract}

\maketitle      

\section{Introduction}
Embedding physical knowledge into learning algorithms is at the core of Physics-informed Machine Learning (PIML)~\citep{osti_1852843}, a new line of research that has recently attracted much attention from both physics and ML communities. Beyond addressing issues related to ill-posed problems, data scarcity and solution consistency, PIML can be applied for (i) solving Differential Equations (ODEs or PDEs) ({\it e.g.} PINNs \citep{raissi2019physics}, FNOs \citep{LiKALBSA21}), (ii) leveraging physical priors to accelerate the learning process in hybrid (knowledge+data) modeling ({\it e.g.} PINO \citep{li2023physicsinformed})  or augment  incomplete physical knowledge ({\it e.g.} APHYNITY \citep{Yin_2021} or hybrid PINNs \citep{Doumeche.BERN.2025}) when physics is only partially understood or where it is derived under ideal conditions that do not hold exactly in real applications, and (iii) learning governing  differential equations  directly from data measurements (thus solving an inverse problem) in domains where the theory remains elusive. In this paper, we focus on this latter scenario which has been shown to be of great help for discovering knowledge in various domains, including engineering, climate science, finance, medicine, biology and chemistry.

One of the most important breakthroughs in discovering equations from data came with the SINDy algorithm (see \citep{Brunton2016} for ODEs and \citep{PDE-FIND} for its extension to PDEs with PDE-FIND) which envisions the problem from the perspective of sparse regression performed from a library of functions (typically including partial derivatives, trigonometric or polynomial terms). SINDy leverages the realistic assumption that in most equations, only a few important terms govern the underlying dynamics, prompting us to promote sparsity. The pioneered version of SINDy relies on an  {\it explicit numerical method}, meaning that it calculates the state of the system at $t+h_t$ using 
known values from the current time step $t$.
 Several extensions have been introduced since then, including SINDYc~\citep{brunton2016SINDYc}  to leverage external inputs and feedback control, Reactive SINDy~\citep{reactiveSINDy}  to deal with vector-valued ansatz functions, Ensemble-SINDy~\citep{urban2022a} enabling uncertainty quantification, SINDy-PI \citep{Kaheman_2020} which uses rational functions to discover equations, an extension of SINDy to stochastic dynamical systems~\citep{Boninsegna_2018} or WSINDy \citep{Messenger_2021} which 
eliminates pointwise derivative approximations with a weak formulation providing better robustness to noise.

Despite remarkable performances, SINDy-like explicit methods face a major limitation when deployed on real applications: they rely on accurate time derivative approximations along the identification process, typically using numerical methods like Euler, imposing constraints on the sampling time step sizes $h_t$ of the data. As illustrated in Fig.~\ref{fig:unrolling} (b), this can constitute a serious obstacle in scenarios where data is scarce, leading to large local truncation errors.   To overcome this limitation, a significant improvement has been proposed with RK4-SINDy \citep{RK4-SINDy}, which leverages the {\it explicit} fourth-order Runge-Kutta method and its advantageous convergence properties to better recover the underlying equations. However, even though RK4 pushes the limits of Euler to some extent, its small {\it absolute stability region} still limits RK4-SINDy, particularly for identifying stiff equations, unless very small step sizes are used. 
From an optimization perspective, as for both numerical methods, the local truncation error worsens with the growth of $h_t$, \textbf{the  parameters of the equation underlying the sparse data no longer represent a minimizer} of the considered optimization problem.

To address this major issue, we introduce a new PDE discovery method, namely Unrolled-SINDy, which {\bf decorrelates the numerical time step size from the available data sampling rate}. It consists in unrolling $K$-times each numerical step, as illustrated in Fig.~\ref{fig:unrolling} (center), {\bf without needing additional  data}. To do so, Unrolled-SINDy evaluates the library terms at $K$ iterative estimates of the solution between two successive observations $u(t)$ and $u(t+h_t)$. We show that this way, Unrolled-SINDy leverages an intrinsic smaller time step and thus benefits of a lower local truncation error. If at first sight, our unrolled scheme may seem close to the principle of Runge–Kutta methods, it is important to note that the latter are difficult to generalize to many stages associated with a large number of Butcher order conditions, preventing non-unique and closed-form solutions for the coefficients, and  leading sometimes to numerical instability.

Note that we position this original contribution within a framework where parsimonious models are promoted, and the identification of the governing equations  can be solved in closed-form ({\it i.e.}, no parameter tuning aside a sparsity threshold) and relies on an explicit  numerical technique to approximate the time derivatives. This concerns the wide range of SINDy-like methods that have flourished in the literature during the past few years and that belong to the state of the art in ODE/PDE discovery. However, notice that our proposed unrolling strategy can also be applied to neural network-based PDE discovery methods, including PDE-Net~\citep{long2018pdenetlearningpdesdata}, DeepMoD~\citep{Both_2021}, PDE-LEARN~\citep{STEPHANY2024106242}, PDE-READ~\citep{PDE-READ}, PINN~\citep{raissi2019physics},  ICNET~\citep{ICNET} or iNeural-SINDy~\citep{ForootaniGB25}. Broadly speaking, any neural method that implements a discretization scheme can directly benefit from our unrolling approach.
We illustrate this nice feature by applying the unrolling inside the state-of-the-art noise-robust iNeural-SINDy. 

To recap, the contribution of this paper is four-fold:
\\
{\bf (A)} We propose an unrolling scheme for SINDy-like methods, which decorrelates the integration time step from the data inter-observation time step $h$.
{\bf (B)} Based on this methodology, we propose Unrolled-SINDy, a new explicit ODE/PDE discovery method.
Unrolled-SINDy is simple to implement, relatively fast and comes with a closed-form. It is versatile as it can be adapted to any explicit method based on a finite-difference technique, such as forward Euler, central difference or Runge-Kutta.
{\bf (C)} Given a time step size $h$, we show that $K$ Unrolled-SINDy benefits of a local truncation error in the order of ${\mathcal O}\left ((\frac{h}{K})^p \right )$ 
that holds for any RK method of order $p$.
{\bf (D)} We carry out a comprehensive experimental study on different equations and show that, with both Euler and RK4, our Unrolled-SINDy is more robust at recovering the underlying physics than SINDy and is able to solve problems with large time steps that are inaccessible to current methods.
We also demonstrate that unrolling benefits neural PDE discovery methods: on state-of-the-art noise-robust iNeural-SINDy, unrolling allows to tackle problems with scarcer observations. 

The rest of this paper is organized as follows. In Section~\ref{sec:def}, we introduce the necessary definitions and notations and review the principles of SINDy and RK4-SINDy. Section~\ref{sec:Unrolled-SINDy} presents the Unrolled-SINDy framework.  Section~\ref{sec:expes} 
is devoted to experiments with both Unrolled-SINDy and our unrolling scheme applied to iNeural-SINDy.

\section{Definitions, Notations and Necessary Background} \label{sec:def}
We consider equations of the general form: $\frac{\partial u}{\partial t}={\mathcal N}[u]$ where ${\mathcal N}[\cdot]$ is a (possibly nonlinear) differential operator involving partial space derivatives and $u(t,\mathbf{x}) \in \mathbb{R}^{d_2}$ is the latent, supposedly unique, hidden solution, with $t \in [0,T]$ the temporal variable and $\mathbf{x} \in \Omega \subset \mathbb{R}^d$ the spatial coordinates.
Note that when a higher-order time derivative is involved in the left hand part of the equation, one can rewrite the latter in the form of a system of first-order equations involving  new variables corresponding to $u$ and its respective temporal derivatives. 
Let us consider a library $\Theta$ (of size $|\Theta|$)
of functions typically composed of constants, exponentials,  trigonometric terms as well as partial derivatives and polynomials up to an order $r$. Let  $\Theta_u: [0,T] \times \mathbb{R}^d \to \mathbb{R}^{|\Theta|}$ be the function mapping  $ (t,\mathbf{x}) \mapsto  \left ( \frac{\partial u_1}{\partial \mathbf{x}_1}, \frac{\partial u_1}{\partial \mathbf{x}_2}, \frac{\partial^2 u_1}{\partial \mathbf{x}_1 \partial \mathbf{x}_2}, ...,\frac{\partial^2 u_{1}}{\partial \mathbf{x}^2_d},...,\frac{\partial^r u_{d_2}}{\partial \mathbf{x}^r_d},...  \right )$ which evaluates the library at ($t,\mathbf{x}$). 

For each task, a set $S$ of data is built as follows: we consider $M$ spatial locations $\mathbf{x}_m \in \Omega, m=1,\dots,M$ and collect (by simulation or observation) the corresponding solutions  $u(t_j,\mathbf{x}_m)$ at the time instances $\{t_0,...,t_J \}$, where $h_j=t_{j+1}-t_j$ is the time-step, with $t_0=0$. The resulting set $S=\{u(t_j,\mathbf{x}_m) \}_{j=0..J, m=1..M}$ is composed of $(J+1)\times M$ observations allowing us to build $N=J\times M$ training pairs $(u(t_j,\mathbf{x}_m),u(t_{j+1},\mathbf{x}_m))$ that will be used for learning the underlying dynamics. Let $\mathbf{U}_{prev}$ (resp. $\mathbf{U}_{next}$) denote the $N \times d_2$ matrix composed of the first (resp. second) elements of the $N$ pairs. Moreover, let us define the  $N \times |\Theta|$ matrix $\mathbf{\Theta_u}=(\Theta_u(t_j,\mathbf{x}_m))_{j=0..J-1,m=1..M}$ and the $N \times N$ matrix $\mathbf{H}$ containing $N \times M$ copies of the time-steps $\mathbf{h} = (h_0,..,h_{J-1})$. All these notations  are illustrated in Fig.~\ref{fig:unrolling} (e). Finally, we make use in this paper of the notation  ${\mathcal \bm{H}}$ as a $N \times d_2$ matrix composed of $M\times d_2$ repetitions of $\mathbf{h}$.


\paragraph{Euler-SINDy \citep{PDE-FIND}:}
  SINDy, that we will call Euler-SINDy when run with the  Euler method, aims to solve the following problem:
\begin{align}
\min_{\bm{\alpha}} \ 
 \sum_{j=0}^{J-1} \sum_{m=1}^M 
\Big\| u(t_{j+1},\mathbf{x}_m)  - \Big( 
   \underbrace{u(t_j,\mathbf{x}_m) 
   + h_j \cdot \Theta_u(t_j,\mathbf{x}_m)\bm{\alpha}}_{
      \mathclap{\text{Euler estimate}}
   }
   \Big) 
\Big\|_2^2 
 + \lambda \|\bm{\alpha}\|_{\mathcal F}^2 
\label{eq:sindySTR}
\end{align}
or
\begin{align}
\min_{\bm{\alpha}} \ 
& \left\| \mathbf{U}_{next} - 
\Big( 
   \underbrace{\mathbf{U}_{prev} 
   + \mathbf{H} \cdot \mathbf{\Theta}_u \cdot \bm{\alpha}}_{
      \mathclap{\text{Euler estimate}}
   }
\Big) \right\|_{\mathcal F}^2+\lambda \|\bm{\alpha}\|_{\mathcal F}^2 \nonumber
\end{align}
where $||.||^2_{\mathcal F}$ is the Frobenius norm and $\bm{\alpha}$ is a $|\Theta| \times d_2$ matrix containing the coefficients associated with each term of $\Theta$ 
for the $d_2$ governing equations.
While a $l_1$-regularization can be directly used for promoting sparsity, Euler-SINDy makes use instead of a sequential threshold ridge regression (namely, STRidge) to prevent pathological behaviors that might occur with highly correlated data.

\paragraph{
RK4-SINDy~\citep{RK4-SINDy}:}
Consider a function $f:\mathbb{R}^2 \to \mathbb{R}$ and the ODE $\frac{du}{dt}=f(u,t)$. 
The  $s$-stage Runge-Kutta (RK) approximation improves Euler method by starting with an initial estimate of the solution and using the latter to calculate a second, more accurate approximation, and so on. Formally, $u(t+h_t)\approx u(t)+h_t \sum_{i=1}^sb_i k_i(t, h_t),$
where
\[
\left\{\begin{array}{rcl}
k_1(t, h_t) & = & f(t, u(t)) \\
k_2(t, h_t) & = & f(t+c_2h_t, \, u(t)+h_t a_{21} k_1) \\
& \vdots & \\
k_s(t, h_t) & = & f\!\left(t+c_s h_t, \, u(t)+h_t \sum_{j=1}^{s-1} a_{sj} k_j\right)
\end{array}\right.
\]
and $a_{i,j}$, $b_l$ and $c_r$ are real coefficients defining the specific RK instance. 
In RK4-SINDy, a $4$-stage RK scheme is used, offering a good balance between accuracy and cost of computation. RK4 local truncation error is of order $4$ 
with $k_1 = f(t, u(t))$, $k_2 = f(t+\frac{h_t}{2}, u(t)+h_t\frac{k_1}{2})$, $k_3 = f(t+\frac{h_t}{2}, u(t)+h_t\frac{k_2}{2})$ and $k_4 = f(t+h_t, u(t)+h_tk_3)$.  The approximation of $u(t+h_t)$ is then defined as $u(t)+ \frac{h_t}{6}(k_1+2k_2+2k_3+k_4)$ and is used in RK4-SINDy instead of the Euler estimate in Eq.~\eqref{eq:sindySTR}.

\section{Unrolled SINDy}\label{sec:Unrolled-SINDy}
\begin{figure*}[t]
\begin{center}
\vspace*{-.03\textwidth}\hspace*{-.03\textwidth}\includegraphics[width=1.06\textwidth]{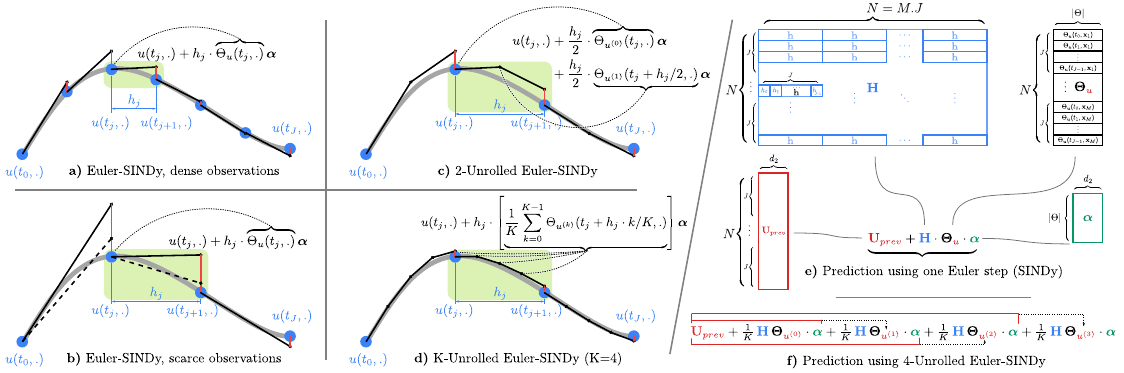} 
\caption{Left: With the solid black lines showing the predictions using the true coefficients (hence tangent to the trajectory), illustration of how SINDy operates with the Euler method associated with (a) a small time step $h_j$, and (b) where data is sparsely sampled; In the center,  we show the  benefit of (c) unrolling $K=2$ times, and (d) $K=4$ times each numerical step which leads to a smaller local truncation error; Right: (e) the matrices involved in prediction steps, and (f) the 4-Unrolled Euler-SINDy Algorithm~\ref{algo:Unrolled-SINDy}.}\label{fig:unrolling}
\end{center}
\end{figure*}

Relying on the illustrations in Fig.~\ref{fig:unrolling}, we present the principles of our unrolled scheme followed by the corresponding 
guarantees on the local truncation error.

\subsection{Unrolling the Time Derivative Numerical Approximation}\label{sec:unrolled-principles}
\newcommand{\fig}[1]{Fig.~\ref{fig:unrolling}#1}
\newcommand{\two}{\textcolor{blue}{2}}

As introduced in Sec.~\ref{sec:def}, Euler-SINDy relies on the observations $u(t_j, .)$ and the evaluation of the dictionary $\Theta_u(t_j,\mathbf{x}_m)$ at time $t_j$ to predict the solution at time $t_{j+1}$.
This is adequate when the data is densely sampled ({\it i.e.} $h_j = t_{j+1}-t_j$ is small) and the dictionary $\Theta$ contains the correct terms, as illustrated in \fig{a}.
However, when observations are sparse (\fig{b}), due to large local truncation errors, \textbf{the true coefficients no longer minimize the loss of Eq.~\eqref{eq:sindySTR}}. 
As a result, the optimization introduces additional non-zero coefficients, 
leading to an overfitting phenomenon. 
Employing RK4 makes it possible to push the limits to some extent, but RK4-SINDy   still remains constrained by small $h_j$, as it would otherwise be unable to handle certain problems.

To address this strong limitation, we thus propose Unrolled-SINDy which consists in unrolling the numerical method used to compute the prediction of the next observation.
The goal is to {\bf decorrelate the numerical integration time step from 
the data sampling rate}, thus allowing to recover the true coefficients even in case of high $h_j$.
Intuitively, when unrolled K times, the numerical method  uses $K$ intermediate steps of size $\tfrac{h_j}{K}$ to compute the next observation, instead of a single step in SINDy. This way, \textbf{the targeted theoretical parameters once again become candidates for the minimization of the loss}, as long as the effective time step remains within the numerical method’s stability region. While excessively large time steps can still break this property, increasing the unrolling depth restores consistency and enables us to solve problems that could not otherwise 
be addressed by classical methods or RK4-SINDy.

For simplicity of  exposition, in \fig{c} we show $\two$-Unrolled Euler-SINDy, the unrolling of the Euler method with $\two$ steps, but the same principle applies to RK4 and other explicit numerical methods.
Similarly to Euler-SINDy, the first step is to compute the next observation using the current one $u(t_j,.)$ and the dictionary at the current time (denoted $\Theta_{u^{(0)}}$) but with a time step that is divided by $\two$.
The second step computes the final prediction using the first-step prediction and the dictionary (denoted $\Theta_{u^{(1)}}$) evaluated using this first-step prediction and time $t_j + \tfrac{h_j}{\two}$.

More generally, $K$-Unrolled Euler-SINDy works similarly, but doing steps of size $\tfrac{h_j}{K}$, as shown in \fig{d} with $K=4$.
The dictionary (denoted $\Theta_{u^{(k)}}$, $k=0,\dots,K-1$) is thus evaluated at $K$ different time steps, $t_j + k\cdot\tfrac{h_j}{K}$, on the initial state $u(t_j,.)$ and the $K-1$ intermediate states.
The final prediction, $u^{(K)}$, can be expressed recursively as follows:
\begin{align}
u^{(k+1)} = u^{(k)} + \frac{h_j}{K} \cdot 
   \Theta_{u^{(k)}}\!\left(t_j+ k\cdot \tfrac{h_j}{K}, \,\bcdot\right) \cdot \bm{\alpha} \quad
\text{and} \quad 
u^{(0)} = u(t_j, \bcdot) ,
\label{eq:yourlabel}
\end{align}
and, by developing the recursion and factorizing, 
\begin{align}
u^{(K)}(t_j + h_j, \bcdot) = u(t_j, \bcdot) + h_j \cdot \left[\tfrac{1}{K}\sum_{k=0}^{K-1}
   \Theta_{u^{(k)}}\!\left(t_j+ k\cdot \tfrac{h_j}{K}, \bcdot\right)\right] \cdot \bm{\alpha}.
\label{eq:unrolled-factorized}
\end{align}
This formula is similar to the one used in Euler-SINDy, but with a dictionary evaluated  as the {\bf average of dictionaries at $K$ intermediate time steps}.
The formula for $K$-Unrolled RK4-SINDy (see Appendix \ref{sec:unrolled-RK4}) can be similarly factorized with an effective dictionary but with four times more terms and non-uniform weights.\\
As shown in the next section, it is worth noticing that by unrolling $K$ times the numerical step,  the Euler estimation performed in Eq.~\eqref{eq:unrolled-factorized} benefits of a truncation error on the order of ${\mathcal O} \left ( {h^2}/{K} \right )$. This is almost similar as carrying out one Euler step with a time step of $h/K$, {\bf but without requiring additional data.}

\subsection{Analysis of the Truncation Errors} \label{sec:theo}

The local truncation error of a numerical method is defined as the error induced for each approximation step. In the following, we derive an upper bound of this error when a $K$-unrolled scheme is applied on Euler, and then, more generally, on a $s$-stage Runge-Kutta method. 

\begin{theorem}\label{thm:truncation-error}
The local truncation error suffered by the unrolled Euler estimate of Eq.~\eqref{eq:unrolled-factorized} (assuming that $\forall j, h_j=h$)
is on the order ${\mathcal O} \left ( \frac{h^2}{K} \right )$ such that:
\begin{eqnarray}
\epsilon={\left (\frac{h^2}{2K} \right )\cdot \displaystyle\left\lvert  \frac{1}{K} \sum_{i=0}^{K-1} u''\left(t+\frac{hi}{K}\right)\right\rvert} \leq \left (\frac{h^2}{2K} \right )\cdot M, \label{eq:U-EULER}
\end{eqnarray}
with the constant $M=\max_{t' \in [t,t+h]}|u''(t')|.$
\end{theorem}

The proof is presented in  Appendix \ref{proof-th1}. It is based on the equality of a recursion applied $K$ times of Taylor’s theorem and Euler’s approximation. Th.~\ref{thm:truncation-error} states that as $K$ tends to infinity, $\epsilon$ converges towards 0. This means that, provided that the dictionary $\Theta$ contains the correct terms, there exists an $\bm{\alpha}$ in Eq.~\eqref{eq:unrolled-factorized} that corresponds to the governing equation and which allows a correct prediction of the next observation. The method for finding these coefficients is covered in the following section. 
Before that, and since the unrolling  can be  applied on other numerical methods, the next theorem generalizes this result when the unrolling is embedded in an $s$-stage RK method.

\begin{theorem}\label{thm:truncation-error-RK}
Assume that $\left|\frac{\partial f}{\partial u}\right|\leq L$. Then there exists $C \in \mathbb{R}^+$ such that the error $\epsilon$ of a $K$-unrolled one-step of an $s$-stage Runge-Kutta method of order $p$, with a time-step $h$ satisfies:
\begin{eqnarray}
\epsilon \leq \left(\frac{h}{K}\right)^p \frac{C}{L}\left(e^{Lh}-1\right). \label{eq:U-RKP}
\end{eqnarray}
\end{theorem}

    The proof is presented in Appendix (\ref{proof-th2}). Note that for Euler's method, {\it i.e.} $p=1$, Thm.~\ref{thm:truncation-error-RK} holds since the right hand part of the inequality of Eq.~\eqref{eq:U-EULER} is upper bounded by that of Eq.~\eqref{eq:U-RKP}.

\subsection{The Unrolled Euler-SINDy Algorithm}\label{sec:unrolled-algorithm}

Similarly to what we have shown for Euler-SINDy, 
 we can express the $K$-Unrolled Euler-SINDy algorithm as a penalized linear regression:
\begin{align}\label{eq:unrolledsindy-mat}
\min_{\bm{\alpha}} \  \left\| \mathbf{U}_{next} - \left(
    \mathbf{U}_{prev} + \mathbf{H} \cdot
    \mathmakebox[0pt][l]{\phantom{\left[ \frac{1}{K} \sum_{k=0}^{K-1} \mathbf{\Theta}_{u^{(k)}} \right]}}
    \smash{\underbrace{\left[ \frac{1}{K} \sum_{k=0}^{K-1} \mathbf{\Theta}_{u^{(k)}} \right]}_{\text{effective dictionary}}}
    \cdot \bm{\alpha} \right) \right\|_{\mathcal F}^2 \nonumber + \lambda \|\bm{\alpha}\|_{\mathcal F}^2.
\end{align}
\medskip

\fig{f} shows the prediction formula without the sum, for $K=4$.
The highlight in red emphasizes that the dictionary $\mathbf{\Theta}_{u^{(k)}}$, evaluated at intermediate steps $u^{(k)}$, has a dependency on $\bm{\alpha}$ (the coefficients to be learned). To address this problem, a first solution, fast and simple to implement~(see runtimes in Appendix~\ref{suppl:expes}), and that corresponds to the core contribution of this paper, is based on an {\bf iterative closed-form solution} of the linear regression problem. Within one iteration, this effectively discards the dependency of the dictionary on $\bm{\alpha}$. The pseudo-code of $K$-Unrolled Euler-SINDy  is presented in Algo.~\ref{algo:Unrolled-SINDy} (the algorithm is implemented with {\tt PyTorch} and {\tt sklearn}), noting that when $K=1$, we recover the original Euler-SINDy. Its extension to Unrolled RK4-SINDy in closed-form is described in the Appendix (see Algo.~\ref{algo:Unrolled-RK4SINDy}). 
Inspired by the original RK4-SINDy algorithm~\citep{RK4-SINDy} which faces the similar problematic with intermediate steps, our second solution (see Appendix \ref{sec:unrolled-sgd}),  denoted as Unrolled Euler-SINDy-SGD, consists in using a {\bf gradient descent approach}, effectively backpropagating through the unrolled prediction scheme. 

\begin{algorithm}[h]
    \newcommand{\dict}{\mathcal{D}ict}
    \newcommand{\dicteval}{\mathbf{\Theta}_{\tilde{u}}}
    \caption{$K$-Unrolled Euler-SINDy} 
    \begin{algorithmic}[1]
        \State \textbf{Input:} time steps $\mathbf{t} = \{t_j\}_{j=0..J}$; $J \times M$ Training pairs $(u(t_j,\mathbf{x}_m),u(t_{j+1},\mathbf{x}_m))$;
        \State $K$: nb of unrolling steps; $\lambda$: regularization parameter; ${\mathcal I}$: nb of iterations; $\alpha_{th}$: threshold
        \State $\dict \in \mathbb{R}^{N \times d_2}\times\mathbb{R}^{J} \rightarrow \mathbb{R}^{N \times |\Theta|}$ : a function to evaluate the dictionary
        \vspace{1mm}
        \State Initialize coefficient matrix $\bm{\alpha} \gets \mathbf{0}$ ; $\mathbf{h}$, $\mathbf{H}$ and $\mathbf{\mathcal \bm H}$ using $\{t_j\}_j$ \Comment{$\bm{\alpha} \in \mathbb{R}^{|\Theta| \times d_2}$}
        \RepeatN{${\mathcal I}$}
        \State Initialize $\tilde{\mathbf{\Theta}} \gets \mathbf{0}$\Comment{$\tilde{\mathbf{\Theta}} \in \mathbb{R}^{N \times |\Theta|}$}
        \State $\mathbf{\tilde{U}} \gets \mathbf{U}_{prev}$ \Comment{$\mathbf{{U}}_{prev}= \left ( u(t_j,\mathbf{x}_m) \right )_{j=0..J-1,m=1..M} \in \mathbb{R}^{N \times d_2}$}
        \For{$k = 0$ to $K-1$}
            \State $\dicteval \gets \dict(\mathbf{\tilde{U}}, \mathbf{t} + \frac{k}{K}\mathbf{h})$ \Comment{$\dicteval=(\Theta_{u^{(k)}}(t_j+\frac{k}{K}h_j,\mathbf{x}_m))_{j,m} \in \mathbb{R}^{N \times |\Theta|}$}
            \State $\mathbf{\tilde{U}} \gets \mathbf{\tilde{U}} + \frac{1}{K}\cdot \mathbf{H} \cdot \dicteval \cdot \bm{\alpha}$ 
            \State $\tilde{\mathbf{\Theta}} \gets \tilde{\mathbf{\Theta}} + \frac{1}{K}\cdot\dicteval$  
        \EndFor
        \State $\dot{\mathbf{U}}=(\mathbf{U}_{next}-\mathbf{U}_{prev}) \oslash \mathbf{\mathcal \bm H}$ \Comment{$\dot{\mathbf{U}} \in \mathbb{R}^{N \times d_2}$; $\mathbf{\mathcal \bm H} \in \mathbb{R}^{N \times d_2}$; $\oslash$: element-wise division}
        \State $\bm{\alpha} \gets \left( \tilde{\mathbf{\Theta}}^\top \tilde{\mathbf{\Theta}} + \lambda \mathbf{I}_{|\Theta| \times |\Theta|} \right)^{-1} \tilde{\mathbf{\Theta}}^\top \dot{\mathbf{U}}$ 
        \State Hard thresholding: $\alpha_{ij} = 0$ if $|\alpha_{ij}| < \alpha_{th}$
        \End
    \State \textbf{Output:} Final sparse coefficient matrix $\bm{\alpha}$
    \end{algorithmic} \label{algo:Unrolled-SINDy}
\end{algorithm}

\section{Experimental Results} \label{sec:expes}
In this section, we first present a comprehensive experimental study on two PDEs to highlight the main strengths of our unrolled method when combined with both Euler-SINDy and RK4-SINDy. We specifically investigate how unrolling the numerical scheme improves the equation recovery for problems with an {\bf increasing inter-observation time step} and {\bf scarcer data}. Then, we evaluate our unrolling approach on the state of the art noise-robust iNeuralSINDy with both Euler and RK4 schemes.
Note that all along these experiments, we evaluate the behavior of the methods through the $\ell_1$-norm between the ground truth coefficients $\bm{\alpha_{GT}}$ and the predicted ones $\bm{\alpha_{pred}}$. 
Additional experiments are reported in Appendix ~\ref{suppl:expes}.

\subsection{Improvements of SINDy by Unrolling}

We conduct a comparative study on two PDEs. We first consider the $2D+t$ {\it reaction–diffusion} system, which involves multiple nonlinear interaction terms together with second-order spatial derivatives. This choice highlights the ability of our method to handle systems with rich dynamics. 
Next, we address the more challenging {\it Kuramoto–Sivashinsky} PDE, which is well known for its chaotic spatio-temporal behavior and the presence of fourth-order spatial derivatives.
Due to the page limit constraints, the tables showing the recovered analytical expressions of the governing equations 
 are provided respectively in Appendix ~\ref{sec:expes-2Dreacdiffus} and Sec.~\ref{sec:kuramoto_sup}.

\subsubsection{2D Reaction-Diffusion PDE system} \label{sec:PDE}


Reaction–diffusion PDEs model physical phenomena such as the change in space and time of the concentration chemical substances. The dynamics is as follows:
\begin{eqnarray}
\left\{\begin{array}{rcl}
    u_t & = & 0.1 \, \Delta u + u - u^3 + v^3 + u^2 v - u v^2\\
    v_t & = & 0.1 \, \Delta v + v - u^3 - v^3 - u^2 v - u v^2 
\end{array}\right.\label{eq:2dreacdiff}
\end{eqnarray}
where $\Delta = \partial_{xx} + \partial_{yy}$ denotes the Laplacian operator in two dimensions, with $\partial_{xx}$ (resp. $\partial_{yy}$) representing the second-order partial derivative w.r.t $x$ (resp. $y$).

\paragraph{Experimental setup:} To create the initial dataset, using the code from PDE-FIND~\citep{PDE-FIND}, we simulate a set of 20,000 pairs with  $t \in [0,10]$ and $\Omega = [-10,10]^2$ with a time step $h=5\!\cdot\!10^{-4}$
under the initial conditions $u(x,y,t=0)=\exp(-(x^2 + y^2)/2)$, $v(x,y,t=0)=0$. The sparsity threshold $\alpha_{th}$ is set to $0.05$ while the regularization parameter $\lambda$ is set to $10^{-1}$. 
 We keep the spatial discretization fixed and produce harder and harder problems by varying the temporal resolution, i.e., adjusting the inter-observation time step $h$. By subsampling the 20,000 data points, we construct several sub-problems with time steps ranging from $1.25 \cdot 10^{-2}$ up to $1$. As $h$ increases, the number $N$ of available pairs $(u(t), u(t+h))$ decreases while the delay between observations grows.

\paragraph{Results:} We compare both Euler-SINDy and RK4-SINDy to their unrolled variants, reporting the results in Tab.~\ref{tab:reacdiff_l1_loss}.
For compactness, we present the results associated with the unrolling value $K$ yielding the best training error as the behavior with respect to $K$ is stable as illustrated in Tab.~\ref{tab:reacdiff_k_h_evolution}.

\begin{table}[t]
\centering
\caption{Robustness of Euler-SINDy (resp. RK4-SINDy) and its unrolled version on reaction-diffusion (Eq.~\eqref{eq:2dreacdiff}), on sparser and sparser observations (higher $h$). Unrolling lowers error and solves sparser problems.}
\begin{tabular}{c}
\vspace*{-.03\textwidth}\hspace*{-.03\textwidth}    \includegraphics[width=\textwidth]{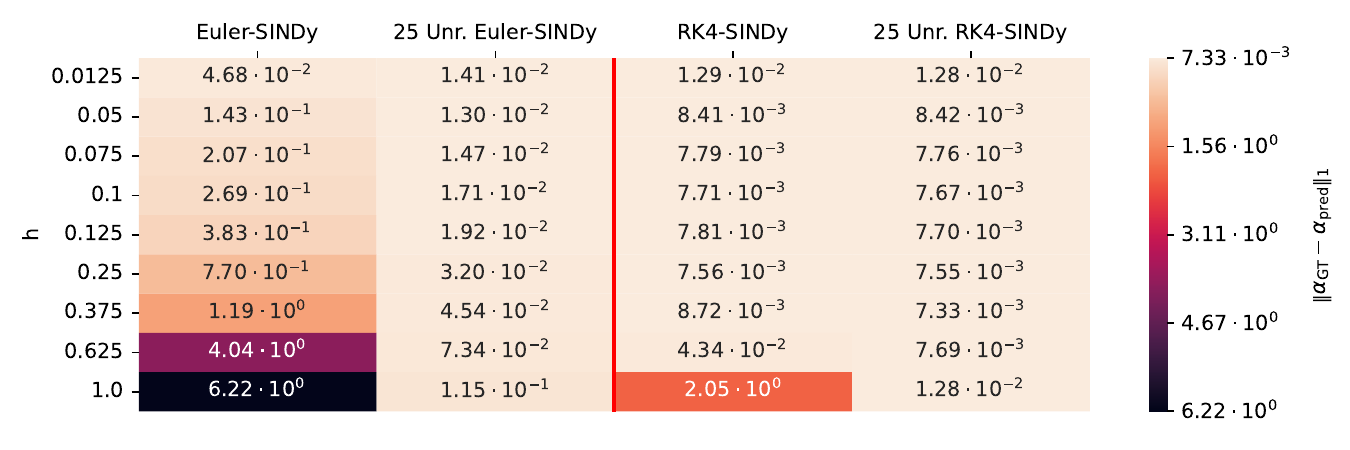}
\end{tabular}
\label{tab:reacdiff_l1_loss}
\end{table}

As expected, Euler-SINDy remains stable for simpler problems (very small step sizes) and then begins to diverge ({\it i.e.} progressively darker colors, with additional non zero terms as described in Sec.~\ref{sec:expes-2Dreacdiffus}). By contrast, its unrolled variant with $K=25$ significantly improves the robustness to large inter-observation time steps, successfully recovering the governing equations. 
A similar trend is observed for RK4-SINDy. It is able to correctly identify the dynamics up to $h=0.625$, but it fails from $h=1$ (much larger $\ell_1$ with additional terms) while its unrolled variant with $K=25$ still recovers the correct underlying equation.

\begin{table}[t]
\centering
\caption{Accuracy of $K$-Unrolled Euler-SINDy 
with different unrolling depths $K$ and observation step sizes $h$. Lighter colors indicate more accurate recovery, crosses are NaN values (details and exact values in~\ref{suppl:errvshk}).}
\begin{tabular}{c}
    \includegraphics[width=0.9\textwidth]{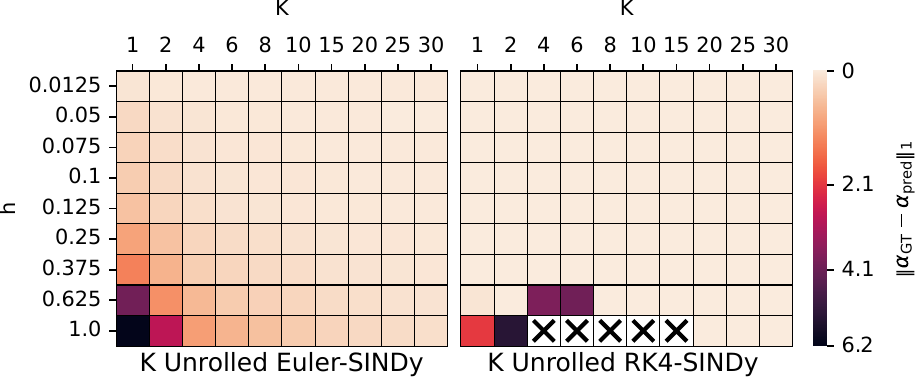}
\end{tabular}
\label{tab:reacdiff_k_h_evolution}
\end{table}

\begin{figure}[h]
\centering
\includegraphics[width=\textwidth]{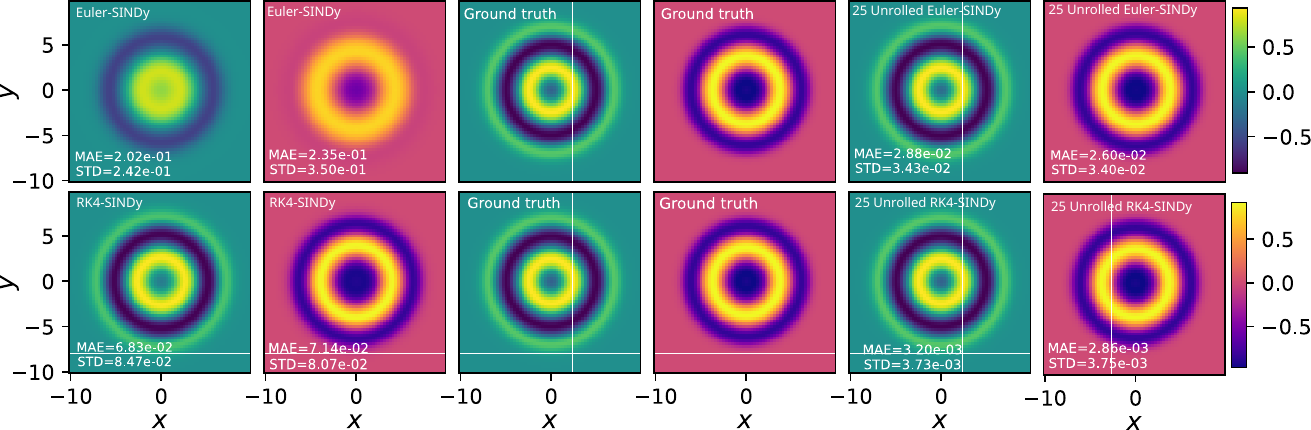}
\caption{Solutions of the 2D Reaction-Diffusion PDEs (with $h=1$ and $K=25$) 
simulated from the ground truth and the learned PDEs 
($u$ in green and $v$ in red).}
\label{fig:2D-React-Diff}
\end{figure} 

We illustrate in Fig.~\ref{fig:2D-React-Diff} the capacity of our unrolled methods to recover the governing equations when the observations are widely spaced ($h=1$). These plots come from a regular grid  $64 \times 64$  obtained at time $t=10$ of the simulation. Note that two figures per method are reported, since it is a $2D$ PDE.
While 25 Unrolled Euler-SINDy recovers quite well the ground truth  (with an MAE $\approx 10^{-2}$), it is visually apparent that Euler-SINDy without unrolling is unable to handle such a large time step and suffers from an error one order of magnitude higher. As for RK4, the difference is less visible to the eye, but is still numerically one order higher ($10^{-2}$ vs $10^{-3}$).

We report the additional computational burden of a $K$ unrolling in suppl. material (\ref{suppl:runtimes}). For instance, note that when $K=25$, the process is on average 6 to 7 times longer. Even though it may appear significant at first sight, it improves accuracy and most importantly, it allows to recover the governing equation in situations inaccessible without unrolling. 

\begin{table}[t]
\centering
\caption{Robustness of Euler-SINDy (resp. RK4-SINDy) and its unrolled version on KS (Eq.~\eqref{eq:1Dkuramoto-1}), on sparser and sparser observations (higher $h$).}
\includegraphics[width=\textwidth]{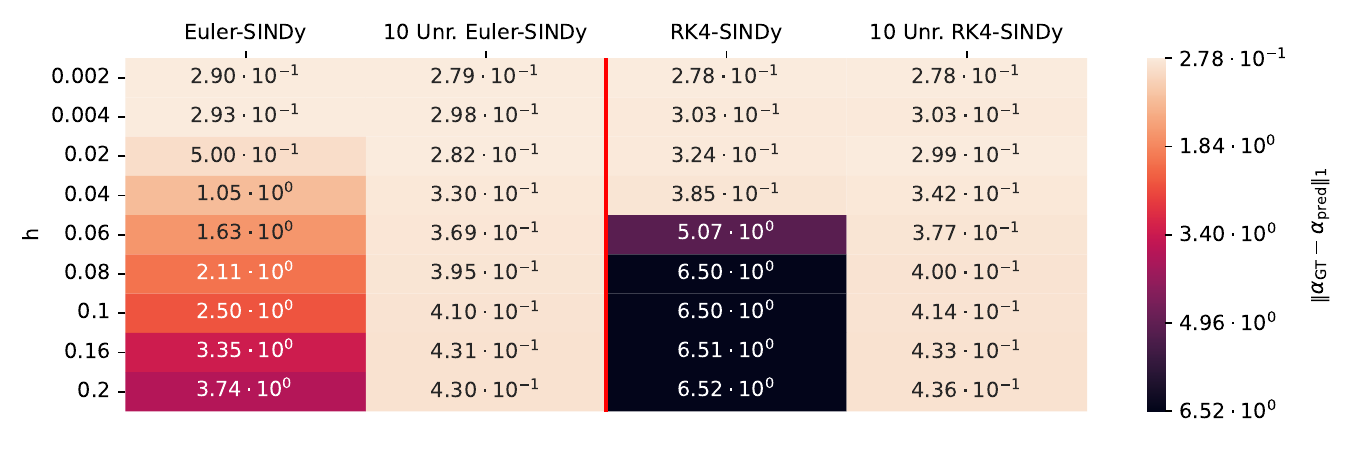}

\label{tab:kuramoto_l1_loss}
\end{table}

\subsubsection{Kuramoto–Sivashinsky PDE} \label{sec:kuramoto}

The Kuramoto–Sivashinsky (KS) equation describes spatiotemporal dynamics commonly observed in pattern-forming physical systems, such as instabilities in laminar flame fronts or fluid flows. In one spatial dimension, it is expressed as the following PDE:
\begin{eqnarray}
    u_t =  - u_{xx} - u_{xxxx} - 5 uu_x , \label{eq:1Dkuramoto-1}
\end{eqnarray}
where $u_t$ denotes the time derivative, $u_{x}$, $u_{xx}$ and $u_{xxxx}$ denote the first, second and fourth spatial derivatives, respectively, and $uu_x$ is the nonlinear term.

\paragraph{Experimental setup:} We simulate the solution using a solver implemented in JAX based on Exponential Time Differencing. We simulate the equation on the spatial domain \(\Omega = [0, L]\) with \(L = 64\), discretized into \(N = 100\) points with spatial step \(\Delta x = \frac{L}{N}\), over the time interval \(t \in [0, 200]\) with time step \(\Delta t = 0.001\). The initial condition is chosen as gaussian $u(x,t=0) = 0.5 \exp\left(-100 \left(x - \frac{L}{2}\right)^2 \right).$ $\alpha_{th}$ is set to $0.1$ and the regularization parameter $\lambda$ to $10^{-6}$.

\begin{figure}[t]
\centering
\includegraphics[width=0.9\textwidth]{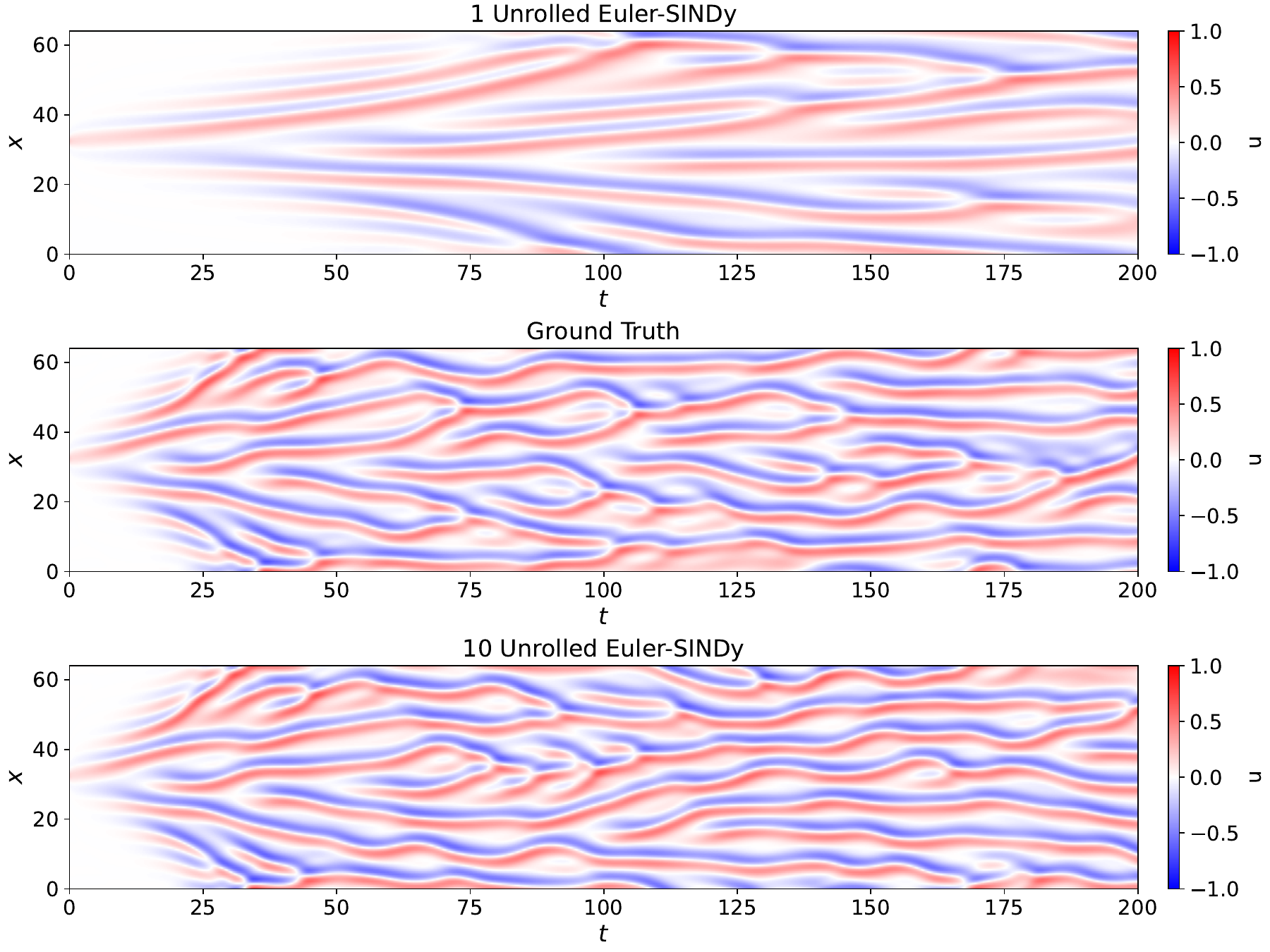}
\caption{Solutions of the Kuramoto–Sivashinsky PDE with $h=0.2$. From top to bottom: Euler-SINDy, ground truth, and 10 Unrolled Euler-SINDy.}
\label{fig:kuramoto_figure_main}
\end{figure}

\paragraph{Results:} Tab.~\ref{tab:kuramoto_l1_loss} shows the results on the KS equation for Euler-SINDy, RK4-SINDy and their unrolled versions. 
Again, without unrolling, the existing methods quickly diverge and produce incorrect equations while their respective unrolled variants  recover this complex fourth-order PDE even in challenging scenarios (i.e. higher $h$).
We can note that the transition to failure is more abrupt with RK4-SINDy.
Fig.~\ref{fig:kuramoto_figure_main} highlights the qualitative gap between Euler-SINDy and 10 Unrolled Euler-SINDy in this difficult setting. The former totally fails to capture the essential dynamics of the system.
In contrast, even though local differences appear between the ground truth and Unrolled Euler-SINDy due to the chaotic nature of this PDE, our method tracks the dominant trends and spatial structures much more accurately, yielding coefficients that remain very close to the true KS dynamics. The same behavior is observed for RK4-SINDy as shown in the suppl. material \eqref{suppl:expes}. 

\subsection{Unrolling iNeural-SINDy}
iNeural-SINDy~\citep{ForootaniGB25} is a robust extension of the SINDy method, designed to {\bf handle both noisy and scarce data}. It combines neural networks, sparse regression, and an integral formulation to stabilize the discovery of governing equations. In this last series of experiments, we embed our unrolling scheme into iNeural-SINDy, showing that unrolling also benefits this neural method, for both Euler and RK4.

We focus here on the cubic damped oscillator (results on other equations are reported in suppl.~\ref{suppl:ineural}).
This ODE is a two-dimensional nonlinear dynamical system governed by polynomial interactions of degree three. The governing equations are given by:
\begin{eqnarray}
\left\{\begin{array}{rcl}
\dot{x}(t) & = & -0.1x^3(t)+2.0y^3(t) \\
\dot{y}(t) & = & -2.0x^3(t)-0.1y^3(t)
\end{array}\right. \label{eq:oscillator}
\end{eqnarray}
where $\dot{x}(t)$ and $\dot{y}(t)$ denote the time derivatives of the state variables $x(t)$ and $y(t)$, respectively.



\begin{table*}[t]
\centering

\begin{subtable}{0.7\textwidth}
    \includegraphics[width=\linewidth]{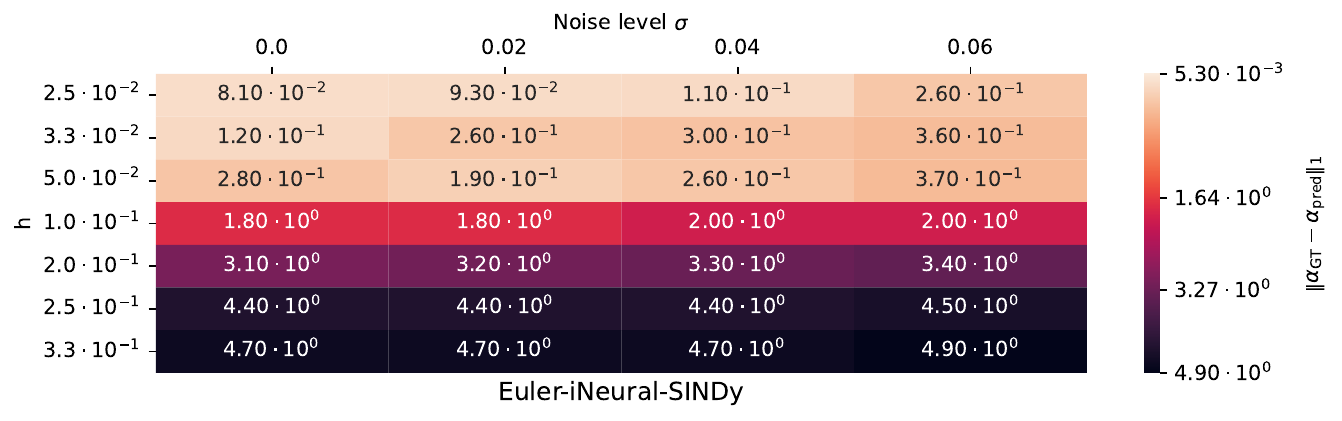}
    \caption{}
\end{subtable}

\begin{subtable}{0.7\textwidth}
    \includegraphics[width=\linewidth]{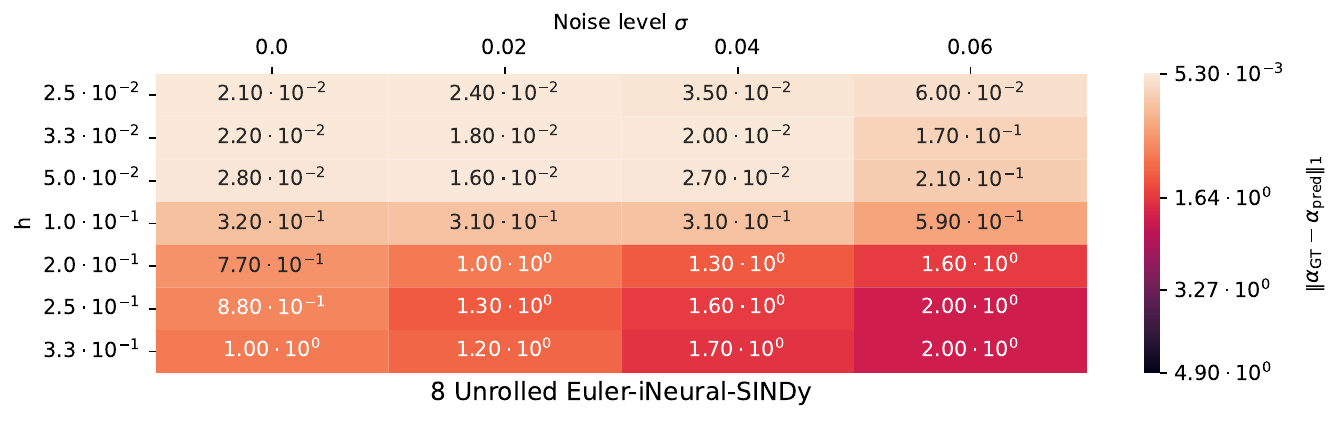}
    \caption{}
\end{subtable}

\begin{subtable}{0.7\textwidth}
    \includegraphics[width=\linewidth]{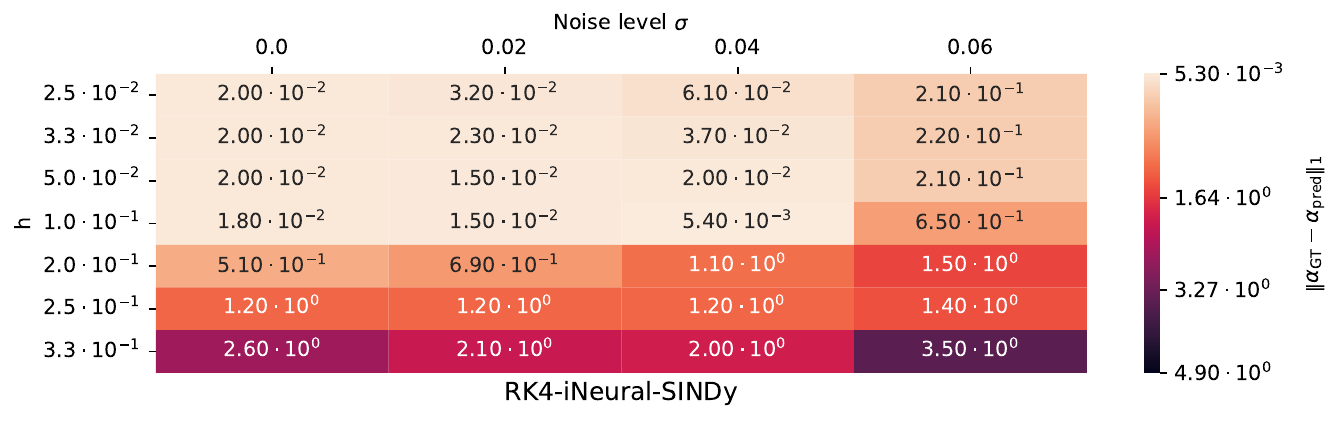}
    \caption{}
\end{subtable}

\begin{subtable}{0.7\textwidth}
    \includegraphics[width=\linewidth]{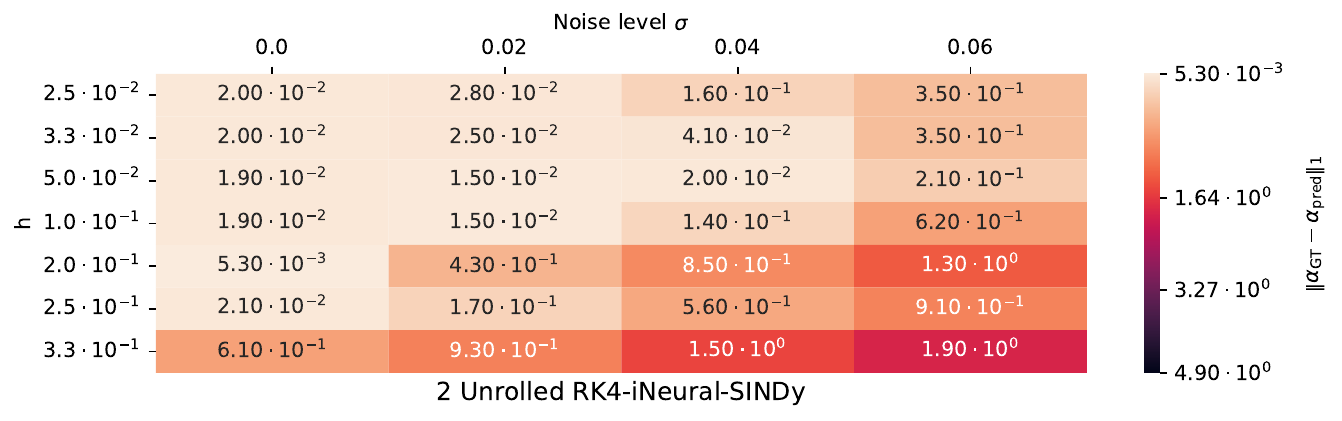}
    \caption{}
\end{subtable}

\caption{Robustness of iNeural-SINDy on the cubic damped oscillator (Eq.~\ref{eq:oscillator}), evaluated with increasing time step $h$ and  noise level $\sigma$ with A) Euler-iNeural-SINDy, B) 8 Unrolled Euler-iNeural-SINDy, C) RK4-iNeural-SINDy, and D) 2 Unrolled RK4-iNeural-SINDy.}
\label{tab:oscillator_l1_loss_ineural}
\end{table*}


\paragraph{Experimental setup:} 
We adopt the same setup as in the original paper. Noisy observations are generated as follows: $\tilde{x}(t) = x(t) + \eta_x(t),$ and $\tilde{y}(t) = y(t) + \eta_y(t),$
where $\eta_x(t)$ and $\eta_y(t)$ are independent Gaussian noise with standard deviations $\sigma_x = \sigma \, \mathrm{std}(x(t))$ and $\sigma_y = \sigma \, \mathrm{std}(y(t))$ respectively, with $\sigma \in [0,0.06]$ controlling the relative noise amplitude. The experiments are conducted over a range of observation intervals $h \in [0.025, 0.333]$, allowing us to systematically evaluate the robustness of each method to increasing $h$ and $\sigma$.

\begin{figure}[h]
\centering
\includegraphics[width=0.8\textwidth]{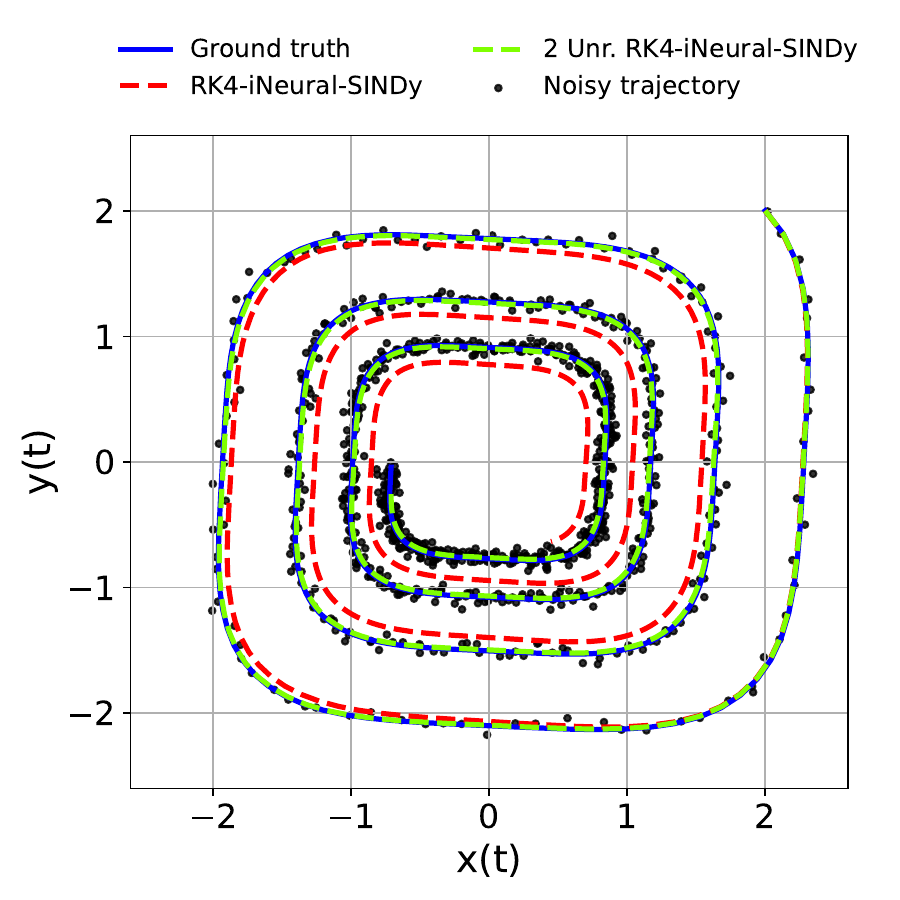}
\caption{Solution of cubic damped oscillator with noisy data ($\sigma=0.02$) using a) RK4 iNeural-SINDy and b) 2 Unrolled RK4-iNeural-SINDy.} \label{fig:oscillator-ineural_plot}
\end{figure}

\paragraph{Results:}
The comparative results are reported in Tab.~\ref{tab:oscillator_l1_loss_ineural}. As already shown in the original paper, iNeural-SINDy remains relatively robust to the presence of noise. However, we can note that for both Euler (Tab.~\ref{tab:oscillator_l1_loss_ineural}.a) and RK4 (\ref{tab:oscillator_l1_loss_ineural}.c),  it becomes less accurate as the gap between the data increases. This limitation is substantially mitigated by the unrolled variants (see Tab.~\ref{tab:oscillator_l1_loss_ineural}.b and ~\ref{tab:oscillator_l1_loss_ineural}.d). Indeed, a  8 Euler-iNeural-SINDy allows to better capture the system’s underlying structure and achieve a consistent reduction of the $\ell_1$ norm between the theoretical and predicted coefficients. A similar trend is observed for a simple 2 RK4-iNeural-SINDy.
Fig.~\ref{fig:oscillator-ineural_plot} illustrates the trajectory reconstructed by RK4 iNeural-SINDy and its unrolled variant with noisy data ($\sigma=0.02$). While iNeural-SINDy is designed to be robust to noise, the standard RK4 implementation (in red)  still gradually deviates from the true dynamics (in blue) as the simulation progresses. In contrast, the 2 Unrolled RK4-iNeural-SINDy (green curve) keeps coinciding almost perfectly with the ground truth.   

\section{Limitations}\label{sec:limitations}
We observed that our closed-form-based method occasionally produces NaN values. This typically occurs when an estimate diverges during the unrolling process, rendering the matrix inversion step numerically unstable. While such occurrences are difficult to anticipate, a practical fallback is to use our SGD-based solution in these cases (see Appendix~\ref{sec:locally-linearCF}).
We have shown that unrolling benefits different combinations of integration schemes (Euler and RK4) and methods (vanilla SINDy and state-of-the-art iNeural-SINDY).
While we believe that unrolling can benefit many SINDy variants, its effect needs to be quantified experimentally.
Like most SINDy settings, we assumed that the coefficients $\bm{\alpha}$ do not change over time, while some dynamics are modeled by varying-coefficient PDEs. Adapting our method to this complex scenario requires substantial additional work.


\section{Conclusion}

We propose Unrolled-SINDy, a PDE discovery algorithm that relies on an unrolling scheme applied at each numerical approximation stage of the time derivatives. This unrolling strategy, independent of any specific integration method, allows the classic SINDy-based algorithms to overcome the strict time step limitations that typically hinder explicit schemes. We further demonstrate that our methodology can be effciently embedded in PDE discovery neural methods.  
To benefit from the best of both worlds, it would be appropriate to explore
 the extension of our approach to implicit numerical methods. 

\appendix
\thispagestyle{empty}

\section{Algorithmic Details}\label{suppl:algo}

\subsection{Pseudo-code of Unrolled RK4-SINDy} \label{sec:unrolled-RK4}

The equivalent of Eq.~\ref{eq:unrolled-factorized} for the factorized expression of our unrolled scheme for RK4 is defined as follows:

\begin{equation}
\begin{aligned}
u^{(K)}(t_j + h_j,\, \cdot)
  &= \underbrace{
      u(t_j,\, \cdot)
      + h_j \cdot
      \left[
        \frac{1}{K}
        \sum_{k=0}^{K-1}
        \Theta^{(k)}\!\left(t_j + k \cdot \frac{h_j}{K},\, \cdot\right)
      \right]
      \cdot \bm{\alpha}
     }_{\text{K-Unrolled RK4 estimate}}
\end{aligned}
\label{eq:unrolled-factorized-RK4}
\end{equation}
where $\Theta{^{(k)}}$ is the weighted average estimate as computed with the RK4 method (see Sec.~\ref{sec:def}) and used (with matrix notations) in line 13 in the following pseudo-code of our Unrolled RK4-SINDy algorithm.

\begin{algorithm}
    \newcommand{\dict}{\mathcal{D}ict}
    \newcommand{\dicteval}[1]{\mathbf{\Theta}_{#1}}
    \caption{Unrolled RK4-SINDy}
    \begin{algorithmic}[1]
        \State \textbf{Input:} Training pairs $\{(u(t_j,\mathbf{x}_m),u(t_{j+1},\mathbf{x}_m))\}_{j=0..J-1, m=1..M}$; time steps $\mathbf{t} = \{t_j\}_{j=0..J}$
        \State $K$: nb of unrolling steps; $\lambda$: regularization parameter; ${\mathcal I}$: nb of iterations; $\alpha_{th}$: threshold
        \State $\dict \in \mathbb{R}^{N \times d_2}\times\mathbb{R}^{J} \rightarrow \mathbb{R}^{N \times |\Theta|}$ : a function to evaluate the dictionary
        \vspace{1mm}
        \State Initialize coefficient matrix $\bm{\alpha} \gets \mathbf{0}$ ; $\mathbf{h}$, $\mathbf{H}$ and $\mathbf{\mathcal \bm H}$ using $\{t_j\}_j$ \Comment{$\bm{\alpha} \in \mathbb{R}^{|\Theta| \times d_2}$}
        \RepeatN{${\mathcal I}$}
        \State Initialize $\tilde{\mathbf{\Theta}} \gets \mathbf{0}$\Comment{$\tilde{\mathbf{\Theta}} \in \mathbb{R}^{N \times |\Theta|}$}
        \State $\mathbf{\tilde{U}} \gets \mathbf{U}_{prev}$
        \For{$k = 0$ to $K-1$}
            \State $\dicteval{1} \gets \dict(\mathbf{\tilde{U}}, \mathbf{t} + \frac{k}{K}\mathbf{h})$
            \State $\dicteval{2} \gets \dict(\mathbf{\tilde{U}} + \frac{1}{2K}\cdot \mathbf{H} \cdot \dicteval{1} \cdot \bm{\alpha}, \mathbf{t} + \frac{k + \sfrac{1}{2}}{K}\mathbf{h})$
            \State $\dicteval{3} \gets \dict(\mathbf{\tilde{U}} + \frac{1}{2K}\cdot \mathbf{H} \cdot \dicteval{2} \cdot \bm{\alpha}, \mathbf{t} + \frac{k + \sfrac{1}{2}}{K}\mathbf{h})$
            \State $\dicteval{4} \gets \dict(\mathbf{\tilde{U}} + \frac{1}{K}\cdot \mathbf{H} \cdot \dicteval{3} \cdot \bm{\alpha}, \mathbf{t} + \frac{k + 1}{K}\mathbf{h})$
            \State $\dicteval{} \gets \frac{1}{6}(\dicteval{1}+2 \dicteval{2}+2\dicteval{3}+\dicteval{4})$
            \State $\mathbf{\tilde{U}} \gets \mathbf{\tilde{U}} + \frac{1}{K}\cdot \mathbf{H} \cdot \dicteval{} \cdot \bm{\alpha}$ 
            \State $\tilde{\mathbf{\Theta}} \gets \tilde{\mathbf{\Theta}} + \frac{1}{K}\cdot\dicteval{}$ 
        \EndFor
        \State $\dot{\mathbf{U}}=(\mathbf{U}_{next}-\mathbf{U}_{prev}) \oslash \mathbf{\mathcal \bm H}$ \Comment{$\dot{\mathbf{U}} \in \mathbb{R}^{N \times d_2}$; $\mathbf{\mathcal \bm H} \in \mathbb{R}^{N \times d_2}$; $\oslash$: element-wise division}
        \State $\bm{\alpha} \gets \left( \tilde{\mathbf{\Theta}}^\top \tilde{\mathbf{\Theta}} + \lambda \mathbf{I}_{|\Theta| \times |\Theta|} \right)^{-1} \tilde{\mathbf{\Theta}}^\top \dot{\mathbf{U}}$ 
        \State Hard thresholding~\cite{PDE-FIND}: $\alpha_{ij} = 0$ if $|\alpha_{ij}| < \alpha_{th}$
        \End
    \State \textbf{Output:} Final sparse coefficient matrix $\bm{\alpha}$
    \end{algorithmic} \label{algo:Unrolled-RK4SINDy}
\end{algorithm}

\subsection{Unrolled *-SINDy-SGD} \label{sec:unrolled-sgd}

Instead of using the closed-form solution as used in the core of the paper, another strategy consists in resorting to a stochastic gradient descent (SGD) approach.\\
To design the Unrolled Euler-SINDy-SGD (resp. Unrolled RK4-SINDy-SGD) stochastic gradient descent version of the algorithm, one can initialize the vector $\bm{\alpha}$ randomly, choose a learning rate $\eta$ and repeatedly do stochastic gradient descent by running the algorithm where lines 13-14 in Algorithm~\ref{algo:Unrolled-SINDy} (resp. lines 17-18 in Algorithm~\ref{algo:Unrolled-RK4SINDy}) are replaced by a gradient step using the gradient obtained through auto-differentiation:

$$
\bm{\alpha} \gets \bm{\alpha} - \eta \cdot \nabla_{\bm{\alpha}} ||\mathbf{U}_{next} - \mathbf{\tilde{U}}||_\mathcal{F}^2.
$$
It then becomes unnecessary in the algorithm to keep track of $\tilde{\bm{\Theta}}$.

\subsection{Locally Linearized Closed-Form Resolution} \label{sec:locally-linearCF}

Even though we leave the evaluation of this approach as future work, it is also possible to use a hybrid approach which iterates on a closed form (like the closed form version) but uses automatic differentiation (like the SGD version) to estimate a more accurate effective dictionary that takes into account the dependency of the effective dictionary on the parameters $\bm{\alpha}$.

\newcommand{\pred}{pred}
\colorlet{highlightvar}{red!50!black}
\newcommand{\dalpha}{\textcolor{highlightvar}{\delta \bm{\alpha}}}
We denote as $\pred(\bm{\alpha})$ the prediction of the unrolled SINDy algorithm, based on a current estimate of $\bm{\alpha}$.
For the sake of generality (to include $K$-Unrolled Euler-SINDy, $K$-Unrolled RK4-SINDy and RK4-SINDy), we assume the prediction is a function of the following form, involving an effective dictionary that is a function of $\bm{\alpha}$ (denoted $\Theta(\bm{\alpha})$ to insist on this dependence, and which corresponds to $\tilde{\bm{\Theta}}$ in Algorithms~\ref{algo:Unrolled-SINDy} and~\ref{algo:Unrolled-RK4SINDy}):
\begin{align}
\forall j,m \;\; \pred(\bm{\alpha})(t_j, x_m) = u(t_j, x_m) + h_j \Theta(\bm{\alpha})(t_j, x_m) \cdot \bm{\alpha}
\end{align}
in more compact algorithmic/matrix form:
\begin{align}
\pred(\bm{\alpha}) = \mathbf{U}_{prev} + \mathbf{H} \cdot \Theta(\bm{\alpha}) \cdot \bm{\alpha}
\end{align}
reminding that $\mathbf{U}_{prev} \in \mathbb{R}^{N \times d_2}$, $\mathbf{H} \in \mathbb{R}^{N \times N}$, $\Theta(\bm{\alpha}) \in \mathbb{R}^{N \times |\Theta|}$ and $\bm{\alpha} \in \mathbb{R}^{|\Theta| \times d_2}$, and thus $\pred \in (\mathbb{R}^{|\Theta| \times d_2} \rightarrow \mathbb{R}^{N \times d_2})$.

\newcommand{\NOcdot}{} 

We can linearize the prediction using the Taylor expansion at the first order, around the current estimate $\bm{\alpha}$, with a variation $\dalpha \in \mathbb{R}^{|\Theta| \times d_2}$:
\begin{align}
\pred(\bm{\alpha} + \dalpha) & \approx \pred(\bm{\alpha}) + \frac{\partial \pred}{\partial \bm{\alpha}}(\bm{\alpha}) \cdot \dalpha \\
& = \pred(\bm{\alpha}) + \mathbf{H} \cdot \left[\Theta(\bm{\alpha}) + \frac{\partial \Theta}{\partial \bm{\alpha}}(\bm{\alpha}) \cdot \bm{\alpha}\right] \cdot \dalpha \\
& = \pred(\bm{\alpha}) + \mathbf{H} \cdot \left[\Theta(\bm{\alpha}) + \bm{J} \NOcdot \bm{\alpha}\right] \cdot \dalpha \\
& = \mathbf{U}_{prev} + \mathbf{H} \cdot \Theta(\bm{\alpha}) \cdot \bm{\alpha} + \mathbf{H} \cdot \left[\Theta(\bm{\alpha}) + \bm{J} \NOcdot \bm{\alpha}\right] \cdot \dalpha \\
\end{align}
where $\bm{J} = \frac{\partial \Theta}{\partial \bm{\alpha}}(\bm{\alpha}) \in \mathbb{R}^{(N \times |\Theta|) \times (|\Theta| \times d_2)}$ is the Jacobian of the effective dictionary with respect to $\bm{\alpha}$, evaluated at the current estimate $\bm{\alpha}$, which can be obtained by automatic differentiation.
The tensor product of $\bm{J}$ by $\bm{\alpha}$ is to be understood as
$\left(\bm{J}\NOcdot\bm{\alpha} \right)_{ij} = \sum_{k,l} \bm{J}_{ijkl} \bm{\alpha}_{kl}$.

Neglecting higher order terms (in the Taylor expansion), we can iteratively update the estimate of $\bm{\alpha}$ by solving the following linear system, in $\dalpha$:
\begin{align}
    \min_{\dalpha} \left\lVert \mathbf{U}_{next} - \mathbf{U}_{prev} - \mathbf{H} \cdot \Theta(\bm{\alpha}) \cdot \bm{\alpha} - \mathbf{H} \cdot \left[\Theta(\bm{\alpha}) + \bm{J} \NOcdot \bm{\alpha}\right] \cdot \dalpha \right\rVert_\mathcal{F}^2 + \lambda \left\lVert \bm{\alpha} + \dalpha \right\rVert_\mathcal{F}^2
\end{align}
i.e.

\begin{align}
    \min_{\dalpha} \left\lVert \left[ \mathbf{U}_{next} - \mathbf{U}_{prev} - \mathbf{H} \cdot \Theta(\bm{\alpha}) \cdot \bm{\alpha} \right] - \mathbf{H} \cdot \tilde{\tilde{\bm{\Theta}}} \cdot \dalpha \right\rVert_\mathcal{F}^2 + \lambda \left\lVert \left[-\bm{\alpha}\right] - \dalpha \right\rVert_\mathcal{F}^2
\end{align}
with

\begin{align}
\tilde{\tilde{\bm{\Theta}}}
&= \Theta(\bm{\alpha}) + \bm{J} \NOcdot \bm{\alpha}
 = \tilde{\bm{\Theta}} + \bm{J} \NOcdot \bm{\alpha}.
\end{align}

In the end, we are using an updated effective dictionary $\tilde{\tilde{\bm{\Theta}}}$ that is the original effective dictionary $\tilde{\bm{\Theta}}$ plus $\bm{J}\bm{\alpha}$.
This minimization can be achieved in closed form:

\begin{align}
    \dalpha & =
    \left[
    \tilde{\tilde{\bm{\Theta}}}^\top \tilde{\tilde{\bm{\Theta}}} + \lambda \mathbf{I}_{|\Theta| \times |\Theta|}
    \right]^{-1} 
    \left[
     \tilde{\tilde{\bm{\Theta}}}^\top
     \dot{\bm{U}} - \lambda \bm{\alpha} \right]
\end{align}
with

\begin{align}
    \dot{\bf{U}}
    &= \left( \mathbf{U}_{next} - \mathbf{U}_{prev} - \mathbf{H} \cdot \tilde{\bm{\Theta}} \cdot \bm{\alpha} \right) \oslash \mathbf{\mathcal \bm H} \\
    &= \left( \mathbf{U}_{next} - \mathbf{U}_{prev} \right) \oslash \mathbf{\mathcal \bm H} - \tilde{\bm{\Theta}} \cdot \bm{\alpha}. 
\end{align}
Then the estimate of $\bm{\alpha}$ can be updated as:

\begin{align}
    \bm{\alpha} \leftarrow \bm{\alpha} + \dalpha.
\end{align}

\section{Proofs}\label{suppl:proofs}
We detail in the following the proofs of Theorems~\ref{thm:truncation-error} and \ref{thm:truncation-error-RK}.

\subsection{Proof of Theorem.~\ref{thm:truncation-error}} \label{proof-th1}

{\bf Theorem 1.}
The local truncation error of a $K$-unrolled one-step of Euler's method of time-step $h$ of the (at least) twice time-differentiable function $u(t)$ is on the order ${\mathcal O} \left ( \frac{h^2}{K} \right )$ such that:
\begin{eqnarray}
\epsilon={\left (\frac{h^2}{2K} \right ). \displaystyle\left\lvert  \frac{1}{K} \sum_{i=0}^{K-1} u''\left(t+\frac{hi}{K}\right)\right\rvert} \leq \left (\frac{h^2}{2K} \right ).M,
\end{eqnarray}
with the constant $M=\max_{t' \in [t,t+h]}|u''(t')|.$

\begin{proof}
By applying recursively $K$ times Taylor's theorem, we get:
\begin{eqnarray}
    u(t+h)=u(t)+\frac{h}{K}\sum_{i=0}^{K-1} u'(t+\frac{hi}{K})+\left ( \frac{h}{K} \right )^2\frac{\sum_{i=0}^{K-1} u''(t+\frac{hi}{K})}{2}. \label{eq:Taylor}
\end{eqnarray}
Then applying recursively $K$ times Euler's approximation, we get:
\begin{eqnarray}
    \tilde{u}(t+h)=u(t)+\frac{h}{K}\sum_{i=0}^{K-1} u'(t+\frac{hi}{K}). \label{eq:Euler}
\end{eqnarray}
From Eq.~\eqref{eq:Taylor} and \eqref{eq:Euler}, we deduce that the local truncation error is equal to: $$\epsilon=| u(t+h)-\tilde{u}(t+h)|=   \left (\frac{h}{K} \right )^2 .{\displaystyle\left\lvert \frac{1}{2}\sum_{i=0}^{K-1} u''(t+\frac{hi}{K}) \right\rvert}. $$
\end{proof}

\subsection{Proof of Theorem~\ref{thm:truncation-error-RK}} \label{proof-th2}

Assume that $\left|\frac{\partial f}{\partial u}\right|\leq L$. Then there exists $C \in \mathbb{R}^+$ such that the error $\epsilon$ of a $K$-unrolled one-step of an $s$-stage Runge-Kutta method of order $p$ of time-step $h$ of the (at least) twice time-differentiable function $u(t,\mathbf{x})$ satisfies:
$$\epsilon \leq \left(\frac{h}{K}\right)^p \frac{C}{L}\left(e^{Lh}-1\right).$$

\begin{proof}
We consider the ODE
\[
\frac{du}{dt}=f(u,t)
\]
with the initial solution $u(t)=u_0$. We want to determine $u(t+h)$ using a $K$-unrolled Runge-Kutta method.

We start at $t$ and compute $u_1$, the approximation of $\widetilde{u}_0(t+\frac{h}{K})$ with the Runge-Kutta method, where $\widetilde{u}_0$ is the solution of the equation such that $\widetilde{u}_0(t)=u_0$. From $u_1$, we compute $u_2$, the approximation of $\widetilde{u}_1(t+2\frac{h}{K})$, where $\widetilde{u}_1$ is the solution of the equation such that $\widetilde{u}_1(t+\frac{h}{K})=u_1$. We repeat the process until we finally obtain $u_K$, the approximation of $\widetilde{u}_{K-1}(t+h)$, where $\widetilde{u}_{K-1}$ is the solution of the equation such that $\widetilde{u}_{K-1}(t+(K-1)\frac{h}{K})=u_{K-1}$. Notice that $\widetilde{u}_0$ is the solution of the initial ODE.

The local error $e_i=|\widetilde{u}_{i-1}(t+i\frac{h}{K})-u_i|\leq C (\frac{h}{K})^{p+1}$ (cf. \cite[Thm. 3.1]{Hairer} for the value of $C$) is transported to the final point, i.e., $\widetilde{u}_i$ and $\widetilde{u}_{i-1}$ will differ at $t+h$ by $E_i=|\widetilde{u}_i(t+h)-\widetilde{u}_{i-1}(t+h)|$. If we apply successively the Runge-Kutta method, then the transported errors will add up, as the global error is $\epsilon=|u(t+h)-u_K|=|\widetilde{u}_0(t+h)-u_K|$ and $E_K=e_K$. One can then show (\cite[Thm. 3.4]{Hairer}) that $\epsilon$ satisfies
\[
\epsilon \leq \left(\frac{h}{K}\right)^p \frac{C}{L}\left(e^{Lh}-1\right).
\]

\end{proof}

\section{Supplementary Experimental Results} \label{suppl:expes}

We report in this section additional results about the running time of the unrolled versions, as well as some details about the experiments performed on the ODE and PDE used in the paper. Finally, we present supplementary experiments about the robustness of the methods in the presence of corrupted data with Gaussian noise.

\subsection{Comparison of the methods in terms of complexity and running time} \label{suppl:runtimes}

In terms of memory, unrolling only requires to store a copy of the coefficients to compute the effective dictionary. In practice, this is negligible compared to the other steps of the algorithms (matrix inversion, ...).

In terms of computations, unrolling $K$ times, multiplies the number of dictionary evaluations by $K$. In practice, the actual running time is often controlled more by the number of iterations necessary for convergence and by the closed-form formula. We resort to empirical measurements of running time to better quantify the cost of unrolling in practice.

We report in the following tables the additional computational burden of the unrolled versions compared with Euler-SINDy and RK4-SINDy on the Advection, 2D-Reaction-Diffusion, Kuramoto-Sivashinsky PDEs. Even though this additional cost may be sometimes significant, it allows to improve the reconstruction accuracy and  to recover the underlying equation in situations inaccessible without unrolling.

\paragraph{Hardware.} All experiments were conducted on a high-performance computing node equipped with dual AMD EPYC 7F72 (Rome) processors (48 physical cores in total) and 512~GB of DDR4 memory. All computations were performed on CPU.

\paragraph{Closed-form execution time measurement.}
For the closed-form SINDy models, the execution time is measured at each iteration for a maximum of 50 iterations. However, note that the training process can be terminated earlier based on an early stopping criterion: if the average change in the coefficient matrix $\bm{\alpha}$ over a sliding window of 5 iterations goes below a tolerance of $10^{-6}$, the optimization stops. At each iteration, a new dictionary is constructed from the current input data, and $\bm{\alpha}$ is learned through a closed-form ridge regression ($\lambda = 10^{-2}$). A thresholding procedure is used to identify active terms (entries of $\bm{\alpha}$  greater the $5 \times 10^{-2}$), and in the subsequent iterations, a reduced dictionary containing only these active terms is used. The full dataset is processed at each iteration (no batching), and the execution time is saved  and accumulated until convergence (which occurred after 7 iterations in this case).

\begin{table}[h]
\caption{Table: Running Time on the Advection PDE}
\label{tab:exec-time-Advection}
\resizebox{\columnwidth}{!}{%
\begin{tabular}{|c|c|c!{\color{red}\vrule width 2pt}c|c|}
\hline
\begin{tabular}{c}$h$=2e-03, $N$=$1000$\end{tabular} & \textbf{Euler-SINDy} & \textbf{25 Unr. Euler-SINDy} & \textbf{RK4-SINDy} & \textbf{25 Unr. RK4-SINDy} \\
\hline
$time[s]$ & \begin{tabular}{l}$7.2 \pm0.5$\end{tabular} & \begin{tabular}{l}$12.0 \pm0.4$\end{tabular} & \begin{tabular}{l}$7.2 \pm0.4$\end{tabular} & \begin{tabular}{l}$27.9 \pm0.7$\end{tabular} \\ \hline
\end{tabular}
}
\end{table}

\begin{table}[h]
\caption{Running Time on the 2D Reaction-Diffusion PDEs}
\label{tab:execution-time-2D-Diff-Reac}
\raggedright
\resizebox{\columnwidth}{!}{%
\begin{tabular}{|c|c|c!{\color{red}\vrule width 2pt}c|c|}
\hline
$h$=$0.1$, $N$=$100$ & \textbf{Euler-SINDy} & \textbf{25 Unr. Euler-SINDy} & \textbf{RK4-SINDy} & \textbf{25 Unr. RK4-SINDy} \\
\hline
$time [s]$ & \begin{tabular}{l}$94 \pm2$\end{tabular} & \begin{tabular}{l}$594 \pm 12$\end{tabular} & \begin{tabular}{l}$298 \pm 7$\end{tabular} & \begin{tabular}{l}$2001 \pm 68$\end{tabular} \\ \hline
\end{tabular}
}
\end{table}

\begin{table}[h]
\caption{Running Time on the Kuramoto-Sivashinsky PDE}
\label{tab:execution-time-KS}
\raggedright
\resizebox{\columnwidth}{!}{%
\begin{tabular}{|c|c|c!{\color{red}\vrule width 2pt}c|c|}
\hline
$h$=$0.02$, $N$=$10000$ & \textbf{Euler-SINDy} & \textbf{10 Unr. Euler-SINDy} & \textbf{RK4-SINDy} & \textbf{10 Unr. RK4-SINDy} \\
\hline
$time [s]$ & \begin{tabular}{l}$27 $\end{tabular} & \begin{tabular}{l}$70 $\end{tabular} & \begin{tabular}{l}$24 $\end{tabular} & \begin{tabular}{l}$200 $\end{tabular} \\ \hline
\end{tabular}
}
\end{table}

In the following, we compare the running time of the closed-form-based methods and their GD counterparts on the 2-dimentional cubic damped oscillator ODE.

\paragraph{Gradient descent execution time measurement.} 
For the SGD-based SINDy models, execution time is measured at each epoch for 600 epochs using mini-batches of size 100. Training was initialized with the RAdam optimizer, using a learning rate of $5 \times 10^{-3}$ and $\ell_2$ regularization ($\lambda = 10^{-2}$). Every 200 epochs, a thresholding operation is applied to the coefficient matrix $\bm{\alpha}$, zeroing entries below $5 \times 10^{-2}$ and masking subsequent updates to enforce sparsity. After each thresholding step, the learning rate is reduced by a factor of 10, and the optimizer is reinitialized with the updated learning rate. A convergence criterion based on the average change in $\alpha$ over a sliding window of 5 epochs (tolerance $10^{-6}$) is monitored but not met during training.

\begin{figure}[h]
\centering
\includegraphics[width=.8\textwidth]{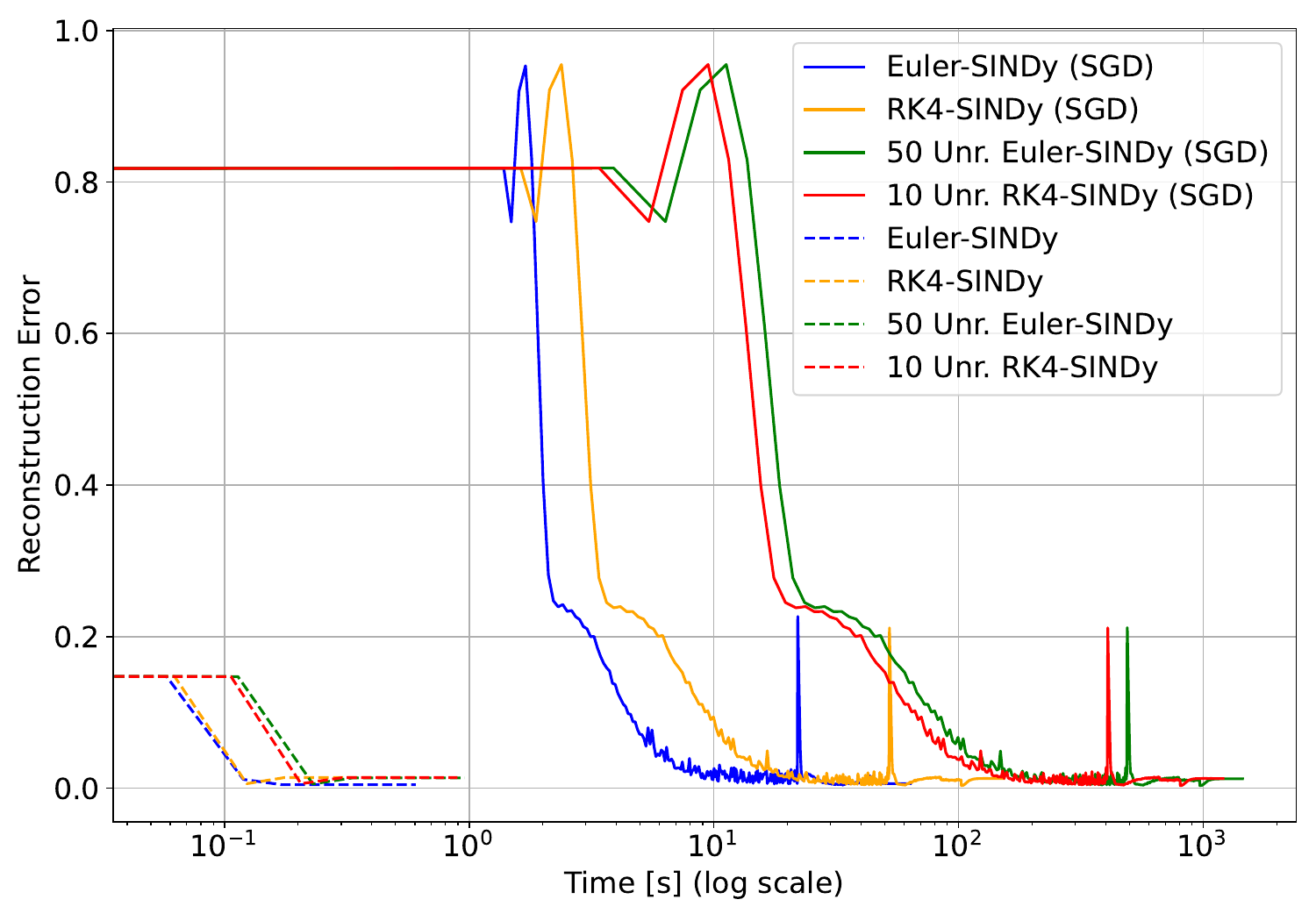}
\caption{Comparison of the reconstruction errors as a function of the execution time (log scale) for gradient descent (SGD) and closed-form SINDy models on the 2-dimentional cubic damped oscillator (where $h=10^{-3}$).} \label{fig:logscale_exectime}
\end{figure} 

We report the results in Fig.~\ref{fig:logscale_exectime} from which three remarks can be made: (i) As expected, the closed-form versions (dashed-lines) are much more efficient than the GD counterparts (solid lines), justifying the focus in the core of this paper on these parameter-free approaches; (ii) Unrolling Euler-SINDy and RK4-SINDy in closed-form is algorithmically efficient, without adding significant additional computational costs (iii) compared to the non unrolled methods, the additional computational burden imposed by the unrolling scheme is amplified when the gradient descent method is used for optimizing (solid lines);

\subsection{Cubic Damped Oscillator ODE} \label{sec:cubic-supp}

We recall the cubic damped oscillator equation as follows:
\begin{eqnarray}
\left\{\begin{array}{rcl}
\frac{d(x(t))}{dt} & = & -0.1x^3(t)+2.0y^3(t) \\
\frac{d(y(t))}{dt} & = & -2.0x^3(t)-0.1y^3(t)
\end{array}\right. \label{eq:oscillator-suppl}
\end{eqnarray}
where, $\dot{x}(t)$ and $\dot{y}(t)$ denote the time derivatives of the state variables $x(t)$ and $y(t)$, respectively.

\paragraph{Experimental setup} Using the adaptive solver {\tt solve$\_$ivp} from the library {\tt SciPy} with default settings, we simulate a  set $S$ of 50,000 data in the time domain $[0,10]$ with a time very fine step $h=2\!\cdot\!10^{-4}$ under the initial conditions $x_0=x(t=0)=-0.488$ , $y_0=y(t=0)=1.096$. $\alpha_{th}$ is set to 0.05 and $\lambda=10^{-2}$.

\paragraph{Robustness to increasing time steps and scarce data}
From the original dataset of 50000 points evenly spaced by $2 \cdot 10^{-4}$, we construct several sub-problems by systematically increasing the time step $h$ from $2 \cdot 10^{-4}$ up to $6 \cdot 10^{-1}$. As the time step grows, the number of available training pairs $(u(t), u(t+h))$ decreases, and each pair represents a larger temporal jump. This creates a more challenging scenario for regression-based identification methods: the dynamics between $u(t)$ and $u(t+h)$ are less directly observable, and a single step of standard numerical approximation may no longer be sufficient to capture the true evolution of the system.

We evaluate Euler-SINDy, RK4-SINDy, and their unrolled variants on these sub-problems, reporting the results in Table~\ref{tab:sparsity} and in Table.~\ref{tab:oscillator_sparse_l1_loss}. For compactness, we present the results associated with the unrolling value $K$ yielding the best training error as the behavior with respect to $K$ is stable as illustrated in Table.~\ref{tab:cubic_oscillator_sparse_euler_table_h_k} and Table.~\ref{tab:cubic_oscillator_sparse_rk4_table_h_k}. The numbers shown in red in the table correspond to additional terms that are absent in the true equation. This color coding was chosen to improve readability and allow the table to be interpreted at a glance.

The results illustrate the limitations of standard SINDy approaches as the time step increases. Euler-SINDy quickly fails to recover the correct equation as soon as $h$ reaches $10^{-1}$. RK4-SINDy is more robust, delaying failure until $h = 4 \cdot 10^{-1}$, but eventually also suffers from instabilities and introduces spurious terms when the number of training pairs becomes too small or the time step too large.

\begin{table}[h]
\centering
\caption{Robustness of Euler-SINDy, RK4-SINDy, and their unrolled variants on cubic damped oscillator (Eq.~\ref{eq:oscillator-suppl}), on sparser and sparser observations (higher $h$). Unrolling lowers error and solves sparser problems.}
\includegraphics[width=1\textwidth]{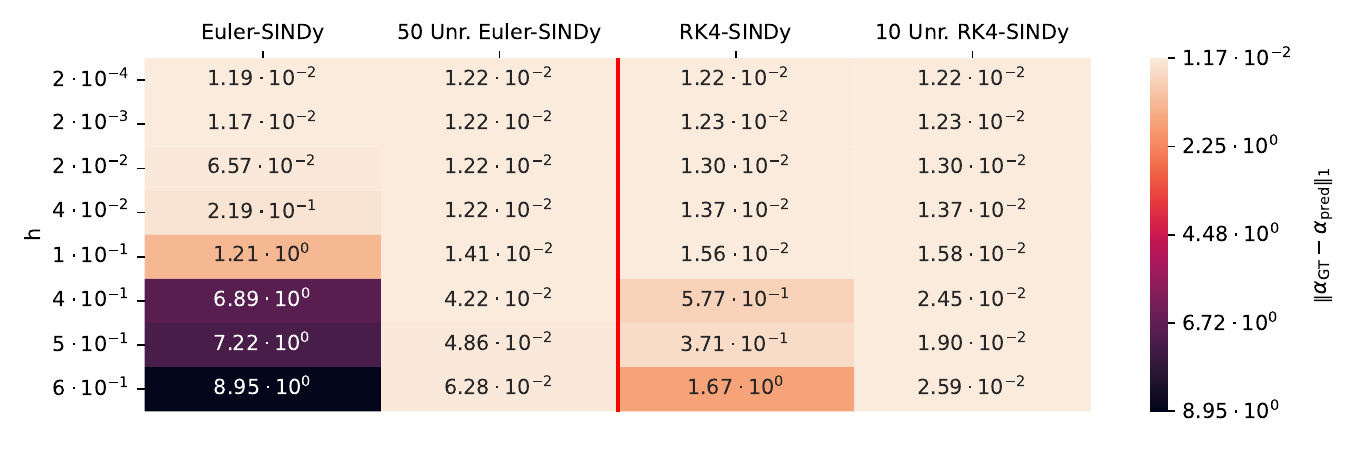}

\label{tab:oscillator_sparse_l1_loss}
\end{table}

\begin{table}[h]
\begin{scriptsize}
\caption{Robustness of Euler-SINDy, RK4-SINDy and their unrolled versions on cubic damped oscillator (Eq.~\ref{eq:oscillator-suppl}), with an increasing time step $h$ and decreasing number of learning pairs $N$. When the method fails to recover the governing equations, only the number of wrong additional terms is indicated.}\label{tab:sparsity}
\resizebox{\columnwidth}{!}{%
\addtolength{\tabcolsep}{-0.4em}
\begin{tabular}{|c|p{3.5cm}|p{3.5cm}!{\color{red}\vrule width 1.5pt}p{3.5cm}|p{3.5cm}|}
\hline
$h$ ($N$) & \multicolumn{1}{c|}{Euler-SINDy} & \multicolumn{1}{c!{\color{red}\vrule width 1.5pt}}{50 Unrolled Euler-SINDy} & \multicolumn{1}{c|}{RK4-SINDy} & \multicolumn{1}{c|}{10 Unrolled RK4-SINDy} \\
\hline
\begin{tabular}{c}$2\!\cdot\!10^{-4}$\\ ($N=5\!\cdot\!10^4$) \end{tabular} & 
\begin{tabular}{c}$ -0.098 x^3 + 1.995 y^3$ \\ $ -1.996 x^3  -0.099 y^3$ \end{tabular} & 
\begin{tabular}{c}$ -0.098 x^3 + 1.995 y^3$ \\ $ -1.996 x^3  -0.099 y^3$ \end{tabular} & 
\begin{tabular}{c}$ -0.098 x^3 + 1.995 y^3$ \\ $ -1.996 x^3  -0.099 y^3$ \end{tabular} & 
\begin{tabular}{c}$ -0.098 x^3 + 1.995 y^3$ \\ $ -1.996 x^3  -0.099 y^3$ \end{tabular} \\ \hline

\begin{tabular}{c}$2\!\cdot\!10^{-3}$\\ ($N=5\!\cdot\!10^3$) \end{tabular} & 
\begin{tabular}{c}$ -0.101 x^3 + 1.994 y^3$ \\ $ -1.995 x^3  -0.101 y^3$ \end{tabular} & 
\begin{tabular}{l}$ -0.098 x^3 + 1.995 y^3$ \\ $ -1.996 x^3  -0.099 y^3$ \end{tabular} & 
\begin{tabular}{l}$ -0.098 x^3 + 1.995 y^3$ \\ $ -1.996 x^3 -0.099 y^3$ \end{tabular} & 
\begin{tabular}{l}$ -0.098 x^3 + 1.995 y^3$ \\ $ -1.996 x^3 -0.099 y^3$ \end{tabular} \\ \hline

\begin{tabular}{c}$2\!\cdot\!10^{-2}$\\ ($N=500$) \end{tabular} &
\begin{tabular}{l}$ -0.124 x^3 + 1.986 y^3$ \\ $ -1.993 x^3  -0.121 y^3$ \end{tabular} & 
\begin{tabular}{c}$ -0.098 x^3 + 1.995 y^3$ \\ $ -1.995 x^3 -0.099 y^3$ \end{tabular} & 
\begin{tabular}{l}$ -0.098 x^3 + 1.995 y^3$ \\ $ -1.995 x^3  -0.099 y^3$ \end{tabular} & 
\begin{tabular}{l}$ -0.098 x^3 + 1.995 y^3$ \\ $ -1.995 x^3  -0.099 y^3$ \end{tabular} \\ \hline

\begin{tabular}{c}$4\!\cdot\!10^{-2}$\\ ($N=249$) \end{tabular} &
\begin{tabular}{c}$-0.124 x^3 + 1.985 y^3$ \plusterms{1} \\ $ -1.983 x^3  -0.122 y^3$ \plusterms{1} \end{tabular} & 
\begin{tabular}{c}$ -0.099 x^3 + 1.995 y^3$ \\ $ -1.995 x^3 -0.100 y^3$ \end{tabular} & 
\begin{tabular}{c}$ -0.098 x^3 + 1.995 y^3$ \\ $ -1.995 x^3 -0.099 y^3$ \end{tabular} & 
\begin{tabular}{c}$ -0.098 x^3 + 1.995 y^3$ \\ $ -1.995 x^3 -0.099 y^3$ \end{tabular} \\ \hline

\begin{tabular}{c}$1\!\cdot\!10^{-1}$\\ ($N=100$) \end{tabular} &
\begin{tabular}{c}$  -0.252 x^3 + 1.900 y^3$ \plusterms{3} \\ $ -1.923 x^3 -0.230 y^3$ \plusterms{2} \end{tabular} & 
\begin{tabular}{c}$ -0.100 x^3 + 1.993 y^3$ \\ $ -1.994 x^3  -0.101 y^3$ \end{tabular} & 
\begin{tabular}{c}$ -0.098 x^3 + 1.994 y^3$ \\ $ -1.994 x^3 -0.099 y^3$ \end{tabular} & 
\begin{tabular}{l}$ -0.098 x^3 + 1.994 y^3$ \\ $ -1.994 x^3 -0.099 y^3$ \end{tabular} \\ \hline

\begin{tabular}{c}$4\!\cdot\!10^{-1}$\\ ($N=24$) \end{tabular} &
\begin{tabular}{c}$ -0.388 x^3 + 1.313 y^3$ \plusterms{9} \\ $ -1.420 x^3   -0.595 y^3$ \plusterms{7} \end{tabular} & 
\begin{tabular}{c}$ -0.108 x^3 + 1.989 y^3$ \\ $ -1.984 x^3  -0.107 y^3$ \end{tabular} & 
\begin{tabular}{c}$ -0.099 x^3 + 1.987 y^3$ \plusterms{3} \\ $ -1.997 x^3  -0.108 y^3$ \plusterms{1} \end{tabular} & 
\begin{tabular}{l}$ -0.098 x^3 + 1.993 y^3$ \\ $ -1.985 x^3 -0.100 y^3$ \end{tabular} \\ \hline

\begin{tabular}{c}$5\!\cdot\!10^{-1}$\\ ($N=19$) \end{tabular} &
\begin{tabular}{c}$ -0.525 x^3 + 1.260 y^3$ \plusterms{9} \\ $  -1.446 x^3 -0.561 y^3$ \plusterms{9} \end{tabular} & 
\begin{tabular}{l}$ -0.114 x^3 + 1.990 y^3$ \\ $ -1.985 x^3  -0.110 y^3$ \end{tabular} & 
\begin{tabular}{c}$ -0.119 x^3 + 1.961 y^3$ \plusterms{1} \\ $ -1.942 x^3  -0.113 y^3$ \plusterms{2} \end{tabular} & 
\begin{tabular}{c}$ -0.100 x^3 + 1.994 y^3$ \\ $ -1.989 x^3 -0.101 y^3$ \end{tabular} \\ \hline

\begin{tabular}{c}$6\!\cdot\!10^{-1}$\\ ($N=16$) \end{tabular} &
\begin{tabular}{c}$  -0.681 x^3 + 1.067 y^3$ \plusterms{11} \\ $  -1.330 x^3 -0.461 y^3$ \plusterms{11} \end{tabular} & 
\begin{tabular}{l}$ -0.114 x^3 + 1.989 y^3$ \\ $ -1.978 x^3  -0.116 y^3$ \end{tabular} & 
\begin{tabular}{c}$ -0.123 x^3 + 1.952 y^3$ \plusterms{4} \\ $ -1.872 x^3 -0.169 y^3$ \plusterms{4} \end{tabular} & 
\begin{tabular}{l}$ -0.100 x^3 + 1.989 y^3$ \\ $ -1.988 x^3 -0.103 y^3$ \end{tabular} \\ \hline
\end{tabular}
}
\end{scriptsize}
\end{table}

In contrast, both Unrolled Euler-SINDy and Unrolled RK4-SINDy exhibit remarkable robustness across all time steps. By subdividing each integration step into $K$ smaller sub-steps of size $h/K$, these methods effectively reduce local truncation errors, yielding more accurate approximations of the derivative even for widely spaced observations. This allows the unrolled methods to recover the underlying equations almost perfectly, even from as few as 16 training pairs when $h = 6 \cdot 10^{-1}$. The unrolled scheme compensates for the sparsity of data by effectively “filling in” the missing temporal information between successive observations, which standard Euler or RK4 approaches cannot do.

\begin{figure}[h]
\centering
\begin{subfigure}{0.5\textwidth}
    \includegraphics[width=\linewidth]{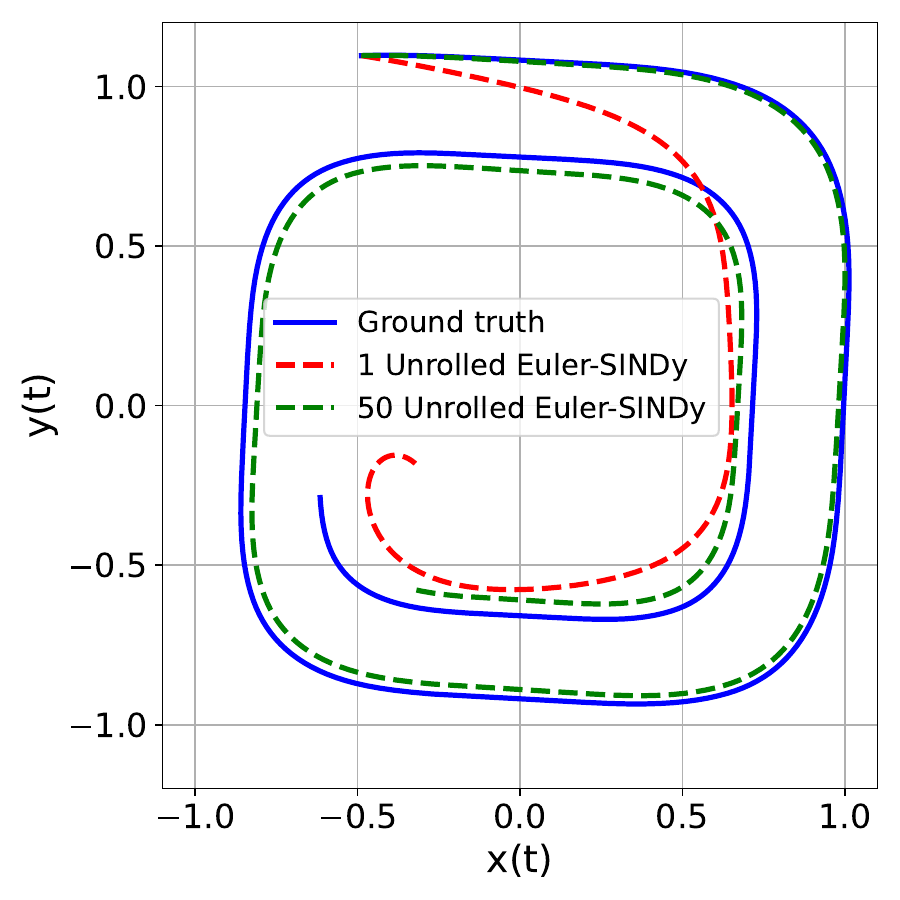}
    \caption{}
\end{subfigure}\hfill
\begin{subfigure}{0.5\textwidth}
    \includegraphics[width=\linewidth]{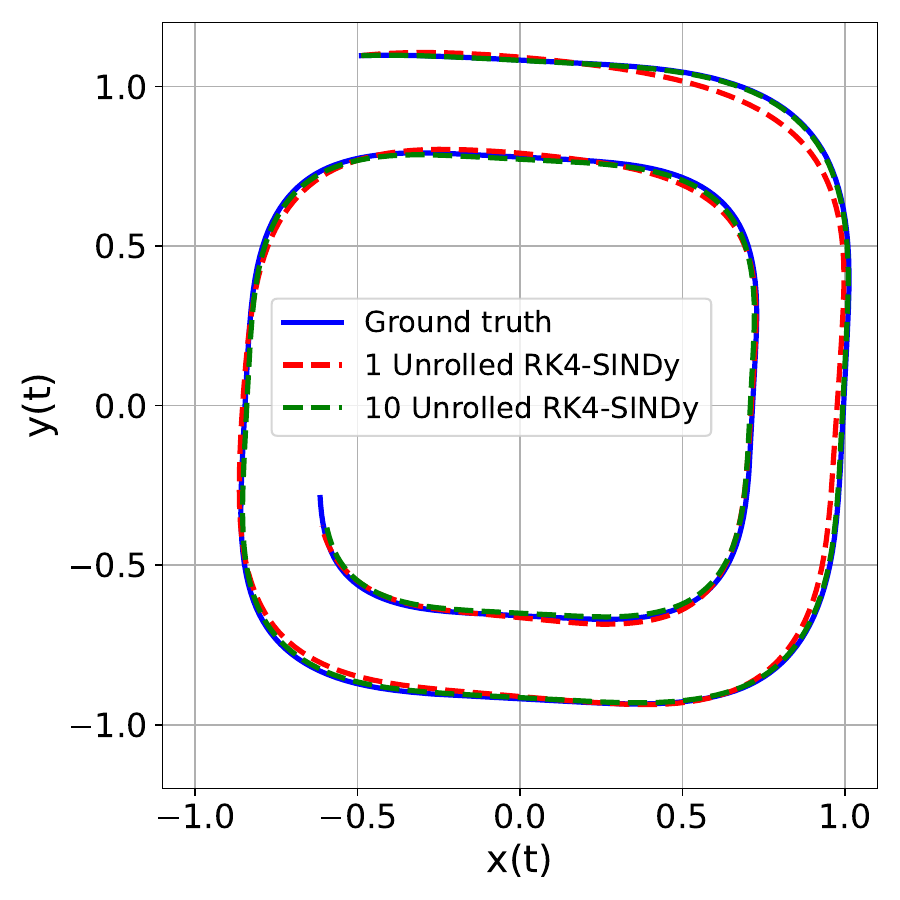}
    \caption{}
\end{subfigure}
\caption{Comparison of the cubic damped oscillator solutions (Eq.~\eqref{eq:oscillator-suppl}) with trajectories obtained from the learned ODEs using (a) Euler-SINDy and (b) RK4-SINDy, including their unrolled variants, at $h=0.6$.}

\label{fig:oscillator-unrolled-sindy}
\end{figure}

Fig.~\ref{fig:oscillator-unrolled-sindy}.a and Fig.~\ref{fig:oscillator-unrolled-sindy}.b visually illustrate this effect by comparing the equations learned by Euler-SINDy, RK4-SINDy, and their unrolled variants to the ground truth defined in Eq.~\eqref{eq:oscillator-suppl}. The differences are striking: Euler-SINDy fails to capture the cubic interactions entirely at large time steps, producing spurious coefficients or missing key terms. RK4-SINDy performs better, recovering some of the terms correctly, but still introduces multiple incorrect terms, demonstrating that higher-order integration alone cannot fully overcome the issues caused by sparse observations. By contrast, the unrolled variants maintain high fidelity to the true equations, clearly showing that unrolling is a highly effective strategy for mitigating both the loss of information due to large time steps and the reduction in the number of available training pairs.

\begin{table}[h]
\centering
\caption{Accuracy of $K$-Unrolled Euler-SINDy with different unrolling depths $K$ and observation step sizes $h$ for the cubic damped oscillator (Eq.~\ref{eq:oscillator-suppl}). Lighter colors indicate more accurate recovery.}
\includegraphics[width=1\textwidth]{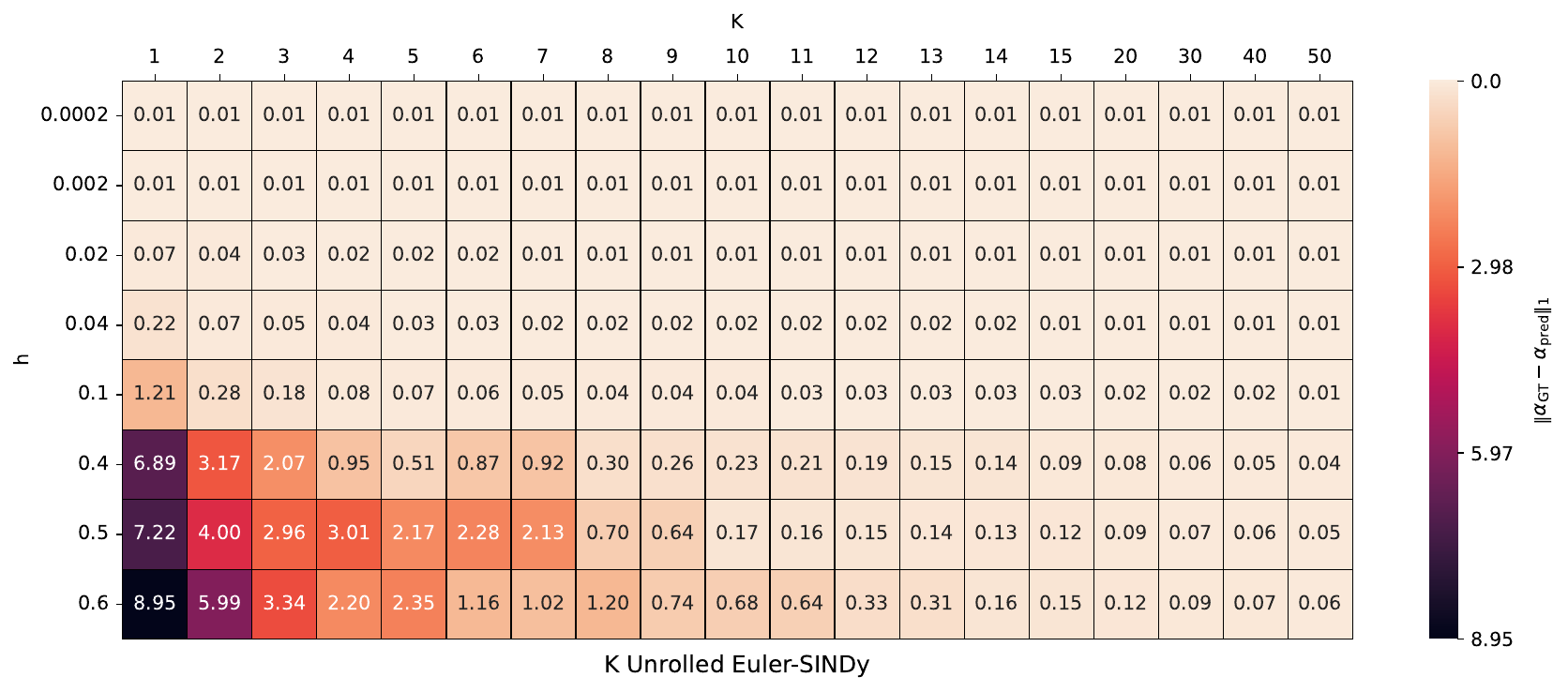}

\label{tab:cubic_oscillator_sparse_euler_table_h_k}
\end{table}

\begin{table}[h]
\centering
\caption{Accuracy of $K$-Unrolled RK4-SINDy with different unrolling depths $K$ and observation step sizes $h$ for cubic damped oscillator (Eq.~\ref{eq:oscillator-suppl}).} 
\includegraphics[width=\textwidth]{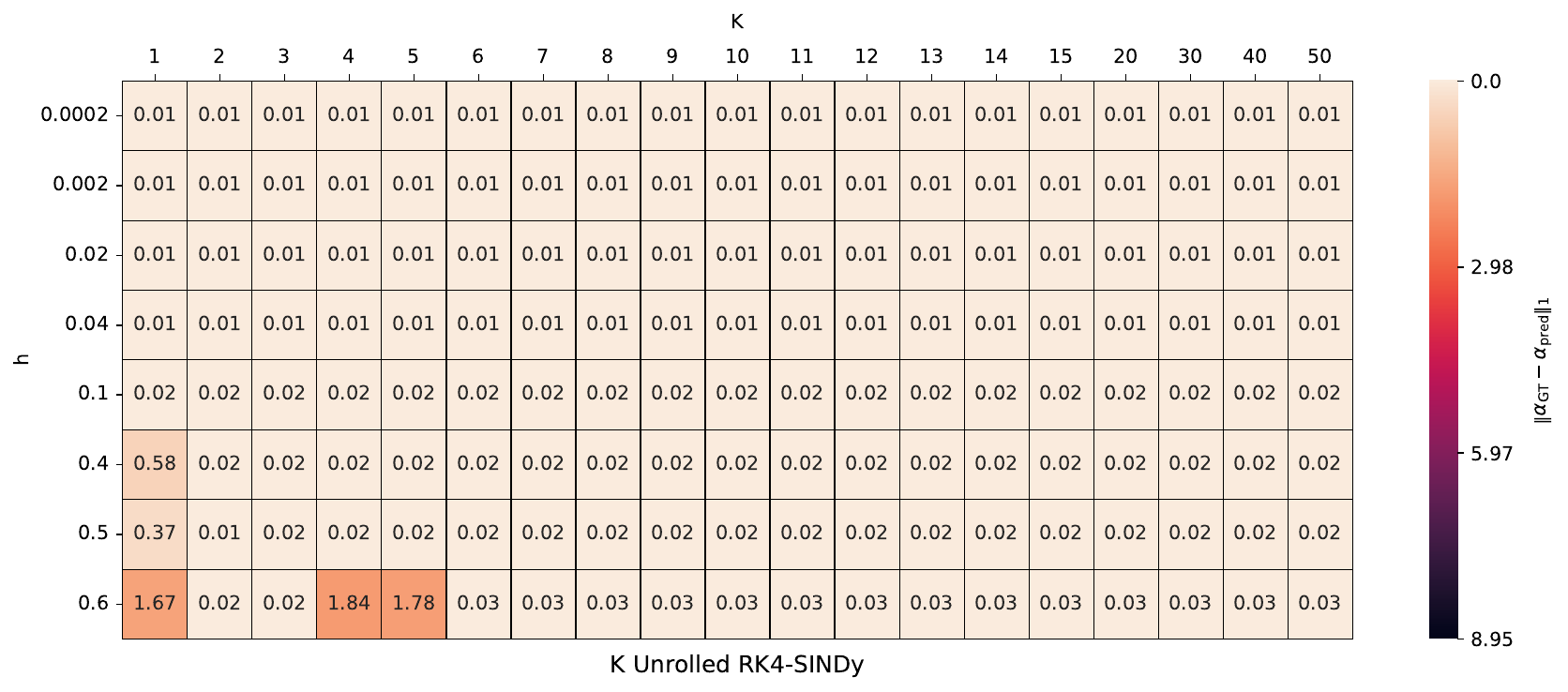}
\label{tab:cubic_oscillator_sparse_rk4_table_h_k}
\end{table} 

\paragraph{Robustness to increasing time steps and constant data}
We have seen in Table~\ref{tab:sparsity} and in Table.~\ref{tab:oscillator_sparse_l1_loss} that Euler-SINDy performs very poorly as $h$ increases. However, in that earlier experiment, the number of training pairs $N$ also decreased as $h$ increased, introducing an additional confounding factor: inability to recover the equation could come not only from integration errors but also from a lack of training data. To isolate the effect of increasing $h$ on equation recovery, we keep the number of training pairs $N$ constant, ensuring that any degradation is solely due to larger time steps rather than a reduction in training data.

Both Euler-SINDy and Unrolled Euler-SINDy results are shown in Table~\ref{tab:cubic-regular} and Table.~\ref{tab:cubic_oscillator_regular_euler_table}, where the results associated with the unrolling value $K$ yielding the best training error as the behavior with respect to $K$ is stable are illustrated in Table.~\ref{tab:cubic_oscillator_regular_euler_table_h_k}.

Euler-SINDy deteriorates rapidly as $h$ increases. In particular, for $h=5 \cdot 10^{-2}$, it fails to recover the correct governing equations, highlighting its limited ability to handle larger time steps. In contrast, the unrolled variant remains robust for much larger $h$. For instance, 50-Unrolled Euler-SINDy successfully identifies the correct dynamics even where standard Euler-SINDy fails. This improved performance arises because unrolling effectively divides each step into smaller internal updates, allowing the model to capture the dynamics accurately while still training on data sampled with a large $h$.

\begin{table}[h]
\centering
\caption{Robustness of Euler-SINDy and its unrolled version cubic damped oscillator (Eq.~\ref{eq:oscillator-suppl}) evaluated with increasing time step $h$ and constant number of training pairs $N$ .} 
\includegraphics[width=\textwidth]{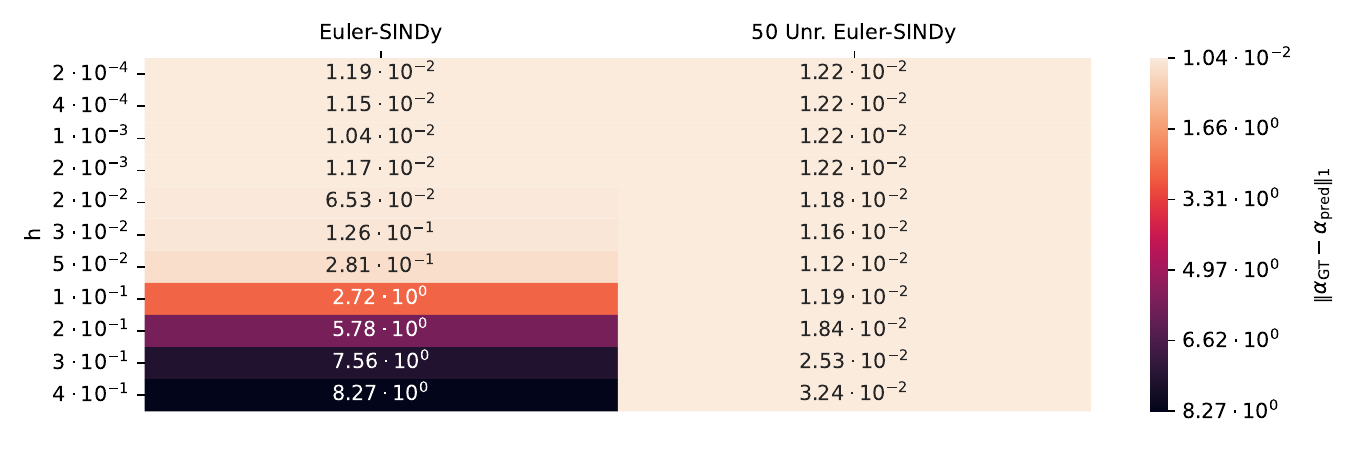}
\label{tab:cubic_oscillator_regular_euler_table}
\end{table}

\begin{table*}[h]
\center{ 
\caption{Robustness of Euler-SINDy and its unrolled version on Eq.~\ref{eq:oscillator-suppl}, with an increasing time step $h$ and a \textbf{constant number of learning pairs $N=50000$}. When Euler-SINDy fails to recover the governing equations, only the number of wrong additional terms is indicated.}
\label{tab:cubic-regular}
\begin{small}
\raggedright
\begin{tabular}{|c|c|c|}
\hline
$h$ & \textbf{Euler-SINDy} & \textbf{50 Unrolled Euler-SINDy} \\
\hline
\begin{tabular}{c}$h=2e-04$\\\end{tabular} & \begin{tabular}{l}$ -0.098 x^3 + 1.995 y^3$ \\ $ -1.996 x^3  -0.099 y^3$\end{tabular} & \begin{tabular}{l}$ -0.098 x^3 + 1.995 y^3$ \\ $ -1.996 x^3  -0.099 y^3$\end{tabular} \\ \hline
\begin{tabular}{c}$h=4e-04$\\\end{tabular} & \begin{tabular}{l}$ -0.098 x^3 + 1.995 y^3$ \\ $ -1.996 x^3  -0.099 y^3$\end{tabular} & \begin{tabular}{l}$ -0.098 x^3 + 1.995 y^3$ \\ $ -1.996 x^3  -0.099 y^3$\end{tabular} \\ \hline
\begin{tabular}{c}$h=1e-03$\\\end{tabular} & \begin{tabular}{l}$ -0.099 x^3 + 1.995 y^3$ \\ $ -1.996 x^3  -0.100 y^3$\end{tabular} & \begin{tabular}{l}$ -0.098 x^3 + 1.995 y^3$ \\ $ -1.996 x^3  -0.099 y^3$\end{tabular}  \\ \hline
\begin{tabular}{c}$h=2e-03$\\\end{tabular} & \begin{tabular}{l}$ -0.101 x^3 + 1.994 y^3$ \\ $ -1.995 x^3  -0.101 y^3$\end{tabular} & \begin{tabular}{l}$ -0.098 x^3 + 1.995 y^3$ \\ $ -1.996 x^3  -0.099 y^3$\end{tabular} \\ \hline
\begin{tabular}{c}$h=2e-02$\\\end{tabular} & \begin{tabular}{l}$ -0.124 x^3 + 1.986 y^3$ \\ $ -1.993 x^3  -0.121 y^3$\end{tabular} & \begin{tabular}{l}$ -0.098 x^3 + 1.995 y^3$ \\ $ -1.996 x^3 -0.099 y^3$\end{tabular} \\ \hline
\begin{tabular}{c}$h=3e-02$\\\end{tabular} & \begin{tabular}{l}$ -0.135 x^3 + 1.989 y^3$ \\ $ -1.988 x^3  -0.117 y^3 \plusterms{1} \hspace*{-3mm}$\end{tabular} & \begin{tabular}{l}$ -0.099 x^3 + 1.995 y^3$ \\ $ -1.995 x^3  -0.099 y^3$\end{tabular} \\ \hline
\begin{tabular}{c}$h=5e-02$\\\end{tabular} & \begin{tabular}{l}$ -0.129 x^3 + 1.981 y^3 \plusterms{1} \hspace*{-3mm}$ \\ $ -1.979 x^3  -0.126 y^3 \plusterms{1} \hspace*{-3mm}$\end{tabular} & \begin{tabular}{l}$ -0.099 x^3 + 1.995 y^3$ \\ $ -1.995 x^3  -0.100 y^3$\end{tabular} \\ \hline
\begin{tabular}{c}$h=1e-01$\\\end{tabular} & \begin{tabular}{l}$ -0.247 x^3 + 1.948 y^3 \plusterms{8} \hspace*{-3mm}$ \\ $ -1.983 x^3  -0.223 y^3 \plusterms{10} \hspace*{-3mm}$\end{tabular} &  \begin{tabular}{l}$ -0.100 x^3 + 1.994 y^3$ \\ $ -1.995 x^3  -0.101 y^3$\end{tabular} \\ \hline
\begin{tabular}{c}$h=2e-01$\\\end{tabular} & \begin{tabular}{l}$ -0.371 x^3 + 1.834 y^3 \plusterms{11} \hspace*{-3mm}$ \\ $ -1.913 x^3  -0.350 y^3 \plusterms{12} \hspace*{-3mm}$\end{tabular} &\begin{tabular}{l}$ -0.103 x^3 + 1.993 y^3$ \\ $ -1.994 x^3  -0.103 y^3$\end{tabular} \\ \hline
\begin{tabular}{c}$h=3e-01$\\\end{tabular} & \begin{tabular}{l}$ -0.465 x^3 + 1.648 y^3 \plusterms{11} \hspace*{-3mm}$ \\ $ -1.786 x^3  -0.478 y^3 \plusterms{12} \hspace*{-3mm}$\end{tabular} &  \begin{tabular}{l}$ -0.105 x^3 + 1.991 y^3$ \\ $ -1.993 x^3  -0.105 y^3$\end{tabular} \\ \hline
\begin{tabular}{c}$h=4e-01$\\\end{tabular} & \begin{tabular}{l}$ -0.525 x^3 + 1.406 y^3 \plusterms{9} \hspace*{-3mm}$ \\ $ -1.601 x^3  -0.598 y^3 \plusterms{13} \hspace*{-3mm}$\end{tabular} &  \begin{tabular}{l}$ -0.108 x^3 + 1.990 y^3$ \\ $ -1.992 x^3  -0.107 y^3$\end{tabular}  \\ \hline
\end{tabular}
\end{small}}
\end{table*}

\begin{table}[t]
\centering
\caption{Accuracy of $K$-Unrolled Euler-SINDy with different unrolling depths $K$ and observation step sizes $h$ for the cubic damped oscillator (Eq.~\ref{eq:oscillator-suppl}). While value $h$ increases \textbf{the number of training pairs $N$ is kept constant}.} 
\includegraphics[width=\textwidth]{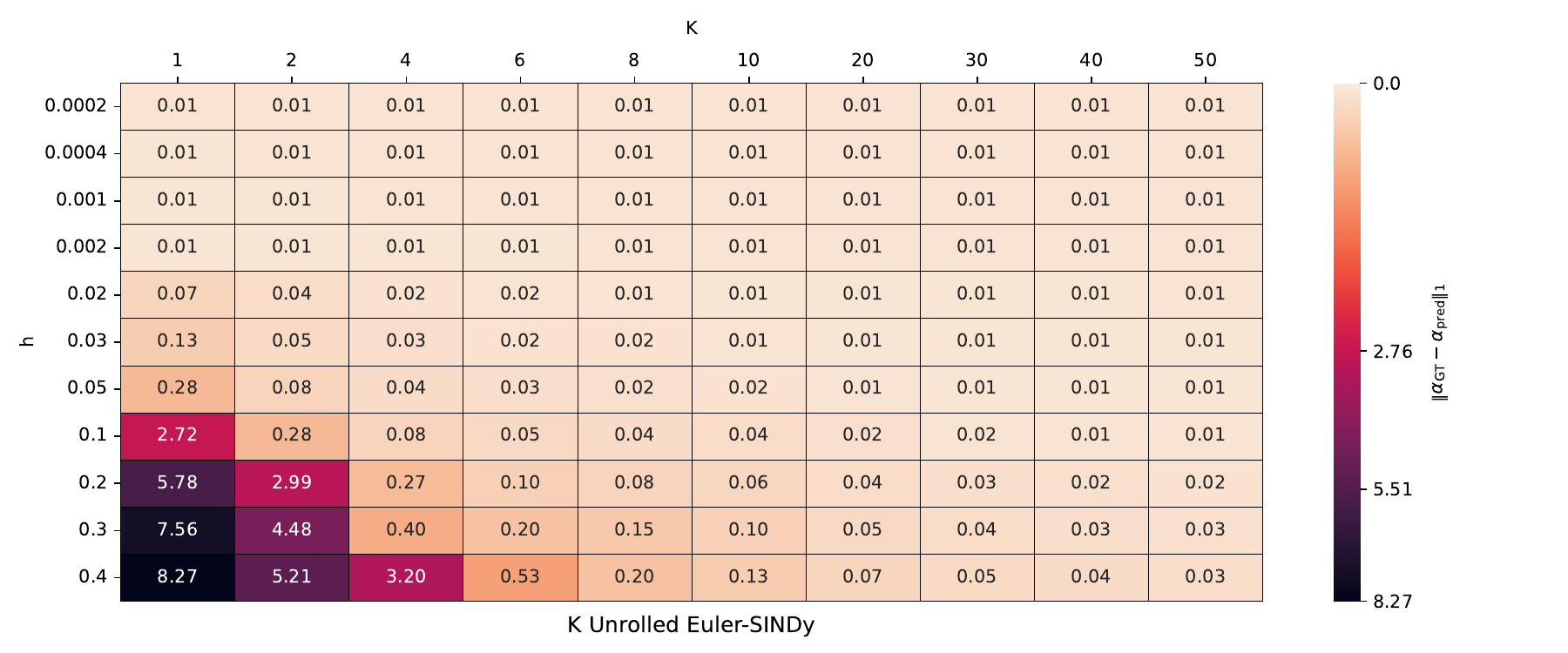}
\label{tab:cubic_oscillator_regular_euler_table_h_k}
\end{table}

\subsection{Advection} \label{sec:expes-Advection}

Let us remind that this PDE describes the motion of $u$ as it is advected by a  velocity field as follows: 
\begin{eqnarray}
    u_{t}(x,t) = -0.4 u_{x}(x,t), \label{eq:1Dadvection}
\end{eqnarray}
where $u_{t}$ (resp. $u_{x}$) denotes the partial derivative w.r.t $t$ (resp. $x$). 

\paragraph{Experimental setup} 
Using {\tt pde-bench}, we simulate  100,000 pairs with  $t \in [0,2]$ and $\Omega = [0,1]$ with $h=5\!\cdot\!10^{-4}$
under the initial conditions $sin(x,t=0)=sin(2 \pi\cdot x)$. The sparsity threshold $\alpha_{th}=0.01$. The regularization parameter $\lambda=10^{-2}$. 

\paragraph{Results}
From the original evenly spaced dataset, we construct sub-problems by progressively increasing the time step $h$. As $h$ grows, the number of available training pairs decreases, and each pair corresponds to a larger temporal jump. For each setting, we report results with the unrolling value $K$ that yields the best training error, since the behavior with respect to $K$ remains stable (see Tables~\ref{tab:advection_sparse_euler_table_h_k} and \ref{tab:advection_sparse_rk4_table_h_k}). The results for Euler-SINDy are given in Tables~\ref{tab:advection-Euler} and \ref{tab:advection_euler_l1_loss}. They show that the method diverges once $h$ reaches $4\cdot 10^{-3}$, whereas its unrolled counterpart ($K=25$) extends this stability limit up to $4\cdot 10^{-2}$. RK4-SINDy performs better overall (Tables~\ref{tab:advection RK4} and \ref{tab:advection_rk4_l1_loss}), but fails to recover the PDE beyond $h=10^{-1}$. In contrast, 25-Unrolled RK4-SINDy remains accurate even for $h=0.15$.

\begin{figure}[t]
\centering
\includegraphics[width=\textwidth]{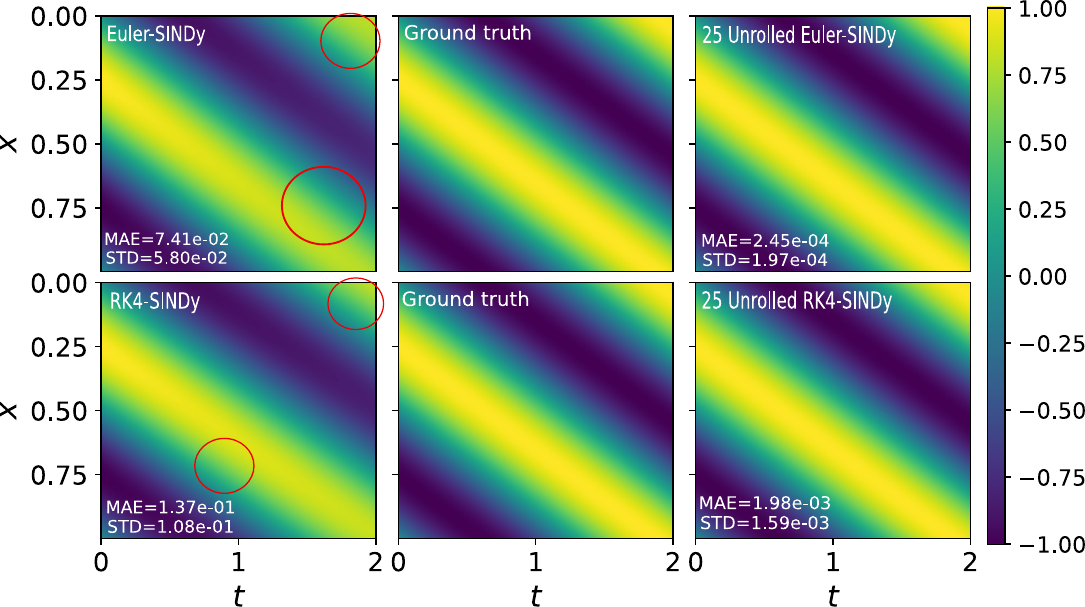}
\caption{Solutions of the Advection PDE for different time steps. A time step of $h=0.04$ is used for Euler-based methods and $h=0.15$ for RK4-based methods. The first row (with $h=0.04$) shows Euler-SINDy, the Ground Truth (GT), and 25 Unrolled Euler-SINDy. The second row (with $h=0.15$) shows RK4-SINDy, GT, and 25 Unrolled RK4-SINDy. For each learned equation, the corresponding Mean Absolute Error (MAE) is indicated.
} 
\label{fig:advection_plot}
\end{figure}

The performance thresholds at which Euler-SINDy and RK4-SINDy saturate, compared with the continued robustness of their unrolled variants, are clearly illustrated in Fig.~\ref{fig:advection_plot}. At first sight, the visual comparison of the reconstructed solutions might suggest that all methods perform similarly. However, closer inspection reveals significant qualitative differences: non-unrolled models produce visibly blurred structures in some regions (highlighted by red-circled areas in Fig.~\ref{fig:advection_plot}). These blurred regions are indicative of accumulated numerical dispersion and instability, typical when the time step is too large for the scheme to handle reliably. 

\begin{table}[t]
\centering
\caption{Robustness of Euler-SINDy and its unrolled version on advection equation (Eq.~\ref{eq:1Dadvection}), with an increasing time step $h$ and a decreasing number of learning pairs.}
\includegraphics[width=\textwidth]{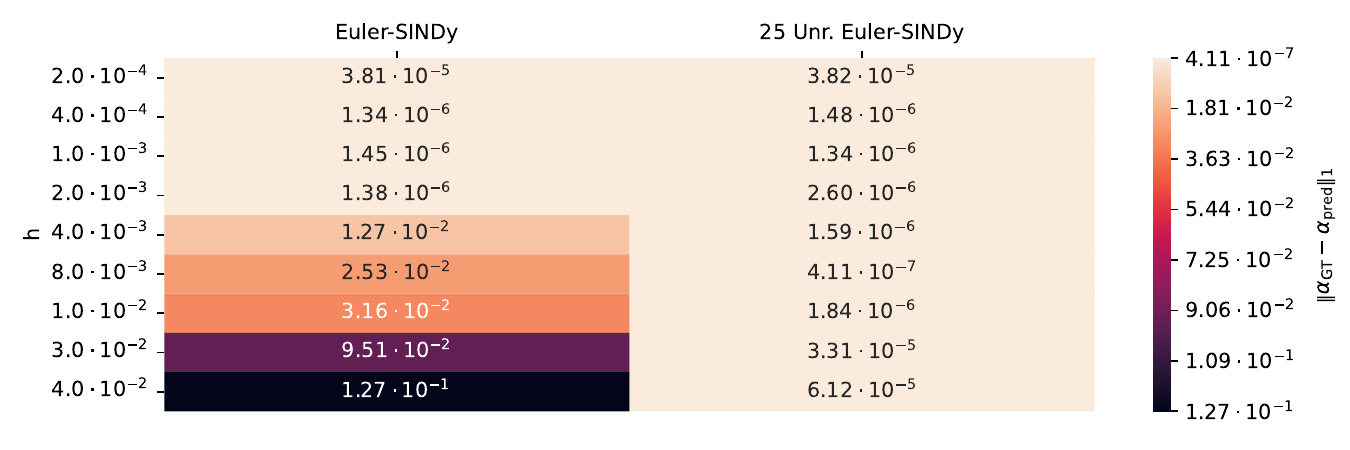}
\label{tab:advection_euler_l1_loss}
\end{table}

\begin{table}[t]
\centering
\caption{Robustness of RK4-SINDy and its unrolled version on advection equation (Eq.~\ref{eq:1Dadvection}), with an increasing time step $h$ and a decreasing number of learning pairs.}
\includegraphics[width=\textwidth]{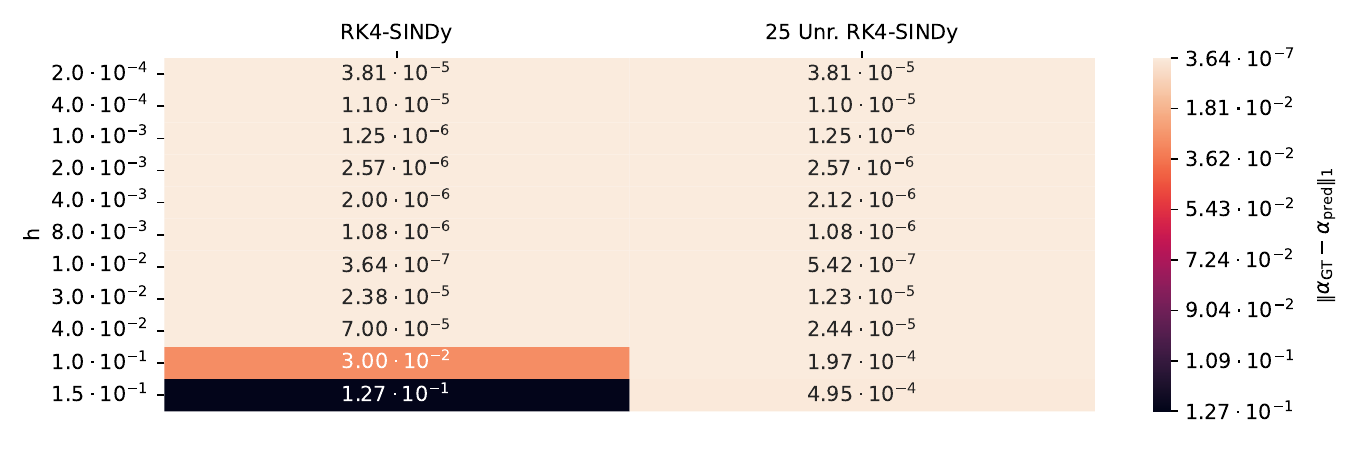}
\label{tab:advection_rk4_l1_loss}
\end{table}

\begin{table*}[t]
\caption{Robustness of Euler-SINDy and its unrolled version on Eq.~\ref{eq:1Dadvection}, with an increasing time step $h$ and a decreasing number of learning pairs. When Euler-SINDy fails to recover the governing equations, only the number of wrong additional terms is indicated.}
\label{tab:advection-Euler}
\vspace*{\fill}
\centering
\scalebox{0.8}{%
\begin{tabular}{|c|c|c|}
\hline
$h(N)$ & \textbf{Euler-SINDy} & \textbf{25 Unrolled Euler-SINDy} \\
\hline
\begin{tabular}{c}$h=2e-04$\\$(N=10000)$\end{tabular} & \begin{tabular}{l}$-0.400 u_x$\end{tabular} & \begin{tabular}{l}$-0.400 u_x$\end{tabular} \\ \hline
\begin{tabular}{c}$h=4e-04$\\$(N=5000)$\end{tabular} & \begin{tabular}{l}$-0.400 u_x$\end{tabular} & \begin{tabular}{l}$-0.400 u_x$\end{tabular} \\ \hline
\begin{tabular}{c}$h=1e-03$\\$(N=2000)$\end{tabular} & \begin{tabular}{l}$-0.400 u_x$\end{tabular} & \begin{tabular}{l}$-0.400 u_x$\end{tabular} \\ \hline
\begin{tabular}{c}$h=2e-03$\\$(N=1000)$\end{tabular} & \begin{tabular}{l}$-0.400 u_x$\end{tabular} & \begin{tabular}{l}$-0.400 u_x$\end{tabular} \\ \hline
\begin{tabular}{c}$h=4e-03$\\$(N=500)$\end{tabular} & \begin{tabular}{l}$-0.400 u_x \plusterms{1} \hspace*{-3mm}$\end{tabular} & \begin{tabular}{l}$-0.400 u_x$\end{tabular} \\ \hline
\begin{tabular}{c}$h=8e-03$\\$(N=250)$\end{tabular} & \begin{tabular}{l}$-0.400 u_x \plusterms{1} \hspace*{-3mm}$\end{tabular} & \begin{tabular}{l}$-0.400 u_x$\end{tabular} \\ \hline
\begin{tabular}{c}$h=1e-02$\\$(N=200)$\end{tabular} & \begin{tabular}{l}$-0.400 u_x \plusterms{1} \hspace*{-3mm}$\end{tabular} & \begin{tabular}{l}$-0.400 u_x$\end{tabular} \\ \hline
\begin{tabular}{c}$h=3e-02$\\$(N=67)$\end{tabular} & \begin{tabular}{l}$-0.400 u_x \plusterms{1} \hspace*{-3mm}$\end{tabular} & \begin{tabular}{l}$-0.400 u_x$\end{tabular} \\ \hline
\begin{tabular}{c}$h=4e-02$\\$(N=50)$\end{tabular} & \begin{tabular}{l}$-0.399 u_x \plusterms{1} \hspace*{-3mm}$\end{tabular} & \begin{tabular}{l}$-0.400 u_x$\end{tabular} \\ \hline
\end{tabular}
}
\vspace*{\fill}
\end{table*}

\vspace{1cm}
\begin{table*}[t]
\center{ 
\begin{small}
\caption{Robustness of RK4-SINDy and its unrolled version on Eq.~\ref{eq:1Dadvection}, with an increasing time step $h$ and a decreasing number of learning pairs. When RK4-SINDy fails to recover the governing equations, only the number of wrong additional terms is indicated.}
\label{tab:advection RK4}
\raggedright
\begin{tabular}{|c|c|c|}
\hline
$h(N)$ & \textbf{RK4-SINDy} & \textbf{25 Unrolled RK4-SINDy} \\
\hline
\begin{tabular}{c}$h=2e-04$\\$(N=10000)$\end{tabular} & \begin{tabular}{l}$-0.400 u_x$\end{tabular} & \begin{tabular}{l}$-0.400 u_x$\end{tabular} \\ \hline
\begin{tabular}{c}$h=4e-04$\\$(N=5000)$\end{tabular} & \begin{tabular}{l}$-0.400 u_x$\end{tabular} & \begin{tabular}{l}$-0.400 u_x$\end{tabular} \\ \hline
\begin{tabular}{c}$h=1e-03$\\$(N=2000)$\end{tabular} & \begin{tabular}{l}$-0.400 u_x$\end{tabular} & \begin{tabular}{l}$-0.400 u_x$\end{tabular} \\ \hline
\begin{tabular}{c}$h=2e-03$\\$(N=1000)$\end{tabular} & \begin{tabular}{l}$-0.400 u_x$\end{tabular} & \begin{tabular}{l}$-0.400 u_x$\end{tabular} \\ \hline
\begin{tabular}{c}$h=4e-03$\\$(N=500)$\end{tabular} & \begin{tabular}{l}$-0.400 u_x$\end{tabular} & \begin{tabular}{l}$-0.400 u_x$\end{tabular} \\ \hline
\begin{tabular}{c}$h=8e-03$\\$(N=250)$\end{tabular} & \begin{tabular}{l}$-0.400 u_x$\end{tabular} & \begin{tabular}{l}$-0.400 u_x$\end{tabular} \\ \hline
\begin{tabular}{c}$h=1e-02$\\$(N=200)$\end{tabular} & \begin{tabular}{l}$-0.400 u_x$\end{tabular} & \begin{tabular}{l}$-0.400 u_x$\end{tabular} \\ \hline
\begin{tabular}{c}$h=3e-02$\\$(N=67)$\end{tabular} & \begin{tabular}{l}$-0.400 u_x$\end{tabular} & \begin{tabular}{l}$-0.400 u_x$\end{tabular} \\ \hline
\begin{tabular}{c}$h=4e-02$\\$(N=50)$\end{tabular} & \begin{tabular}{l}$-0.400 u_x$\end{tabular} & \begin{tabular}{l}$-0.400 u_x$\end{tabular} \\ \hline
\begin{tabular}{c}$h=1e-01$\\$(N=20)$\end{tabular} & \begin{tabular}{l}$-0.395 u_x \plusterms{1} \hspace*{-3mm}$\end{tabular} & \begin{tabular}{l}$-0.400 u_x$\end{tabular} \\ \hline
\begin{tabular}{c}$h=1.5e-01$\\$(N=14)$\end{tabular} & \begin{tabular}{l}$-0.367 u_x \plusterms{1} \hspace*{-3mm}$\end{tabular} & \begin{tabular}{l}$-0.400 u_x$\end{tabular} \\ \hline
\end{tabular}
\end{small}}
\end{table*}

\begin{table}[t]
\centering
\caption{Accuracy of $K$-Unrolled Euler-SINDy with different unrolling depths $K$ and observation step sizes $h$ for the advection equation (Eq.~\ref{eq:1Dadvection}).}
\includegraphics[width=\textwidth]{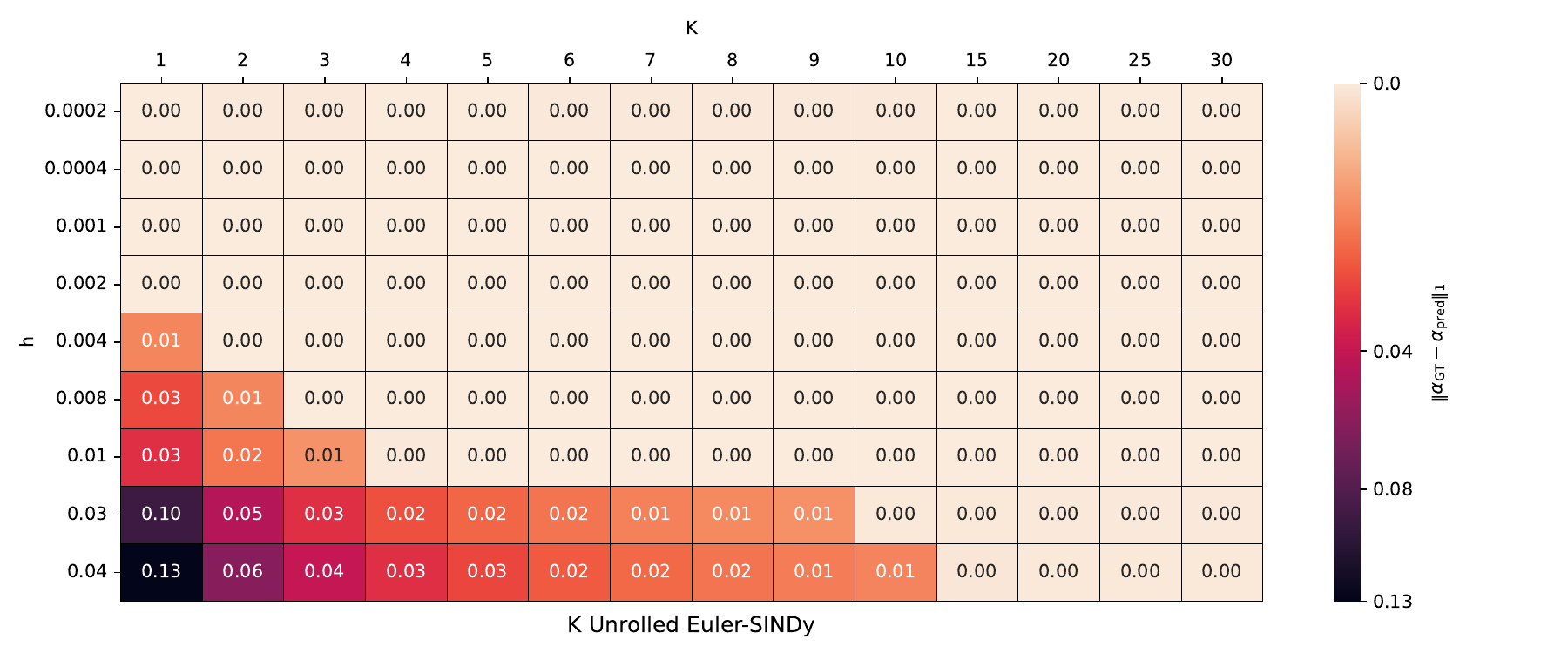}
 \label{tab:advection_sparse_euler_table_h_k}
\end{table}

\begin{table}[t]
\centering
\caption{Accuracy of $K$-Unrolled RK4-SINDy with different unrolling depths $K$ and observation step sizes $h$ for the advection equation (Eq.~\ref{eq:1Dadvection}). The marker $X$ denotes entries where the results are NaN.}
\includegraphics[width=\textwidth]{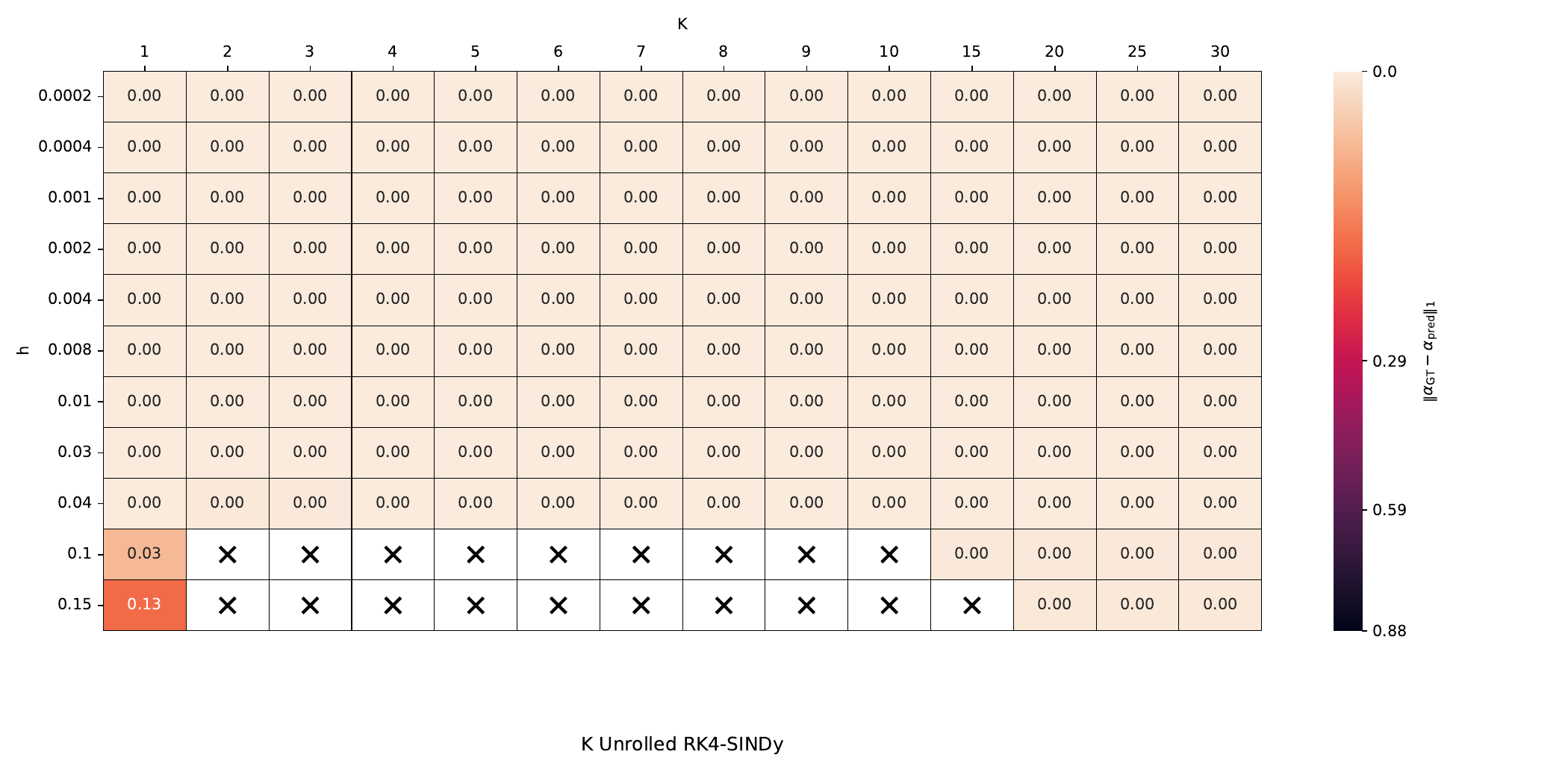}

\label{tab:advection_sparse_rk4_table_h_k}
\end{table}

\subsection{2D Reaction-Diffusion PDEs} \label{sec:expes-2Dreacdiffus}

We recall the reaction–diffusion PDEs model physical phenomena such as the change in space and time of the concentration chemical substances. The dynamics is as follows:
\begin{eqnarray}
\left\{\begin{array}{rcl}
    u_t & = & u - u^3 + v^3 + 0.1 u_{xx} + 0.1 u_{yy} + u^2 v - u v^2\\
    v_t & = & v - u^3 - v^3 + 0.1 v_{xx} + 0.1 v_{yy} - u^2 v - u v^2 
    \end{array}\right.\label{eq:2dreacdiff-V2}
\end{eqnarray}
where $u_{xx}$ (resp. $u_{yy}$) denotes the second-order partial derivative w.r.t $x$ (resp. $y$). 

The experimental setup and results for the reaction–diffusion system are already presented in Sec.~\ref{sec:PDE}. For completeness, we provide in Tables~\ref{tab:2React-Euler} and \ref{tab:2D-react-RK4} the recovered analytical expressions obtained with Euler-SINDy, RK4-SINDy, and their unrolled variants. In addition, we report results using the unrolling value $K$ that achieves the best training error, as the performance is stable with respect to $K$ (see Tables~\ref{tab:reacdiff_sparse_euler_table_h_k} and \ref{tab:reacdiff_sparse_rk4_table_h_k}).

\begin{table}[p]
\vspace*{\fill}    
\centering         
\begin{adjustbox}{angle=90, scale=0.65, center} 
\begin{scriptsize}
\begin{tabular}{|c|l|l|}
\hline
$h(N)$ & \textbf{Euler-SINDy} & \textbf{25 Unr. Euler-SINDy} \\
\hline
\begin{tabular}{c}$h=1.25e-02$\\$(N=800)$\end{tabular} & \begin{tabular}{l}$1.001 x - 1.003 x^3 + 1.001 y^3 + 0.103 x_{11} + 0.102 x_{22} + 1.003 x^2 y - 1.004 x y^2$ \\ $1.005 y - 1.001 x^3 - 1.009 y^3 + 0.102 y_{11} + 0.102 y_{22} - 1.008 x^2 y - 0.998 x y^2$\end{tabular} & \begin{tabular}{l}$1.000 x - 1.000 x^3 + 1.000 y^3 + 0.101 x_{11} + 0.101 x_{22} + 1.001 x^2 y - 1.001 x y^2$ \\ $1.002 y - 1.000 x^3 - 1.002 y^3 + 0.101 y_{11} + 0.100 y_{22} - 1.002 x^2 y - 0.997 x y^2$\end{tabular} \\ \hline
\begin{tabular}{c}$h=5.00e-02$\\$(N=200)$\end{tabular} & \begin{tabular}{l}$1.002 x - 1.010 x^3 + 1.005 y^3 + 0.108 x_{11} + 0.108 x_{22} + 1.009 x^2 y - 1.013 x y^2$ \\ $1.015 y - 1.005 x^3 - 1.028 y^3 + 0.107 y_{11} + 0.107 y_{22} - 1.024 x^2 y - 1.001 x y^2$\end{tabular} & \begin{tabular}{l}$1.001 x - 1.001 x^3 + 1.000 y^3 + 0.101 x_{11} + 0.101 x_{22} + 1.001 x^2 y - 1.001 x y^2$ \\ $1.001 y - 1.001 x^3 - 1.001 y^3 + 0.101 y_{11} + 0.101 y_{22} - 1.001 x^2 y - 0.999 x y^2$\end{tabular} \\ \hline
\begin{tabular}{c}$h=7.50e-02$\\$(N=134)$\end{tabular} & \begin{tabular}{l}$1.002 x - 1.014 x^3 + 1.008 y^3 + 0.112 x_{11} + 0.112 x_{22} + 1.012 x^2 y - 1.019 x y^2$ \\ $1.022 y - 1.008 x^3 - 1.040 y^3 + 0.111 y_{11} + 0.111 y_{22} - 1.035 x^2 y - 1.002 x y^2$\end{tabular} & \begin{tabular}{l}$1.001 x - 1.001 x^3 + 1.000 y^3 + 0.101 x_{11} + 0.101 x_{22} + 1.001 x^2 y - 1.001 x y^2$ \\ $1.001 y - 1.001 x^3 - 1.002 y^3 + 0.101 y_{11} + 0.101 y_{22} - 1.001 x^2 y - 0.999 x y^2$\end{tabular} \\ \hline
\begin{tabular}{c}$h=1.00e-01$\\$(N=100)$\end{tabular} & \begin{tabular}{l}$1.002 x - 1.018 x^3 + 1.010 y^3 + 0.115 x_{11} + 0.115 x_{22} + 1.016 x^2 y - 1.024 x y^2$ \\ $1.028 y - 1.010 x^3 - 1.053 y^3 + 0.114 y_{11} + 0.114 y_{22} - 1.046 x^2 y - 1.003 x y^2$\end{tabular} & \begin{tabular}{l}$1.001 x - 1.001 x^3 + 1.000 y^3 + 0.101 x_{11} + 0.101 x_{22} + 1.002 x^2 y - 1.002 x y^2$ \\ $1.002 y - 1.001 x^3 - 1.002 y^3 + 0.101 y_{11} + 0.101 y_{22} - 1.001 x^2 y - 0.999 x y^2$\end{tabular} \\ \hline
\begin{tabular}{c}$h=1.25e-01$\\$(N=80)$\end{tabular} & \begin{tabular}{l}$1.002 x - 1.021 x^3 + 1.012 y^3 + 0.119 x_{11} + 0.119 x_{22} + 1.019 x^2 y - 1.029 x y^2$ \\ $0.981 y - 0.933 x^3 - 1.021 y^3 + 0.105 y_{11} + 0.105 y_{22} - 1.014 x^2 y - 0.930 x y^2 \plusterms{1} \hspace*{-3mm}$\end{tabular} & \begin{tabular}{l}$1.001 x - 1.001 x^3 + 1.000 y^3 + 0.102 x_{11} + 0.102 x_{22} + 1.002 x^2 y - 1.002 x y^2$ \\ $1.002 y - 1.001 x^3 - 1.003 y^3 + 0.101 y_{11} + 0.101 y_{22} - 1.002 x^2 y - 1.000 x y^2$\end{tabular} \\ \hline
\begin{tabular}{c}$h=2.50e-01$\\$(N=40)$\end{tabular} & \begin{tabular}{l}$0.995 x - 1.035 x^3 + 1.020 y^3 + 0.137 x_{11} + 0.137 x_{22} + 1.035 x^2 y - 1.048 x y^2$ \\ $0.941 y - 0.846 x^3 - 1.022 y^3 + 0.106 y_{11} + 0.106 y_{22} - 1.010 x^2 y - 0.840 x y^2 \plusterms{1} \hspace*{-3mm}$\end{tabular} & \begin{tabular}{l}$1.001 x - 1.002 x^3 + 1.001 y^3 + 0.102 x_{11} + 0.102 x_{22} + 1.002 x^2 y - 1.003 x y^2$ \\ $1.003 y - 1.001 x^3 - 1.005 y^3 + 0.102 y_{11} + 0.102 y_{22} - 1.004 x^2 y - 1.000 x y^2$\end{tabular} \\ \hline
\begin{tabular}{c}$h=3.75e-01$\\$(N=27)$\end{tabular} & \begin{tabular}{l}$0.980 x - 1.039 x^3 + 1.023 y^3 + 0.154 x_{11} + 0.154 x_{22} + 1.049 x^2 y - 1.056 x y^2$ \\ $0.881 y - 0.740 x^3 - 1.003 y^3 + 0.105 y_{11} + 0.105 y_{22} - 0.987 x^2 y - 0.730 x y^2 \plusterms{1} \hspace*{-3mm}$\end{tabular} & \begin{tabular}{l}$1.001 x - 1.003 x^3 + 1.002 y^3 + 0.103 x_{11} + 0.103 x_{22} + 1.003 x^2 y - 1.005 x y^2$ \\ $1.004 y - 1.002 x^3 - 1.008 y^3 + 0.103 y_{11} + 0.103 y_{22} - 1.006 x^2 y - 1.000 x y^2$\end{tabular} \\ \hline
\begin{tabular}{c}$h=6.25e-01$\\$(N=16)$\end{tabular} & \begin{tabular}{l}$1.165 x - 1.386 x^3 + 0.691 y^3 + 0.160 x_{11} + 0.160 x_{22} + 0.733 x^2 y - 1.383 x y^2 \plusterms{3} \hspace*{-3mm}$ \\ $0.712 y - 0.487 x^3 - 0.914 y^3 + 0.097 y_{11} + 0.097 y_{22} - 0.896 x^2 y - 0.467 x y^2 \plusterms{1} \hspace*{-3mm}$\end{tabular} & \begin{tabular}{l}$1.002 x - 1.005 x^3 + 1.003 y^3 + 0.104 x_{11} + 0.104 x_{22} + 1.004 x^2 y - 1.008 x y^2$ \\ $1.007 y - 1.003 x^3 - 1.013 y^3 + 0.104 y_{11} + 0.104 y_{22} - 1.010 x^2 y - 1.001 x y^2$\end{tabular} \\ \hline
\begin{tabular}{c}$h=1.00e+00$\\$(N=10)$\end{tabular} & \begin{tabular}{l}$1.112 x - 1.469 x^3 + 0.834 y^3 + 0.209 x_{11} + 0.209 x_{22} + 0.878 x^2 y - 1.407 x y^2 \plusterms{4} \hspace*{-3mm}$ \\ $0.331 y - 0.641 y^3 + 0.066 y_{11} + 0.066 y_{22} - 0.638 x^2 y \plusterms{1} \hspace*{-3mm}$\end{tabular} & \begin{tabular}{l}$1.002 x - 1.008 x^3 + 1.005 y^3 + 0.106 x_{11} + 0.106 x_{22} + 1.005 x^2 y - 1.012 x y^2$ \\ $1.012 y - 1.005 x^3 - 1.023 y^3 + 0.106 y_{11} + 0.106 y_{22} - 1.015 x^2 y - 1.003 x y^2$\end{tabular} \\ \hline
\end{tabular}
\end{scriptsize}
\end{adjustbox}
\vspace*{\fill}    
\caption{Robustness of Euler-SINDy and its unrolled version on Eq.~\ref{eq:2dreacdiff-V2}, with an increasing time step $h$ and a decreasing number of learning pairs. When Euler-SINDy fails to recover the governing equations, only the number of wrong additional terms is indicated.}
\label{tab:2React-Euler}
\end{table}

\begin{table}[p]
\vspace*{\fill}    
\centering         
\begin{adjustbox}{angle=90, scale=0.8, center} 
\begin{scriptsize}
\begin{tabular}{|c|l|l|}
\hline
$h(N)$ & \textbf{RK4-SINDy} & \textbf{25 Unr. RK4-SINDy} \\
\hline
\begin{tabular}{c}$h=1.25e-02$\\$(N=800)$\end{tabular} & \begin{tabular}{l}$1.001 x -1.000 x^3 + 1.000 y^3 + 0.101 x_{11} + 0.100 x_{22} + 1.001 x^2 y -1.001 x y^2$ \\ $1.002 y -1.000 x^3 -1.002 y^3 + 0.100 y_{11} + 0.100 y_{22} -1.002 x^2 y -0.997 x y^2$\end{tabular} & \begin{tabular}{l}$1.001 x -1.000 x^3 + 1.000 y^3 + 0.101 x_{11} + 0.100 x_{22} + 1.001 x^2 y -1.001 x y^2$ \\ $1.002 y -1.000 x^3 -1.002 y^3 + 0.100 y_{11} + 0.100 y_{22} -1.002 x^2 y -0.997 x y^2$\end{tabular} \\ \hline
\begin{tabular}{c}$h=5.00e-02$\\$(N=200)$\end{tabular} & \begin{tabular}{l}$1.001 x -1.000 x^3 + 1.000 y^3 + 0.101 x_{11} + 0.101 x_{22} + 1.001 x^2 y -1.001 x y^2$ \\ $1.001 y -1.000 x^3 -1.000 y^3 + 0.101 y_{11} + 0.101 y_{22} -1.000 x^2 y -0.999 x y^2$\end{tabular} & \begin{tabular}{l}$1.001 x -1.000 x^3 + 1.000 y^3 + 0.101 x_{11} + 0.101 x_{22} + 1.001 x^2 y -1.001 x y^2$ \\ $1.001 y -1.000 x^3 -1.000 y^3 + 0.101 y_{11} + 0.101 y_{22} -1.000 x^2 y -0.999 x y^2$\end{tabular} \\ \hline
\begin{tabular}{c}$h=7.50e-02$\\$(N=134)$\end{tabular} & \begin{tabular}{l}$1.001 x -1.000 x^3 + 1.000 y^3 + 0.101 x_{11} + 0.101 x_{22} + 1.001 x^2 y -1.001 x y^2$ \\ $1.000 y -1.000 x^3 -1.000 y^3 + 0.101 y_{11} + 0.101 y_{22} -0.999 x^2 y -0.999 x y^2$\end{tabular} & \begin{tabular}{l}$1.001 x -1.000 x^3 + 1.000 y^3 + 0.101 x_{11} + 0.101 x_{22} + 1.001 x^2 y -1.001 x y^2$ \\ $1.000 y -1.000 x^3 -1.000 y^3 + 0.101 y_{11} + 0.101 y_{22} -0.999 x^2 y -0.999 x y^2$\end{tabular} \\ \hline
\begin{tabular}{c}$h=1.00e-01$\\$(N=100)$\end{tabular} & \begin{tabular}{l}$1.001 x -1.000 x^3 + 1.000 y^3 + 0.101 x_{11} + 0.101 x_{22} + 1.001 x^2 y -1.001 x y^2$ \\ $1.000 y -1.000 x^3 -1.000 y^3 + 0.101 y_{11} + 0.101 y_{22} -0.999 x^2 y -0.999 x y^2$\end{tabular} & \begin{tabular}{l}$1.001 x -1.000 x^3 + 1.000 y^3 + 0.101 x_{11} + 0.101 x_{22} + 1.001 x^2 y -1.001 x y^2$ \\ $1.000 y -1.000 x^3 -1.000 y^3 + 0.101 y_{11} + 0.101 y_{22} -0.999 x^2 y -0.999 x y^2$\end{tabular} \\ \hline
\begin{tabular}{c}$h=1.25e-01$\\$(N=80)$\end{tabular} & \begin{tabular}{l}$1.001 x -1.000 x^3 + 1.000 y^3 + 0.101 x_{11} + 0.101 x_{22} + 1.001 x^2 y -1.001 x y^2$ \\ $1.000 y -1.000 x^3 -1.000 y^3 + 0.101 y_{11} + 0.101 y_{22} -0.999 x^2 y -0.999 x y^2$\end{tabular} & \begin{tabular}{l}$1.001 x -1.000 x^3 + 1.000 y^3 + 0.101 x_{11} + 0.101 x_{22} + 1.001 x^2 y -1.001 x y^2$ \\ $1.000 y -1.000 x^3 -1.000 y^3 + 0.101 y_{11} + 0.101 y_{22} -0.999 x^2 y -0.999 x y^2$\end{tabular} \\ \hline
\begin{tabular}{c}$h=2.50e-01$\\$(N=40)$\end{tabular} & \begin{tabular}{l}$1.001 x -1.000 x^3 + 1.000 y^3 + 0.101 x_{11} + 0.101 x_{22} + 1.001 x^2 y -1.001 x y^2$ \\ $1.000 y -1.000 x^3 -1.000 y^3 + 0.101 y_{11} + 0.101 y_{22} -0.999 x^2 y -0.999 x y^2$\end{tabular} & \begin{tabular}{l}$1.001 x -1.000 x^3 + 1.000 y^3 + 0.101 x_{11} + 0.101 x_{22} + 1.001 x^2 y -1.001 x y^2$ \\ $1.000 y -1.000 x^3 -1.000 y^3 + 0.101 y_{11} + 0.101 y_{22} -0.999 x^2 y -0.999 x y^2$\end{tabular} \\ \hline
\begin{tabular}{c}$h=3.75e-01$\\$(N=27)$\end{tabular} & \begin{tabular}{l}$1.001 x -1.001 x^3 + 1.000 y^3 + 0.101 x_{11} + 0.101 x_{22} + 1.001 x^2 y -1.001 x y^2$ \\ $1.000 y -1.001 x^3 -1.000 y^3 + 0.101 y_{11} + 0.101 y_{22} -0.999 x^2 y -1.000 x y^2$\end{tabular} & \begin{tabular}{l}$1.001 x -1.000 x^3 + 1.000 y^3 + 0.101 x_{11} + 0.101 x_{22} + 1.001 x^2 y -1.000 x y^2$ \\ $1.000 y -1.000 x^3 -0.999 y^3 + 0.101 y_{11} + 0.101 y_{22} -0.999 x^2 y -1.000 x y^2$\end{tabular} \\ \hline
\begin{tabular}{c}$h=6.25e-01$\\$(N=16)$\end{tabular} & \begin{tabular}{l}$1.002 x -1.007 x^3 + 1.002 y^3 + 0.101 x_{11} + 0.101 x_{22} + 1.005 x^2 y -1.007 x y^2$ \\ $1.000 y -1.004 x^3 -1.005 y^3 + 0.101 y_{11} + 0.101 y_{22} -1.004 x^2 y -1.003 x y^2$\end{tabular} & \begin{tabular}{l}$1.000 x -1.000 x^3 + 1.000 y^3 + 0.101 x_{11} + 0.101 x_{22} + 1.000 x^2 y -1.000 x y^2$ \\ $1.000 y -1.000 x^3 -0.999 y^3 + 0.101 y_{11} + 0.101 y_{22} -0.998 x^2 y -1.000 x y^2$\end{tabular} \\ \hline
\begin{tabular}{c}$h=1.00e+00$\\$(N=10)$\end{tabular} & \begin{tabular}{l}$0.950 x -0.950 x^3 + 1.169 y^3 + 0.114 x_{11} + 0.114 x_{22} + 1.056 x^2 y -0.833 x y^2 \plusterms{1} \hspace*{-3mm}$ \\ $0.890 y -0.611 x^3 -1.096 y^3 -1.029 x^2 y -0.674 x y^2 \plusterms{1} \hspace*{-3mm}$\end{tabular} & \begin{tabular}{l}$1.000 x -0.999 x^3 + 1.000 y^3 + 0.101 x_{11} + 0.101 x_{22} + 0.999 x^2 y -0.999 x y^2$ \\ $0.999 y -1.000 x^3 -0.998 y^3 + 0.101 y_{11} + 0.101 y_{22} -0.997 x^2 y -1.000 x y^2$\end{tabular} \\ \hline
\end{tabular}
\end{scriptsize}
\end{adjustbox}
\vspace*{\fill}
\caption{Robustness of RK4-SINDy and its unrolled version on Eq.~\ref{eq:2dreacdiff-V2}, with an increasing time step $h$ and a decreasing number of learning pairs. When RK4-SINDy fails to recover the governing equations, only the number of wrong additional terms is indicated.}
\label{tab:2D-react-RK4}
\end{table}

\begin{table}[t]
\centering
\caption{Accuracy of $K$-Unrolled Euler-SINDy with different unrolling depths $K$ and observation step sizes $h$ for the reaction-diffusion equation (Eq.~\ref{eq:2dreacdiff-V2}).}
\includegraphics[width=\textwidth]{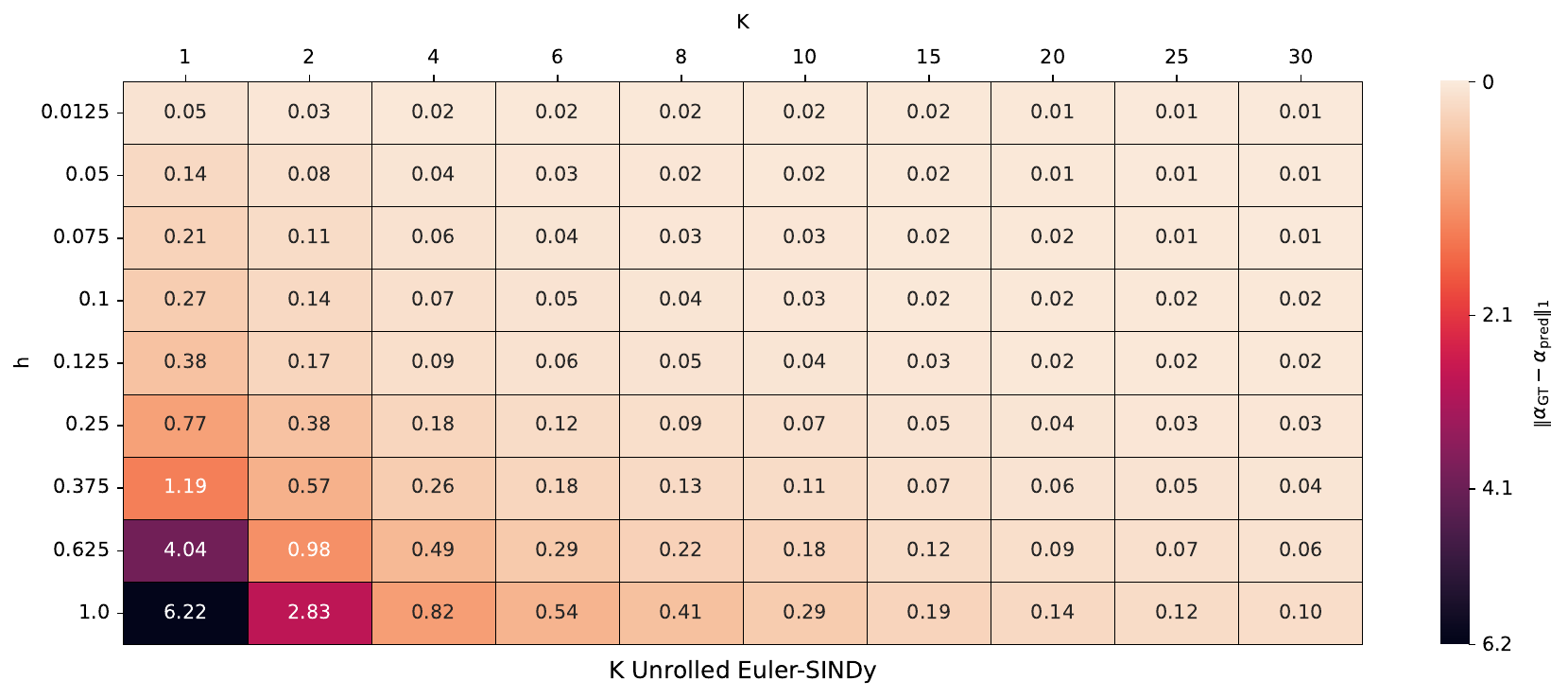}
 \label{tab:reacdiff_sparse_euler_table_h_k}
\end{table}

\begin{table}[t]
\centering
\caption{Accuracy of $K$-Unrolled RK4-SINDy with different unrolling depths $K$ and observation step sizes $h$ for the reaction-diffusion equation (Eq.~\ref{eq:2dreacdiff-V2}).  The marker $X$ denotes entries where the results are NaN.} 
\includegraphics[width=\textwidth]{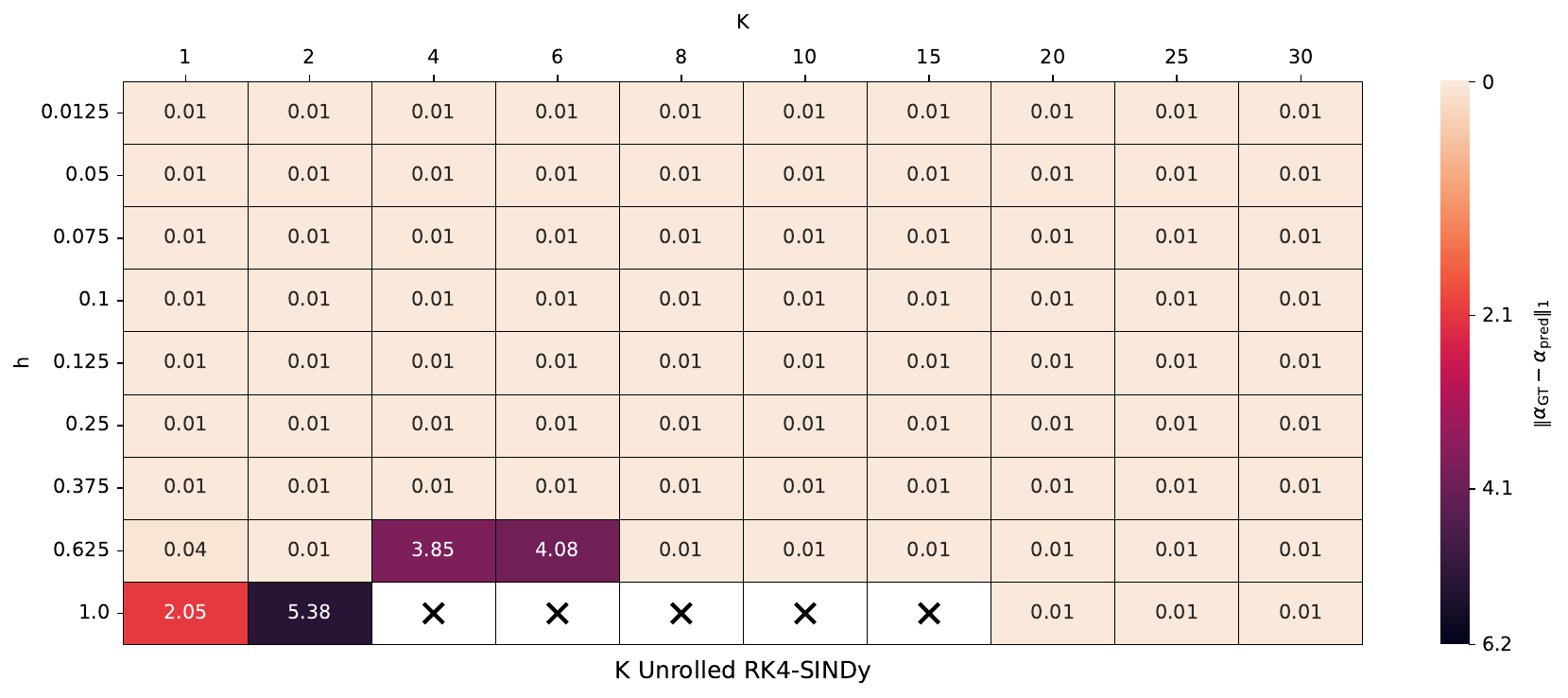}

\label{tab:reacdiff_sparse_rk4_table_h_k}
\end{table}

\subsection{Exact values for tables and NaN discussion } \label{suppl:errvshk}

\subsubsection{Exact values for tables}

In Tables~\ref{tab:exact_reacdiff_euler}, ~\ref{tab:exact_reacdiff_rk4}, ~\ref{tab:exact_kuramoto_euler}, ~\ref{tab:exact_kuramoto_rk4}, ~\ref{tab:exact_advection_euler}, ~\ref{tab:exact_advection_rk4}, ~\ref{tab:exact_oscillator_euler}, ~\ref{tab:exact_oscillator_rk4}, ~\ref{tab:exact_oscillator_sparse_euler}, we provide the precise numerical values from the results tables to ensure reproducibility of our experiments.

\begin{table}[h]
\centering
\caption{Exact numerical values of the accuracy of $K$-Unrolled Euler-SINDy with different unrolling depths $K$ and observation step sizes $h$ for the reaction-diffusion equation}
\begin{tiny}
\begin{verbatim}
unroll          1         2         4         6         8         10  \
dt_train                                                               
0.0125    0.046242  0.029178  0.021072  0.018472  0.016398  0.015680   
0.0500    0.143209  0.075789  0.041940  0.030209  0.024184  0.020664   
0.0750    0.206536  0.108930  0.058231  0.041380  0.032611  0.027218   
0.1000    0.269164  0.141793  0.074531  0.051498  0.039870  0.033129   
0.1250    0.382550  0.174994  0.091947  0.063114  0.048521  0.039855   
0.2500    0.769858  0.384209  0.176360  0.120409  0.091679  0.074206   
0.3750    1.190408  0.574098  0.257273  0.175996  0.134285  0.108614   
0.6250    4.042809  0.982508  0.489266  0.291211  0.222157  0.179478   
1.0001    6.217934  2.830413  0.816992  0.543290  0.407604  0.287270   

unroll          15        20        25        30  
dt_train                                          
0.0125    0.015172  0.014568  0.014150  0.013987  
0.0500    0.016426  0.014451  0.013093  0.012185  
0.0750    0.021055  0.016576  0.014678  0.013787  
0.1000    0.023920  0.019665  0.017072  0.015542  
0.1250    0.028212  0.022653  0.019197  0.017394  
0.2500    0.051073  0.039023  0.031984  0.027136  
0.3750    0.073464  0.056105  0.045445  0.038244  
0.6250    0.121230  0.091557  0.073443  0.061530  
1.0001    0.193090  0.144566  0.115114  0.095276 
\end{verbatim}
\end{tiny}
\label{tab:exact_reacdiff_euler}
\end{table}

\begin{table}[h]
\centering
\caption{Exact numerical values of the accuracy of $K$-Unrolled RK4-SINDy with different unrolling depths $K$ and observation step sizes $h$ for the reaction-diffusion equation}
\begin{tiny}
\begin{verbatim}
unroll          1         2         4         6         8         10  \
dt_train                                                               
0.0125    0.012855  0.012855  0.012834  0.012851  0.012857  0.012840   
0.0500    0.008413  0.008420  0.008420  0.008420  0.008420  0.008414   
0.0750    0.007793  0.007771  0.007771  0.007771  0.007762  0.007766   
0.1000    0.007708  0.007699  0.007671  0.007674  0.007672  0.007669   
0.1250    0.007818  0.007729  0.007713  0.007705  0.007705  0.007703   
0.2500    0.007564  0.007445  0.007520  0.007561  0.007562  0.007550   
0.3750    0.008718  0.007150  0.007353  0.007367  0.007368  0.007362   
0.6250    0.043449  0.007658  3.850548  4.084136  0.007694  0.007713   
1.0001    2.046546  5.381619       NaN       NaN       NaN       NaN   

unroll          15        20        25        30  
dt_train                                          
0.0125    0.012836  0.012857  0.012836  0.012862  
0.0500    0.008414  0.008410  0.008418  0.008421  
0.0750    0.007779  0.007774  0.007763  0.007765  
0.1000    0.007682  0.007670  0.007666  0.007678  
0.1250    0.007708  0.007702  0.007706  0.007713  
0.2500    0.007544  0.007556  0.007547  0.007547  
0.3750    0.007376  0.007382  0.007328  0.007308  
0.6250    0.007739  0.007729  0.007694  0.007728  
1.0001         NaN  0.012770  0.012801  0.012801 
\end{verbatim}
\end{tiny}
\label{tab:exact_reacdiff_rk4}
\end{table}

\begin{table}[h]
\centering
\caption{Exact numerical values of the accuracy of $K$-Unrolled Euler-SINDy with different unrolling depths $K$ and observation step sizes $h$ for the Kuramoto-Sivashinsky equation}
\begin{tiny}
\begin{verbatim}
unroll          1         2         3         4         5         6   \
dt_train                                                               
0.002     0.290433  0.284235  0.282468  0.281390  0.280735  0.280262   
0.004     0.292767  0.277847  0.285475  0.289942  0.292812  0.294319   
0.020     0.500157  0.287455  0.269833  0.261670  0.262745  0.269243   
0.040     1.049682  0.301632  0.298395  0.311471  0.317574  0.321438   
0.060     1.633355  0.330752  0.250329  0.361343  0.362094  0.364552   
0.080     2.107891  0.731860  0.377969  0.388942  0.385852  0.390808   
0.100     2.499690  0.214895  0.264847  0.237942  0.347438  0.410326   
0.160     3.349287  5.207612  5.225735  6.510035  0.595032  0.340408   
0.200     3.743670  6.514644  5.134880  6.515187  6.515296  6.515370   

unroll          7         8         9         10        15        20  
dt_train                                                              
0.002     0.279804  0.279554  0.279266  0.279183  0.279189  0.278954  
0.004     0.295529  0.296454  0.297358  0.298260  0.299678  0.300505  
0.020     0.273735  0.277338  0.279609  0.281728  0.288045  0.290853  
0.040     0.324669  0.326371  0.328593  0.329710  0.333738  0.335783  
0.060     0.366458  0.367389  0.368930  0.369419  0.372146  0.373482  
0.080     0.392260  0.392944  0.393869  0.394542  0.396333  0.396929  
0.100     0.408391  0.409379  0.409867  0.410223  0.411323  0.412225  
0.160     0.184852  0.426776  0.427713  0.430640  0.431321  0.431690  
0.200     5.454909  0.365776  0.203297  0.430028  0.434726  0.434908
\end{verbatim}
\end{tiny}
\label{tab:exact_kuramoto_euler}
\end{table}

\begin{table}[h]
\centering
\caption{Exact numerical values of the accuracy of $K$-Unrolled RK4-SINDy with different unrolling depths $K$ and observation step sizes $h$ for the Kuramoto-Sivashinsky equation}
\begin{tiny}
\begin{verbatim}
unroll          1         2         3         4         5         6   \
dt_train                                                               
0.002     0.277920  0.277938  0.277919  0.277954  0.277925  0.278190   
0.004     0.302744  0.302737  0.302762  0.302830  0.302792  0.302735   
0.020     0.324396  0.299542  0.298787  0.298669  0.298680  0.298673   
0.040     0.384603  0.350953  0.342493  0.341847  0.341809  0.341815   
0.060     5.066454  0.313697  0.379681  0.377625  0.377360  0.377320   
0.080     6.500150  0.469114  0.409304  0.400637  0.399971  0.399840   
0.100     6.502722  6.502722  0.373876  0.418929  0.414179  0.413881   
0.160     6.510501       NaN       NaN       NaN  0.377725  0.436245   
0.200     6.515738  6.515738       NaN       NaN       NaN  5.071077   

unroll          7         8         9         10        15        20  
dt_train                                                              
0.002     0.277946  0.278192  0.277937  0.277936  0.277926  0.277948  
0.004     0.302788  0.302787  0.302748  0.302792  0.302747  0.302763  
0.020     0.298649  0.298658  0.298615  0.298639  0.298683  0.298626  
0.040     0.341817  0.341805  0.341672  0.341676  0.341822  0.341651  
0.060     0.377309  0.377357  0.377366  0.377016  0.377346  0.377378  
0.080     0.399817  0.399830  0.399740  0.399835  0.399712  0.399837  
0.100     0.413932  0.413953  0.413941  0.413968  0.413912  0.413912  
0.160     0.432981  0.432881  0.432846  0.432845  0.432846  0.432831  
0.200     0.383934  0.436598  0.436149  0.436170  0.436023  0.436125 
\end{verbatim}
\end{tiny}
\label{tab:exact_kuramoto_rk4}
\end{table}

\begin{table}[h]
\centering
\caption{Exact numerical values of the accuracy of $K$-Unrolled Euler-SINDy with different unrolling depths $K$ and observation step sizes $h$ for the advection equation}
\begin{tiny}
\begin{verbatim}
unroll              1         2             3             4             5   \
dt_train                                                                     
0.0002    1.173615e-05  0.000038  3.820062e-05  1.164675e-05  1.164675e-05   
0.0004    1.335144e-06  0.000001  1.305342e-06  1.335144e-06  1.099110e-05   
0.0010    2.419949e-06  0.000002  2.509356e-06  2.598763e-06  1.484156e-06   
0.0020    9.000301e-07  0.000002  2.121925e-06  2.062321e-06  2.002716e-06   
0.0040    1.266406e-02  0.000002  9.775162e-07  2.026558e-07  1.549721e-07   
0.0080    2.529175e-02  0.012659  1.006722e-05  7.563829e-06  5.984306e-06   
0.0100    3.162265e-02  0.015826  1.056524e-02  1.325607e-05  1.051426e-05   
0.0300    9.507964e-02  0.047536  3.172120e-02  2.387895e-02  1.920243e-02   
0.0400    1.268950e-01  0.063414  4.230794e-02  3.190674e-02  2.573017e-02   

unroll              6             7             8             9         10  \
dt_train                                                                     
0.0002    3.796220e-05  3.808141e-05  3.808141e-05  3.820062e-05  0.000038   
0.0004    1.116991e-05  1.099110e-05  1.114011e-05  1.102090e-05  0.000011   
0.0010    1.275539e-06  1.335144e-06  1.335144e-06  1.454353e-06  0.000001   
0.0020    2.181530e-06  2.449751e-06  1.943111e-06  2.032518e-06  0.000002   
0.0040    3.933907e-07  6.616116e-07  7.510185e-07  8.106232e-07  0.000001   
0.0080    4.762411e-06  4.136562e-06  3.510714e-06  3.212690e-06  0.000003   
0.0100    8.726120e-06  7.563829e-06  6.729364e-06  5.924702e-06  0.000005   
0.0300    1.608529e-02  1.385471e-02  1.217951e-02  1.087188e-02  0.000063   
0.0400    2.159350e-02  1.859933e-02  1.634081e-02  1.459157e-02  0.013196   

unroll          15            20            25            30  
dt_train                                                      
0.0002    0.000011  1.164675e-05  1.146793e-05  1.182556e-05  
0.0004    0.000001  1.394749e-06  1.484156e-06  1.543760e-06  
0.0010    0.000001  1.275539e-06  2.658367e-06  2.568960e-06  
0.0020    0.000003  2.688169e-06  2.598763e-06  2.628565e-06  
0.0040    0.000001  1.436472e-06  1.585484e-06  1.585484e-06  
0.0080    0.000001  7.092953e-07  4.112720e-07  2.026558e-07  
0.0100    0.000004  2.557039e-06  1.841784e-06  1.633167e-06  
0.0300    0.000047  3.817081e-05  3.301501e-05  2.958775e-05  
0.0400    0.000085  7.038713e-05  6.129742e-05  5.518794e-05 
\end{verbatim}
\end{tiny}
\label{tab:exact_advection_euler}
\end{table}

\begin{table}[h]
\centering
\caption{Exact numerical values of the accuracy of $K$-Unrolled RK4-SINDy with different unrolling depths $K$ and observation step sizes $h$ for the advection equation}
\begin{tiny}
\begin{verbatim}
unroll              1             2             3             4   \
dt_train                                                           
0.0002    1.173615e-05  1.161695e-05  1.173615e-05  1.173615e-05   
0.0004    1.364946e-06  1.096129e-05  1.364946e-06  1.305342e-06   
0.0010    1.245737e-06  1.215935e-06  2.568960e-06  1.215935e-06   
0.0020    2.568960e-06  2.688169e-06  2.568960e-06  2.330542e-06   
0.0040    1.794100e-06  2.092123e-06  2.002716e-06  2.002716e-06   
0.0080    1.078844e-06  1.078844e-06  1.078844e-06  1.168251e-06   
0.0100    2.443790e-07  4.529953e-07  4.231930e-07  4.529953e-07   
0.0300    2.365708e-05  1.632571e-05  1.182556e-05  1.215339e-05   
0.0400    6.988049e-05  3.750861e-04  1.441777e-04  2.601147e-05   
0.1000    3.005036e-02           NaN           NaN           NaN   
0.1500    1.266138e-01           NaN           NaN           NaN   

unroll              5             6             7             8   \
dt_train                                                           
0.0002    1.164675e-05  3.796220e-05  3.805161e-05  3.796220e-05   
0.0004    1.364946e-06  1.096129e-05  1.096129e-05  1.102090e-05   
0.0010    1.215935e-06  1.215935e-06  2.568960e-06  2.539158e-06   
0.0020    2.568960e-06  2.568960e-06  2.568960e-06  2.568960e-06   
0.0040    2.062321e-06  1.853704e-06  1.823902e-06  1.794100e-06   
0.0080    1.078844e-06  1.078844e-06  1.108646e-06  1.108646e-06   
0.0100    4.231930e-07  4.231930e-07  4.231930e-07  2.443790e-07   
0.0300    1.221299e-05  1.230240e-05  1.224279e-05  1.230240e-05   
0.0400    2.452135e-05  2.449155e-05  2.443194e-05  2.449155e-05   
0.1000             NaN           NaN           NaN           NaN   
0.1500             NaN           NaN           NaN           NaN   

unroll              9             10            15            20  \
dt_train                                                           
0.0002    3.805161e-05  3.805161e-05  3.763437e-05  3.802180e-05   
0.0004    1.102090e-05  1.108050e-05  1.114011e-05  1.149774e-05   
0.0010    2.568960e-06  2.568960e-06  2.568960e-06  2.568960e-06   
0.0020    2.717972e-06  2.568960e-06  2.658367e-06  2.717972e-06   
0.0040    1.794100e-06  2.002716e-06  2.062321e-06  1.943111e-06   
0.0080    1.108646e-06  1.168251e-06  1.198053e-06  1.227856e-06   
0.0100    3.039837e-07  2.443790e-07  3.933907e-07  3.039837e-07   
0.0300    1.230240e-05  1.218319e-05  1.218319e-05  1.221299e-05   
0.0400    2.449155e-05  2.449155e-05  2.461076e-05  2.467036e-05   
0.1000             NaN           NaN  1.971364e-04  1.968682e-04   
0.1500             NaN           NaN           NaN  4.986465e-04   

unroll              25            30  
dt_train                              
0.0002    3.811121e-05  3.852844e-05  
0.0004    1.096129e-05  1.069307e-05  
0.0010    1.245737e-06  2.568960e-06  
0.0020    2.568960e-06  2.568960e-06  
0.0040    1.823902e-06  1.704693e-06  
0.0080    1.078844e-06  1.198053e-06  
0.0100    2.443790e-07  3.039837e-07  
0.0300    1.230240e-05  1.218319e-05  
0.0400    2.458096e-05  2.461076e-05  
0.1000    1.966894e-04  1.970768e-04  
0.1500    4.953980e-04  4.950702e-04 
\end{verbatim}
\end{tiny}
\label{tab:exact_advection_rk4}
\end{table}

\begin{table}[h]
\centering
\caption{Exact numerical values of the accuracy of $K$-Unrolled Euler-SINDy with different unrolling depths $K$ and observation step sizes $h$ for the cubic damped oscillator equation}
\begin{tiny}
\begin{verbatim}
unroll          1         2         3         4         5         6   \
dt_train                                                               
0.0002    0.011839  0.012030  0.012096  0.012126  0.012146  0.012159   
0.0020    0.011708  0.010446  0.011063  0.011373  0.011559  0.011683   
0.0200    0.065739  0.035763  0.025860  0.020924  0.017968  0.016004   
0.0400    0.219144  0.065993  0.046025  0.036112  0.030186  0.026249   
0.1000    1.208356  0.280733  0.175847  0.081809  0.066870  0.056980   
0.4000    6.893707  3.170178  2.073294  0.954799  0.513009  0.866129   
0.5000    7.224839  4.004562  2.958772  3.007558  2.172986  2.282994   
0.6000    8.950206  5.993519  3.335531  2.197650  2.351094  1.164204   

unroll          7         8         9         10        11        12  \
dt_train                                                               
0.0002    0.012168  0.012173  0.012179  0.012183  0.012186  0.012190   
0.0020    0.011768  0.011835  0.011889  0.011931  0.011965  0.011993   
0.0200    0.014598  0.013546  0.012727  0.012073  0.011537  0.011102   
0.0400    0.023438  0.021333  0.019696  0.018391  0.017318  0.016431   
0.1000    0.049945  0.044692  0.040617  0.037360  0.034703  0.032492   
0.4000    0.924864  0.300969  0.263655  0.234217  0.210415  0.190783   
0.5000    2.128182  0.703083  0.636216  0.172422  0.158125  0.146251   
0.6000    1.017468  1.204637  0.739615  0.682376  0.635191  0.331186   

unroll          13        14        15        20        30        40        50  
dt_train                                                                        
0.0002    0.012192  0.012194  0.012209  0.012216  0.012222  0.012212  0.012216  
0.0020    0.012015  0.012037  0.012055  0.012116  0.012178  0.012212  0.012229  
0.0200    0.011055  0.011016  0.010982  0.011121  0.011742  0.012051  0.012241  
0.0400    0.015678  0.015036  0.014471  0.012521  0.011580  0.011857  0.012232  
0.1000    0.030625  0.029025  0.027637  0.022798  0.017976  0.015569  0.014128  
0.4000    0.150957  0.140261  0.094612  0.075671  0.056964  0.047699  0.042169  
0.5000    0.136235  0.127672  0.120267  0.094484  0.068915  0.056215  0.048623  
0.6000    0.312322  0.160111  0.150970  0.119183  0.087734  0.072145  0.062834  
\end{verbatim}
\end{tiny}
\label{tab:exact_oscillator_euler}
\end{table}

\begin{table}[h]
\centering
\caption{Exact numerical values of the accuracy of $K$-Unrolled RK4-SINDy with different unrolling depths $K$ and observation step sizes $h$ for the cubic damped oscillator equation}
\begin{tiny}
\begin{verbatim}
unroll          1         2         3         4         5         6   \
dt_train                                                               
0.0002    0.012221  0.012221  0.012221  0.012221  0.012221  0.012221   
0.0020    0.012306  0.012306  0.012306  0.012306  0.012306  0.012307   
0.0200    0.012980  0.012981  0.012981  0.012981  0.012981  0.012981   
0.0400    0.013720  0.013723  0.013724  0.013724  0.013724  0.013724   
0.1000    0.015627  0.015815  0.015825  0.015828  0.015828  0.015828   
0.4000    0.576687  0.022302  0.023917  0.024295  0.024410  0.024453   
0.5000    0.370792  0.014308  0.017563  0.018523  0.018826  0.018941   
0.6000    1.695693  0.019368  0.022926  1.838034  1.782619  0.025580   

unroll          7         8         9         10        11        12  \
dt_train                                                               
0.0002    0.012221  0.012221  0.012221  0.012221  0.012221  0.012221   
0.0020    0.012306  0.012306  0.012306  0.012306  0.012306  0.012306   
0.0200    0.012981  0.012981  0.012981  0.012981  0.012981  0.012981   
0.0400    0.013724  0.013724  0.013724  0.013724  0.013724  0.013724   
0.1000    0.015828  0.015828  0.015828  0.015829  0.015828  0.015828   
0.4000    0.024471  0.024482  0.024487  0.024490  0.024492  0.024493   
0.5000    0.018993  0.019019  0.019033  0.019041  0.019045  0.019049   
0.6000    0.025748  0.025828  0.025871  0.025896  0.025912  0.025920   

unroll          13        14        15        20        30        40        50  
dt_train                                                                        
0.0002    0.012221  0.012221  0.012221  0.012221  0.012221  0.012221  0.012221  
0.0020    0.012306  0.012307  0.012306  0.012306  0.012306  0.012306  0.012306  
0.0200    0.012981  0.012981  0.012981  0.012981  0.012981  0.012981  0.012981  
0.0400    0.013724  0.013724  0.013724  0.013724  0.013724  0.013723  0.013724  
0.1000    0.015828  0.015829  0.015828  0.015829  0.015829  0.015829  0.015828  
0.4000    0.024493  0.024494  0.024495  0.024495  0.024496  0.024495  0.024495  
0.5000    0.019050  0.019052  0.019053  0.019055  0.019055  0.019056  0.019055  
0.6000    0.025926  0.025931  0.025934  0.025940  0.025942  0.025943  0.025941  
\end{verbatim}
\end{tiny}
\label{tab:exact_oscillator_rk4}
\end{table}

\begin{table}[h]
\centering
\caption{Exact numerical values of the accuracy of $K$-Unrolled Euler-SINDy with different unrolling depths $K$ and observation step sizes $h$ for the cubic damped oscillator equation. \textbf{The number of training pairs $N$ is kept constant.}}
\begin{tiny}
\begin{verbatim}
unroll          1         2         4         6         8         10  \
dt_train                                                               
0.0002    0.011862  0.012049  0.012142  0.012172  0.012188  0.012197   
0.0004    0.011495  0.011866  0.012025  0.012082  0.012111  0.012129   
0.0010    0.010372  0.011305  0.011768  0.011918  0.011994  0.012043   
0.0020    0.011667  0.010410  0.011336  0.011643  0.011799  0.011890   
0.0200    0.065306  0.035343  0.020496  0.015573  0.013110  0.011628   
0.0300    0.126438  0.050244  0.027891  0.020494  0.016803  0.014595   
0.0500    0.280594  0.080242  0.042694  0.030326  0.024172  0.020488   
0.1000    2.717628  0.277419  0.079931  0.054965  0.042604  0.035233   
0.2000    5.778956  2.989165  0.266794  0.104895  0.079848  0.065000   
0.3000    7.561143  4.478805  0.403619  0.202406  0.151908  0.096000   
0.4000    8.271039  5.214392  3.197492  0.529979  0.199636  0.129392   

unroll          20        30        40        50  
dt_train                                          
0.0002    0.012203  0.012209  0.012225  0.012228  
0.0004    0.012201  0.012214  0.012220  0.012225  
0.0010    0.012138  0.012169  0.012201  0.012210  
0.0020    0.012055  0.012116  0.012170  0.012190  
0.0200    0.010698  0.011314  0.011600  0.011781  
0.0300    0.010662  0.010855  0.011319  0.011597  
0.0500    0.013145  0.010863  0.010727  0.011175  
0.1000    0.020571  0.015711  0.013291  0.011847  
0.2000    0.035703  0.026062  0.021264  0.018397  
0.3000    0.051465  0.036869  0.029619  0.025283  
0.4000    0.068332  0.048259  0.038312  0.032370
\end{verbatim}
\end{tiny}
\label{tab:exact_oscillator_sparse_euler}
\end{table}

\subsubsection{Discussion about NaN values}

As mentioned in Sec.~\ref{sec:limitations}, a potential issue with unrolling is the appearance of numerical instabilities, which can lead to NaN values in $\alpha_{\text{pred}}$, the vector of predicted coefficients of the governing equations. These instabilities typically arise in challenging scenarios where the time step $h$ is relatively large and the unrolling depth $K$ is insufficient, meaning the numerical scheme has not been unrolled far enough to maintain stability. However, this is not a fundamental limitation of the method. When $K$ is increased appropriately, the NaN issues vanish entirely. In other words, the method’s stability can always be restored by unrolling deeper, ensuring accurate recovery of the system’s coefficients while retaining the advantages of the unrolled approach.

\subsection{Robustness to corrupted data} \label{sec:expes-noise}

\begin{figure}[t]
\centering
\includegraphics[width=\textwidth]{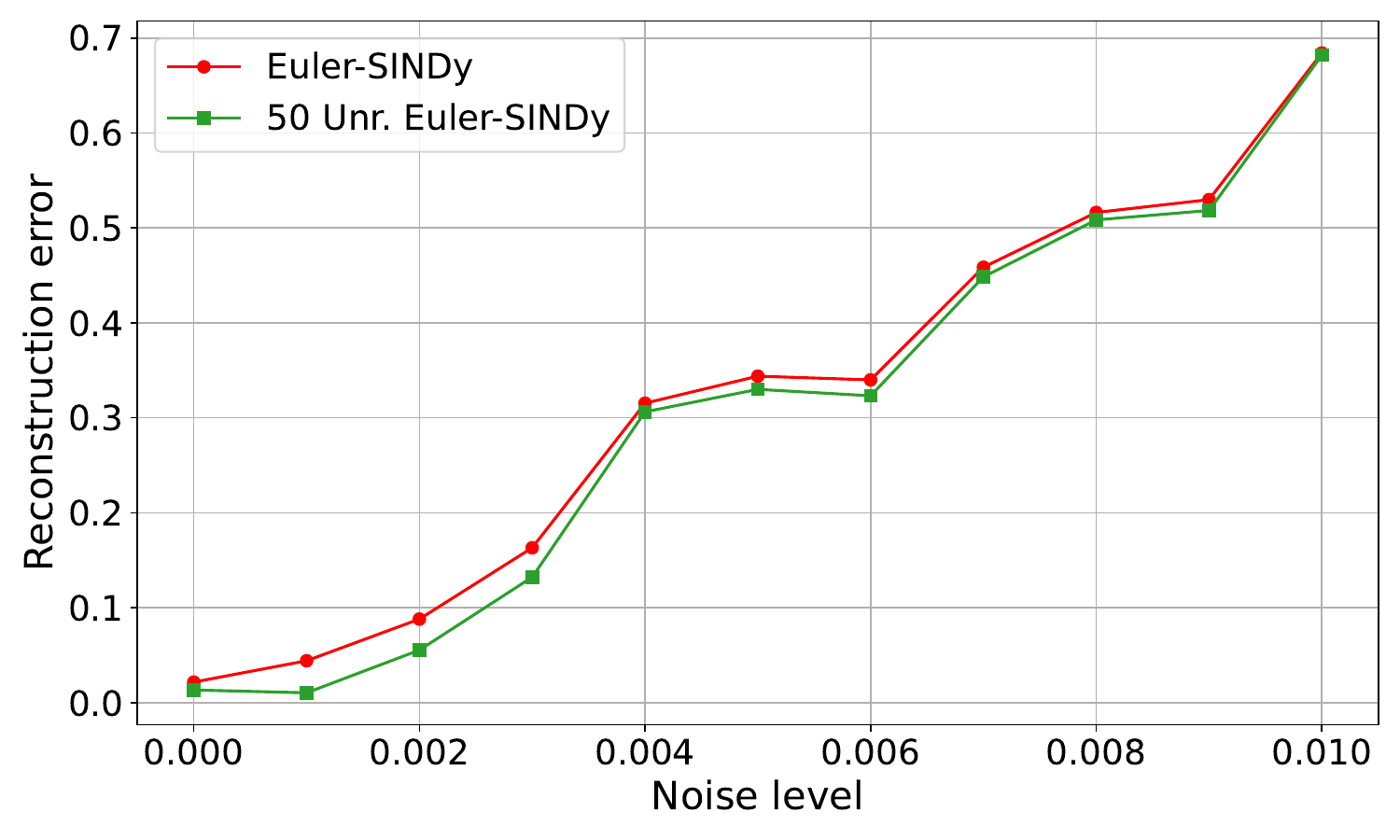}
\caption{Evolution of the reconstruction error of Euler-SINDy and 50 Unrolled Euler-SINDy on the 2-dimensional cubic damped oscillator as the data is more and more corrupted with a gaussian noise.} \label{fig:noise_evolution}
\end{figure}

We would like to point out that Unrolled SINDy (with Euler or RK4) was not specifically designed to handle situations where the data is corrupted. Figure~\ref{fig:noise_evolution} depicts this behavior on the 2-dimensional cubic damped oscillator with an increasing level of Gaussian noise and running Euler-SINDy and its unrolled version. The corrupted data has been generated as follows: 

Noisy observations are generated as 
\[
\tilde{x}(t) = x(t) + \eta_x(t), \quad \tilde{y}(t) = y(t) + \eta_y(t),
\]
where $\eta_x(t)$ and $\eta_y(t)$ are independent Gaussian noise with standard deviations 
\[
\sigma_x = \sigma \, \mathrm{std}(x(t)), \quad \sigma_y = \sigma \, \mathrm{std}(y(t))
\] 
respectively. The parameter $\sigma$ controls the amplitude of the noise before adding random perturbations with zero mean. 

From Figure~\ref{fig:noise_evolution} we can make the following remarks: 
(i) when the noise level is small, Unrolled Euler SINDy retains its advantage over the standard Euler-SINDy version by leveraging the unrolling scheme; 
(ii) as the noise increases, the gap between the two methods tends to narrow; 
(iii) from a certain noise level (here 0.01), both methods behave similarly and unrolling no longer provides any benefit, as the data is too corrupted to allow reliable intermediate estimates.

\begin{figure}[t]
\begin{center}

\includegraphics[width=0.45\textwidth]{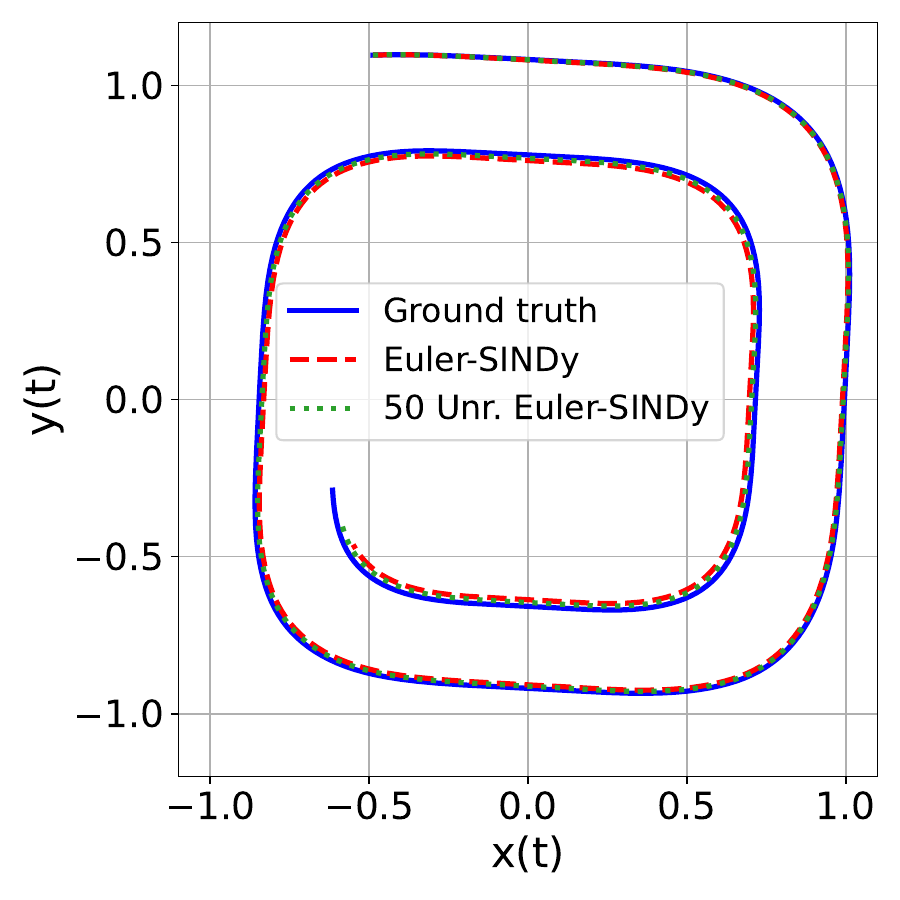}
\hfill
\includegraphics[width=0.45\textwidth]{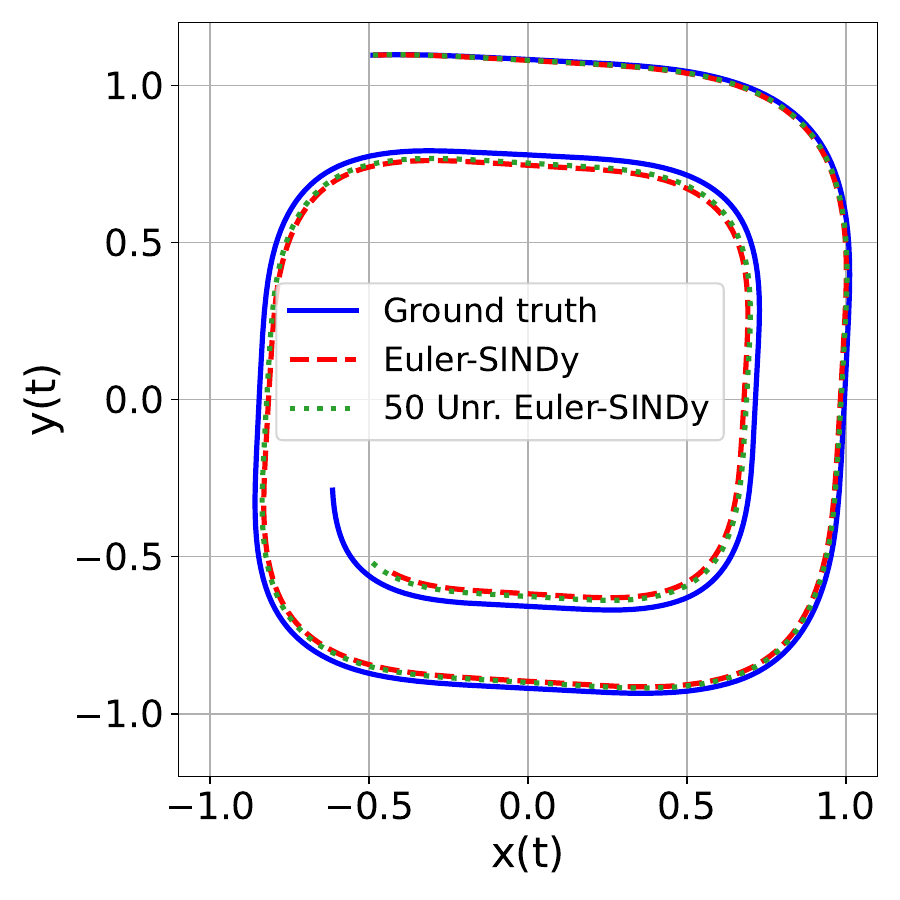}

\vspace{-0.3cm}

\makebox[0.45\textwidth][l]{\hspace{1cm}a)}\hfill
\makebox[0.45\textwidth][l]{\hspace{1cm}b)}\hfill

\caption{Impact of the noise on the capacity of the methods to recover the 2-dimensional cubic damped oscillator (where $h=0.02)$ with an increasing noise rate; a) $\epsilon=2 \cdot10^{-2}$, b) $\epsilon=3 \cdot10^{-2}$}
\label{fig:noise_robustness1}
\end{center}
\end{figure}

\begin{figure}[t]
\begin{center}

\includegraphics[width=0.45\textwidth]{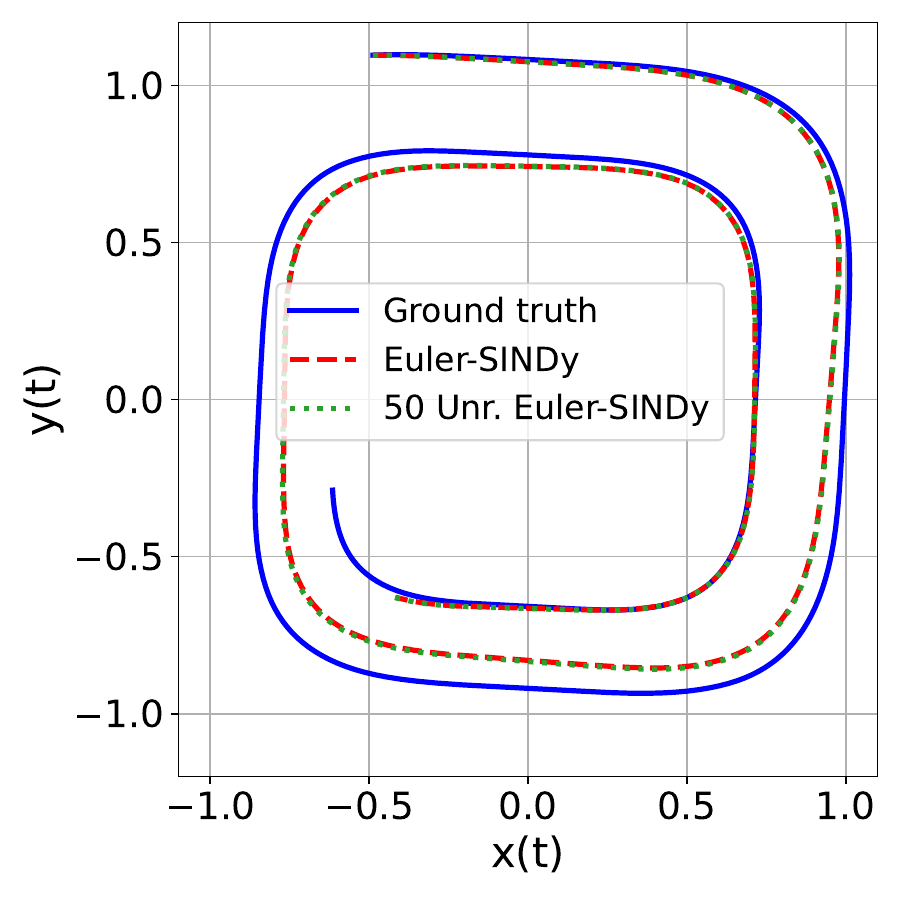}
\hfill
\includegraphics[width=0.45\textwidth]{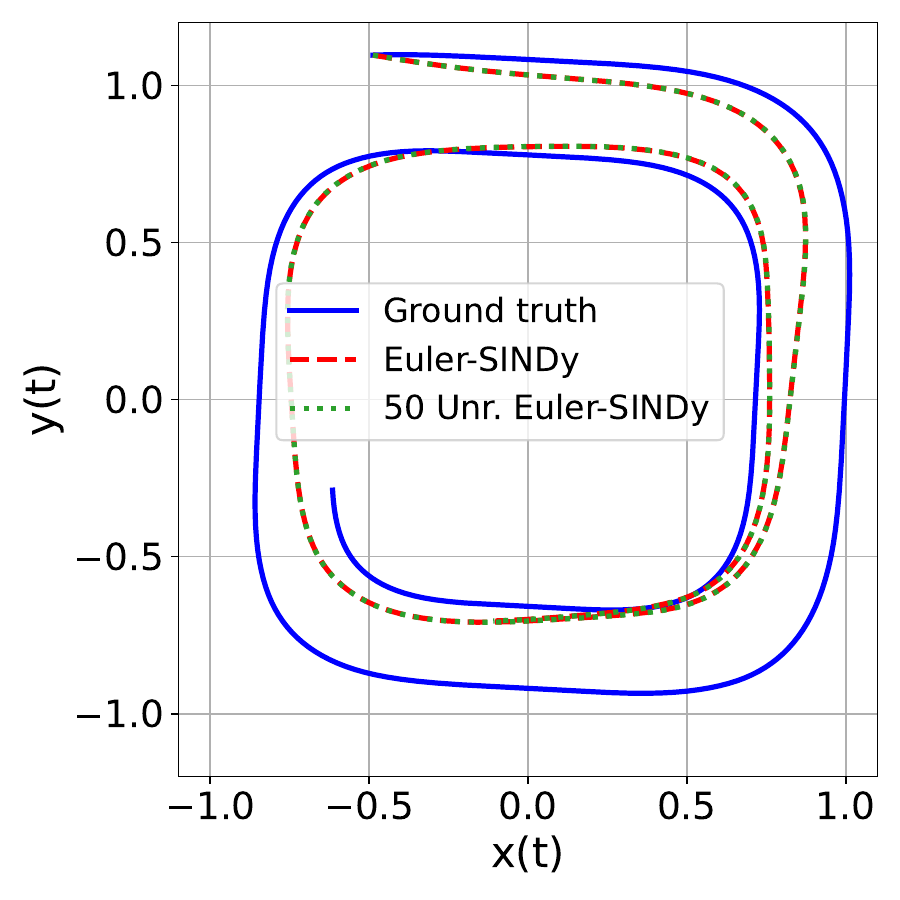}

\vspace{-0.3cm}

\makebox[0.45\textwidth][l]{\hspace{1cm}c)}\hfill
\makebox[0.45\textwidth][l]{\hspace{1cm}d)}\hfill

\caption{Impact of the noise on the capacity of the methods to recover the 2-dimensional cubic damped oscillator (where $h=0.02)$ with an increasing noise rate; c) $\epsilon=5 \cdot10^{-2}$, d) $\epsilon=1 \cdot10^{-1}$}
\label{fig:noise_robustness2}
\end{center}
\end{figure}

In order to highlight the impact of the discrepancies between the red and green curves in Figure~\ref{fig:noise_evolution}, we report in Figures \ref{fig:noise_robustness1} and \ref{fig:noise_robustness2} the solutions of the ODE corresponding to several representative points of the curves.

\subsection{1D Kuramoto–Sivashinsky PDE \label{sec:kuramoto_sup}}

The Kuramoto–Sivashinsky (KS) equation is recalled below:

\begin{eqnarray}
    u_t =  - u_{xx} - u_{xxxx} - 5 uu_x , \label{eq:1Dkuramoto}
\end{eqnarray}
where $u_t$ denotes the time derivative, $u_{x}$, $u_{xx}$ and $u_{xxxx}$ denote the first, second and fourth spatial derivatives, respectively, and $uu_x$ is the nonlinear term.

The experimental setup and results for the Kuramoto–Sivashinsky (KS) equation are detailed in Sec.~\ref{sec:kuramoto}. Table~\ref{tab:kuramoto_table} presents the recovered analytical expressions obtained with Euler-SINDy, RK4-SINDy, and their unrolled variants. For the unrolled methods, we report results corresponding to the value of $K$ that minimizes the training error, as performance remains stable with respect to $K$ (see Tables~\ref{tab:kuramoto_sparse_euler_table_h_k} and \ref{tab:kuramoto_sparse_rk4_table_h_k}).

\begin{figure}[t]
\centering
\includegraphics[width=\textwidth]{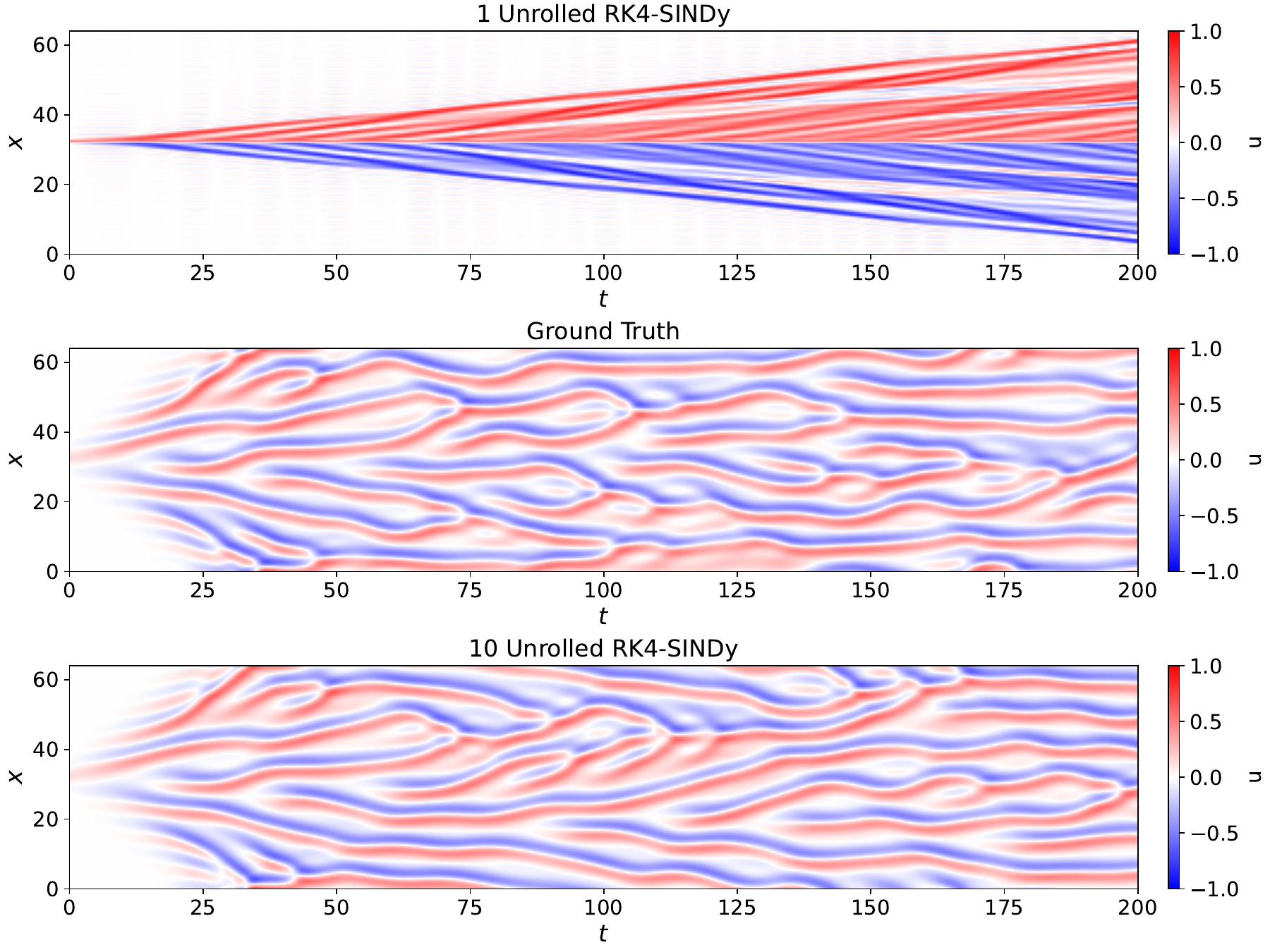}
\caption{Solutions of the Kuramoto–Sivashinsky PDE with $h=0.2$. From top to bottom: RK4-SINDy, ground truth, and 10 Unrolled RK4-SINDy.}
\label{fig:kuramoto_figure_rk4}
\end{figure}

Fig.~\ref{fig:kuramoto_figure_rk4} illustrates the qualitative difference between standard RK4-SINDy and 10-Unrolled RK4-SINDy in this challenging scenario. While RK4-SINDy fails to capture the key dynamics, Unrolled RK4-SINDy closely tracks the dominant trends and spatial structures. Minor local deviations arise due to the chaotic nature of the KS equation, but overall, the recovered coefficients remain very close to the true dynamics.

\begin{table}[p]
\vspace*{\fill}    
\centering         
\begin{adjustbox}{angle=90, scale=0.8, center} 
\begin{scriptsize}
\begin{tabular}{|c|l|l!{\color{red}\vrule width 1.5pt}l|l|}
\hline
$h(N)$ & \textbf{Euler-SINDy} & \textbf{10 Unrolled Euler-SINDy} & \textbf{RK4-SINDy} & \textbf{10 Unrolled RK4-SINDy} \\
\hline

\begin{tabular}{c}$h=2.000e-03$\\$(N=100000)$\end{tabular} & \begin{tabular}{l}$ -1.081 u_{xx} -1.189 u_{xxxx} -4.979 uu_x$\end{tabular} & \begin{tabular}{l}$ -1.084 u_{xx} -1.193 u_{xxxx} -4.998 uu_x$\end{tabular} & \begin{tabular}{l}$ -1.085 u_{xx} -1.193 u_{xxxx} -5.000 uu_x$\end{tabular} & \begin{tabular}{l}$ -1.085 u_{xx} -1.193 u_{xxxx} -5.000 uu_x$\end{tabular} \\ \hline
\begin{tabular}{c}$h=4.000e-03$\\$(N=50000)$\end{tabular} & \begin{tabular}{l}$ -1.079 u_{xx} -1.187 u_{xxxx} -4.974 uu_x$\end{tabular} & \begin{tabular}{l}$ -1.087 u_{xx} -1.196 u_{xxxx} -5.015 uu_x$\end{tabular} & \begin{tabular}{l}$ -1.088 u_{xx} -1.197 u_{xxxx} -5.018 uu_x$\end{tabular} & \begin{tabular}{l}$ -1.088 u_{xx} -1.197 u_{xxxx} -5.018 uu_x$\end{tabular} \\ \hline
\begin{tabular}{c}$h=2.000e-02$\\$(N=10000)$\end{tabular} & \begin{tabular}{l}$ -0.983 u_{xx} -1.087 u_{xxxx} -4.603 uu_x$\end{tabular} & \begin{tabular}{l}$ -1.077 u_{xx} -1.188 u_{xxxx} -5.016 uu_x$\end{tabular} & \begin{tabular}{l}$ -1.084 u_{xx} -1.195 u_{xxxx} -5.046 uu_x$\end{tabular} & \begin{tabular}{l}$ -1.080 u_{xx} -1.191 u_{xxxx} -5.028 uu_x$\end{tabular} \\ \hline
\begin{tabular}{c}$h=4.000e-02$\\$(N=5000)$\end{tabular} & \begin{tabular}{l}$ -0.861 u_{xx} -0.959 u_{xxxx} -4.131 uu_x$\end{tabular} & \begin{tabular}{l}$ -1.078 u_{xx} -1.192 u_{xxxx} -5.060 uu_x$\end{tabular} & \begin{tabular}{l}$ -0.990 u_{xx} -1.099 u_{xxxx} -4.724 uu_x$\end{tabular} & \begin{tabular}{l}$ -1.079 u_{xx} -1.193 u_{xxxx} -5.069 uu_x$\end{tabular} \\ \hline
\begin{tabular}{c}$h=6.000e-02$\\$(N=3334)$\end{tabular} & \begin{tabular}{l}$ -0.761 u_{xx} -0.855 u_{xxxx} -3.750 uu_x$\end{tabular} & \begin{tabular}{l}$ -1.079 u_{xx} -1.194 u_{xxxx} -5.096 uu_x$\end{tabular} & \begin{tabular}{l}$ -0.180 u_{xx} -0.241 u_{xxxx} -1.512 uu_x$\end{tabular} & \begin{tabular}{l}$ -1.079 u_{xx} -1.195 u_{xxxx} -5.103 uu_x$\end{tabular} \\ \hline
\begin{tabular}{c}$h=8.000e-02$\\$(N=2500)$\end{tabular} & \begin{tabular}{l}$ -0.681 u_{xx} -0.771 u_{xxxx} -3.440 uu_x$\end{tabular} & \begin{tabular}{l}$ -1.078 u_{xx} -1.195 u_{xxxx} -5.121 uu_x$\end{tabular} & \begin{tabular}{l}$ -0.500 uu_x$\end{tabular} & \begin{tabular}{l}$ -1.079 u_{xx} -1.195 u_{xxxx} -5.126 uu_x$\end{tabular} \\ \hline
\begin{tabular}{c}$h=1.000e-01$\\$(N=2000)$\end{tabular} & \begin{tabular}{l}$ -0.614 u_{xx} -0.702 u_{xxxx} -3.185 uu_x$\end{tabular} & \begin{tabular}{l}$ -1.077 u_{xx} -1.195 u_{xxxx} -5.139 uu_x$\end{tabular} & \begin{tabular}{l}$ -0.497 uu_x$\end{tabular} & \begin{tabular}{l}$ -1.077 u_{xx} -1.195 u_{xxxx} -5.142 uu_x$\end{tabular} \\ \hline
\begin{tabular}{c}$h=1.600e-01$\\$(N=1250)$\end{tabular} & \begin{tabular}{l}$ -0.468 u_{xx} -0.552 u_{xxxx} -2.631 uu_x$\end{tabular} & \begin{tabular}{l}$ -1.071 u_{xx} -1.192 u_{xxxx} -5.168 uu_x$\end{tabular} & \begin{tabular}{l}$ -0.489 uu_x$\end{tabular} & \begin{tabular}{l}$ -1.071 u_{xx} -1.192 u_{xxxx} -5.170 uu_x$\end{tabular} \\ \hline
\begin{tabular}{c}$h=2.000e-01$\\$(N=1000)$\end{tabular} & \begin{tabular}{l}$ -0.400 u_{xx} -0.482 u_{xxxx} -2.375 uu_x$\end{tabular} & \begin{tabular}{l}$ -1.066 u_{xx} -1.188 u_{xxxx} -5.176 uu_x$\end{tabular} & \begin{tabular}{l}$ -0.484 uu_x$\end{tabular} & \begin{tabular}{l}$ -1.067 u_{xx} -1.189 u_{xxxx} -5.180 uu_x$\end{tabular} \\ \hline
\end{tabular}
\end{scriptsize}
\end{adjustbox}
\vspace*{\fill}    
\caption{Robustness of Euler-SINDy and its unrolled version on Eq.~\ref{eq:1Dkuramoto}, with an increasing time step $h$ and a decreasing number of learning pairs.} \label{tab:kuramoto_table}
\end{table}

\begin{table}[t]
\centering
\includegraphics[width=\textwidth]{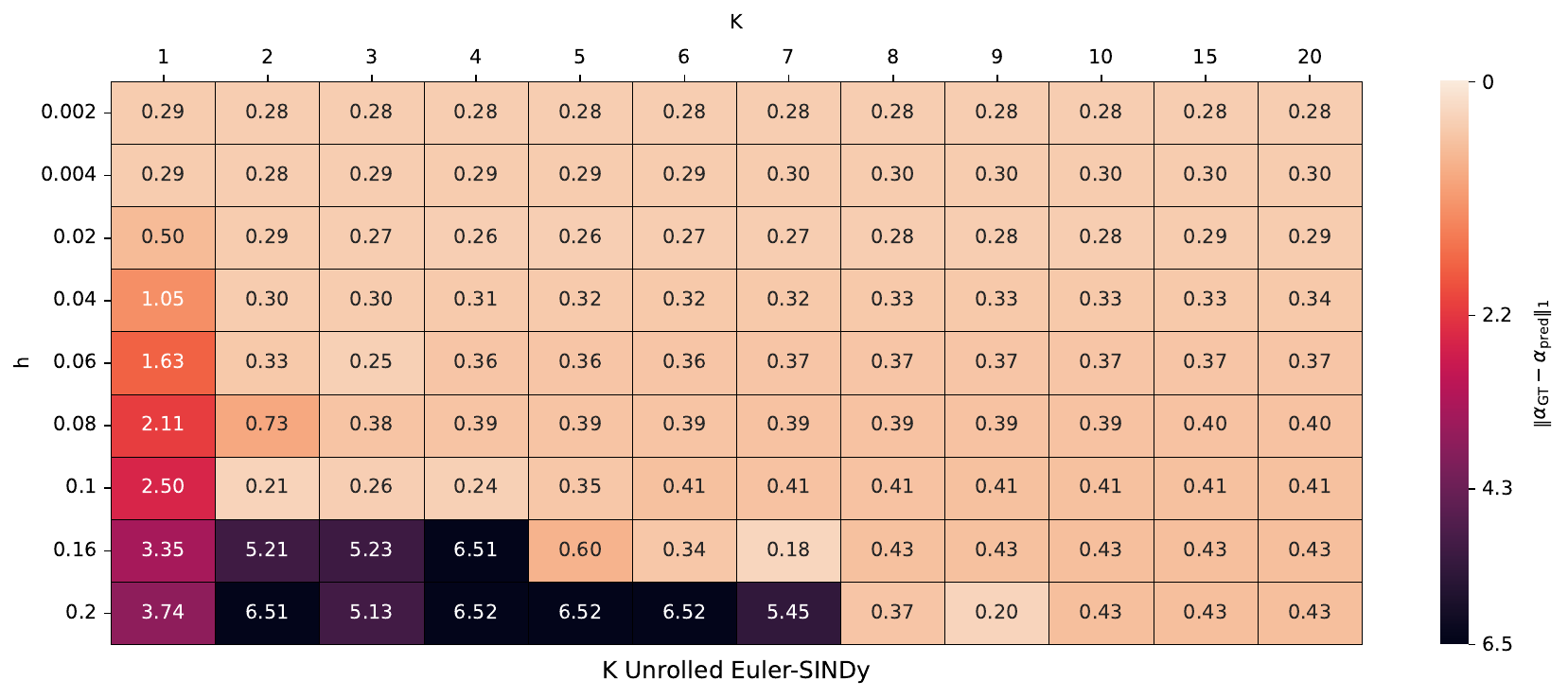}
\caption{Accuracy of $K$-Unrolled Euler-SINDy with different unrolling depths $K$ and observation step sizes $h$ for the KS equation (Eq.~\ref{eq:1Dkuramoto}).} \label{tab:kuramoto_sparse_euler_table_h_k}
\end{table}

\begin{table}[t]
\centering
\includegraphics[width=\textwidth]{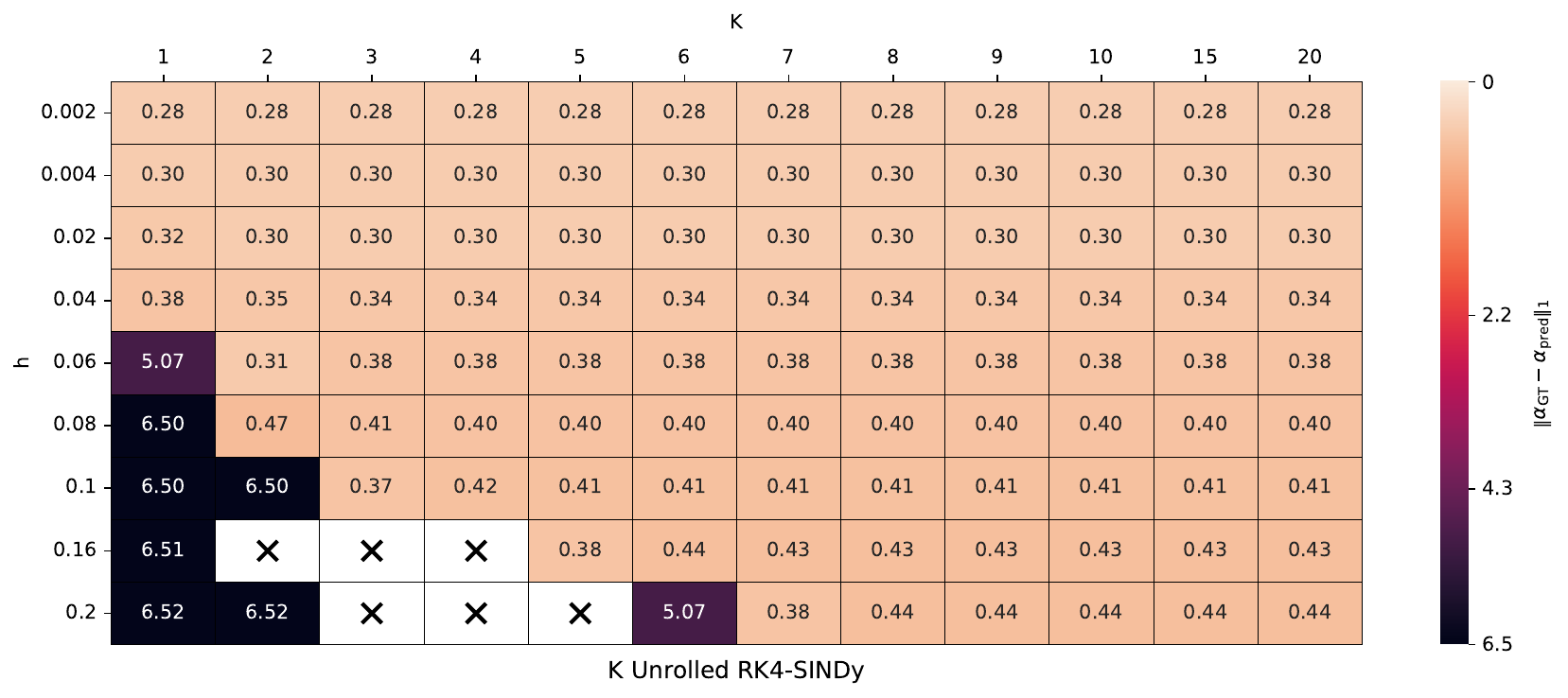}
\caption{Accuracy of $K$-Unrolled RK4-SINDy with different unrolling depths $K$ and observation step sizes $h$ for the KS equation (Eq.~\ref{eq:1Dkuramoto}).  The marker $X$ denotes entries where the results are NaN.} 
\label{tab:kuramoto_sparse_rk4_table_h_k}
\end{table}

\subsection{Numerical Stability versus Success in Identifying Governing Equations}

Both Euler-SINDy and RK4-SINDy are explicit methods that embed a  numerical scheme  during the iterations of the equation discovery. 
In Section~\ref{sec:theo}, we studied the truncation errors of these methods. However, it is worth noticing that these errors only concern the way the time derivative is approximated and do not depend on the complexity of the underlying physics. The stability analysis, based on Jacobian Eigenvalues, allows us to address this task by defining a region in the complex plane where the numerical solutions remain bounded. This region, called {\it absolute stability region}, can then be leveraged to establish a connection between the numerical scheme and its capacity for recovering the considered governing equations. We investigate how the numerical stability of the integration method (Euler or RK4) affects the ability to correctly recover the governing equation, using the cubic damped oscillator as a test case. The absolute stability region $SR$ of Euler and RK4 methods is the set of complex values $z$ defined as follows (see \cite[Sec. IV.2]{Hairer-book} for more details): 

\begin{definition}
    The absolute stability region of the Euler  (resp. RK4) method is defined as the set:
    $SR=\{z \in \mathbb{C} \ | \  |R(z)| \leq 1 \}$, 
    where $R(z)=1+z$ (resp. $R(z)=1+z+\frac{z^2}{2}+\frac{z^3}{6}+\frac{z^4}{24}$). 
\end{definition}

\begin{figure}[t]
\centering
\begin{subfigure}{0.4\textwidth}
    \includegraphics[width=\linewidth]{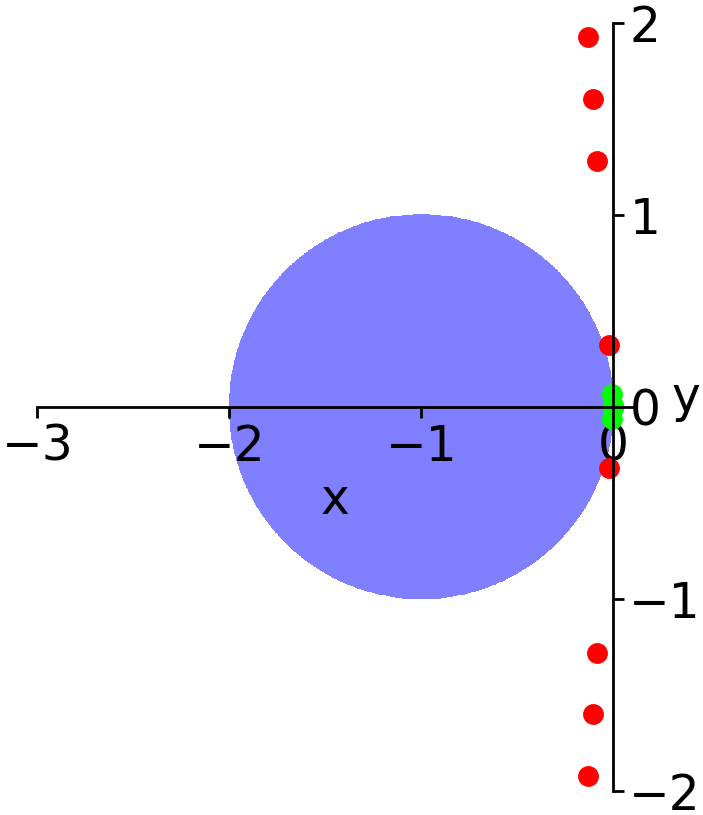}
    \caption{}
\end{subfigure}\quad \quad \quad \quad \quad \quad
\begin{subfigure}{0.25\textwidth}
    \includegraphics[width=\linewidth]{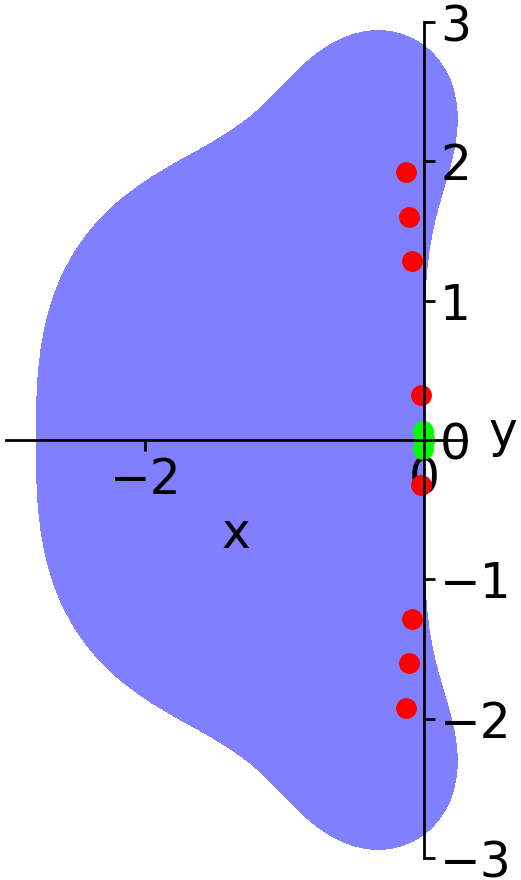}
    \caption{}
\end{subfigure}
\caption{Stability zones (in blue) for (a) Euler and (b) RK4 methods. The points $z_j = h \lambda_j$ are shown in green when the governing equation is successfully recovered, and in red when it is not.}

\label{fig:errors}
\end{figure}

These regions are  depicted in blue in Fig.~\ref{fig:errors}(a) (resp. b) and can  be used to determine if the methods are stable for recovering the cubic damped oscillator ODE, according to the following rule.

\begin{definition}[Stability of a numerical method for a function $f$]
Consider the problem $y'(x)=f(x,y)$ with $y(x_0)=y_0$ as initial condition. Let $\lambda_j$ denote the eigenvalues of the Jacobian matrix of $f$ at $(x_0,y_0)$.  
A numerical method with step size $h$ is stable for $f$ if $z_j=h \lambda_j  \in SR$ for all $j$.
\end{definition}


 From Eq.~\eqref{eq:oscillator-suppl}, we obtain the two eigenvalues $\lambda_1=-0.2159 + 3.206i=\overline{\lambda_2}$. 
As they are conjugate of each other, it follows that $|R(h\lambda_1)|=|R(h\lambda_2)|$ for any $h \in \mathbb{R}$. 
We report in the two stability regions of Fig.~\ref{fig:errors} the complex number $z_j=h \lambda_j$ for the different step sizes $h$ used in Table~\ref{tab:sparsity}.  We use green dots when the method succeeds in recovering the equations and red dots otherwise. 
We can note that all the green dots are inside the stability regions. This means that for the equation discovery method to perform well, a {\bf necessary condition} is that it is stable for the considered step size. This explains why Euler-SINDy fails dramatically at $h=0.6$ on Fig.~\ref{fig:errors}(a), its corresponding red dot being the farthest from the blue region. However, this {\bf condition is not sufficient} as it does not guarantee that the regression within SINDy will recover the governing equation. This is illustrated with the red dots inside the region for RK4-SINDy. 
Because our unrolled versions benefit of an implicit smaller step size $h/K$, the corresponding complex numbers  satisfy the constraint $|R(z)| \leq 1$ much more easily, justifying why they work much better.
The shapes of the stability regions hint that choosing between $4K$-Unrolled Euleur-SINDy and $K$-Unrolled RK4-SINDy (which are using the same number of dictionary evaluations) might depend on the equation (none of the 4-times-bigger circle of Euler and the RK4 region is included in the other) that is to be recovered.

\subsection{Dictionary terms used in the experiments}

We report in Table~\ref{tab:pde_summary} a summary of information on the equations used in the experiments.

\begin{table}[p]
\vspace*{\fill}    
\centering         
\begin{adjustbox}{angle=90, scale=0.8, center}
\begin{tabular}{|c|c|c|}
\hline
\thead{Equation Name} & \thead{Equation Expression} & \thead{Candidate Terms Dictionary} \\
\hline

2D Cubic Damped Oscillator & 
\(
\left\{
\begin{aligned}
u_t & = -0.1u^3+2.0v^3 \\
v_t & = -2.0u^3-0.1v^3
\end{aligned}
\right.
\) &
\begin{tabular}{l}
$\{1,\, u,\, v,\, u^2,\, u v,\, v^2,\, u^3,\, u^2 v,$ \\
$u v^2,\, v^3,\, u^4,\, u^3 v,\, u^2 v^2,\, u v^3,\, v^4\}$
\end{tabular}
\\ \hline

Advection Equation & 
$\displaystyle
u_t = - 0.4 u_x 
$ &
$\{1, u, u^2, u^3, u_x, u_{xx}, u_{xxx}\}$
\\ \hline

2D Reaction-Diffusion Equation & 
\(
\left\{
\begin{aligned}
u_t &= u - u^3 + v^3 + 0.1 u_{xx} + 0.1 u_{yy} + u^2 v - u v^2 \\
v_t &= v - u^3 - v^3 + 0.1 v_{xx} + 0.1 v_{yy} - u^2 v - u v^2
\end{aligned}
\right.
\)
&
\begin{tabular}{l}
$\{1,\, u,\, v,\, u^2,\, v^2,\, u^3,\, v^3, u_x,\, v_x,\, u_y,\, v_y,$ \\
$u_{xx},\, v_{xx},\, u_{yy},\, v_{yy},\, u_{xy},\, v_{xy},\, u v,\, u^2 v,\, u v^2\}$
\end{tabular}
\\[10pt] \hline

Kuramoto-Sivashinsky Equation &
$\displaystyle
u_t = -u u_x - u_{xx} - 5u_{xxxx}
$ &
$\{1, u_x, u_{xx}, u_{xxx}, u_{xxxx}, u u_x\}$
\\ \hline
\end{tabular}
\end{adjustbox}
\vspace*{\fill}    
\caption{Summary of information on the equations used in the experiments. Each row lists the name of the PDE/ODE, its analytical expression, and the dictionary of candidate terms used during the discovery process.} \label{tab:pde_summary}
\end{table}

\section{Supplementary experiments with iNeural-SINDy}\label{suppl:ineural}

\subsection{Linear oscillator equation}

We focus here on the linear oscillator. The governing equations are given by:
\begin{eqnarray}
\left\{\begin{array}{rcl}
\dot{x}(t) & = & -0.1x(t)+2.0y(t) \\
\dot{y}(t) & = & -2.0x(t)-0.1y(t)
\end{array}\right. \label{eq:linear_oscillator}
\end{eqnarray}
where, $\dot{x}(t)$ and $\dot{y}(t)$ denote the time derivatives of the state variables $x(t)$ and $y(t)$, respectively.

\paragraph{Experimental setup:}
We adopt the same setup as in the original iNeural-SINDy paper. Noisy observations are generated as follows: $\tilde{x}(t) = x(t) + \eta_x(t),$ and $\tilde{y}(t) = y(t) + \eta_y(t),$
where $\eta_x(t)$ and $\eta_y(t)$ are independent Gaussian noise with standard deviations $\sigma_x = \sigma \, \mathrm{std}(x(t))$ and $\sigma_y = \sigma \, \mathrm{std}(y(t))$ respectively, with $\sigma \in [0,0.06]$ controlling the relative noise amplitude. The experiments are conducted over a range of observation intervals $h \in [0.025, 0.333]$.

\paragraph{Results}

The comparative results are reported in Tab.~\ref{tab:linear_oscillator_l1_loss_ineural}. iNeural-SINDy is relatively robust to noise. For both Euler (Tab.~\ref{tab:linear_oscillator_l1_loss_ineural}.a) and RK4 (Tab.~\ref{tab:linear_oscillator_l1_loss_ineural}.c), accuracy decreases as the temporal gap increases. This limitation is substantially mitigated by the unrolled variants (Tab.~\ref{tab:linear_oscillator_l1_loss_ineural}.b,d), with 8-step Euler-iNeural-SINDy and 2-step RK4-iNeural-SINDy better capturing the underlying dynamics and consistently reducing the $\ell_1$ error between predicted and theoretical coefficients.

\begin{table*}[t]
\centering

\begin{subtable}{0.9\textwidth}
    \includegraphics[width=\linewidth]{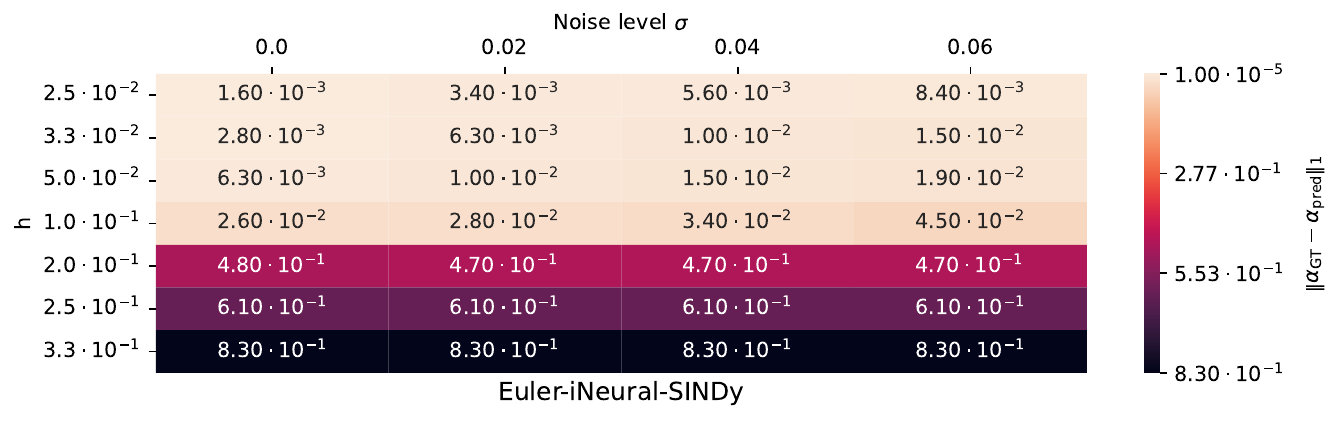}
    \caption{}
\end{subtable}

\begin{subtable}{0.9\textwidth}
    \includegraphics[width=\linewidth]{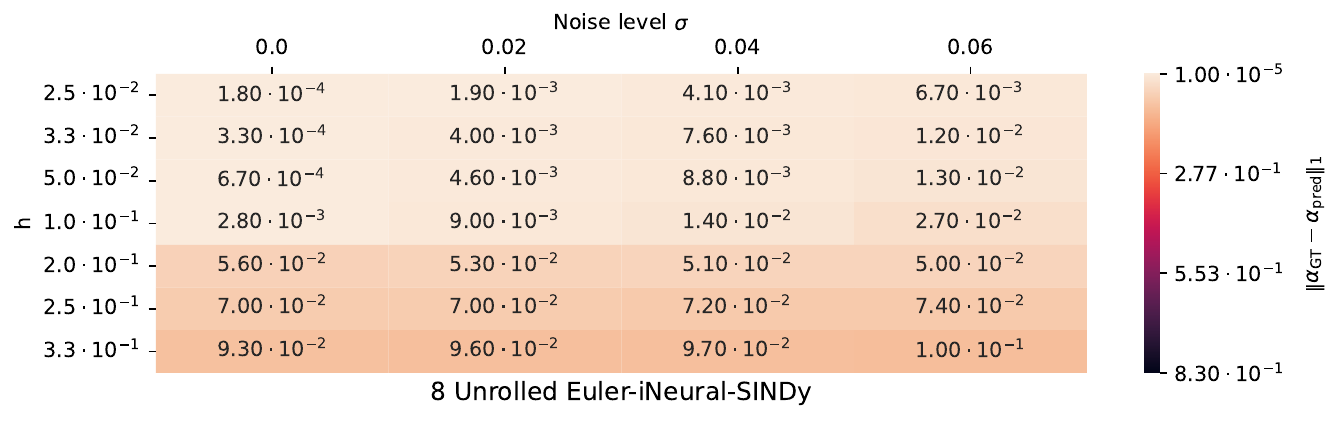}
    \caption{}
\end{subtable}

\begin{subtable}{0.9\textwidth}
    \includegraphics[width=\linewidth]{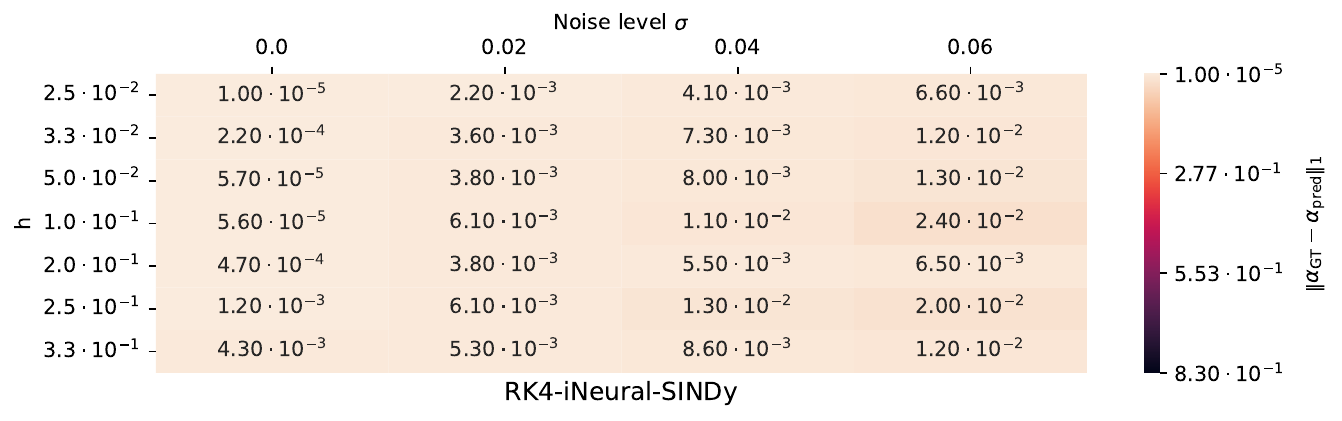}
    \caption{}
\end{subtable}
\hfill
\begin{subtable}{0.9\textwidth}
    \includegraphics[width=\linewidth]{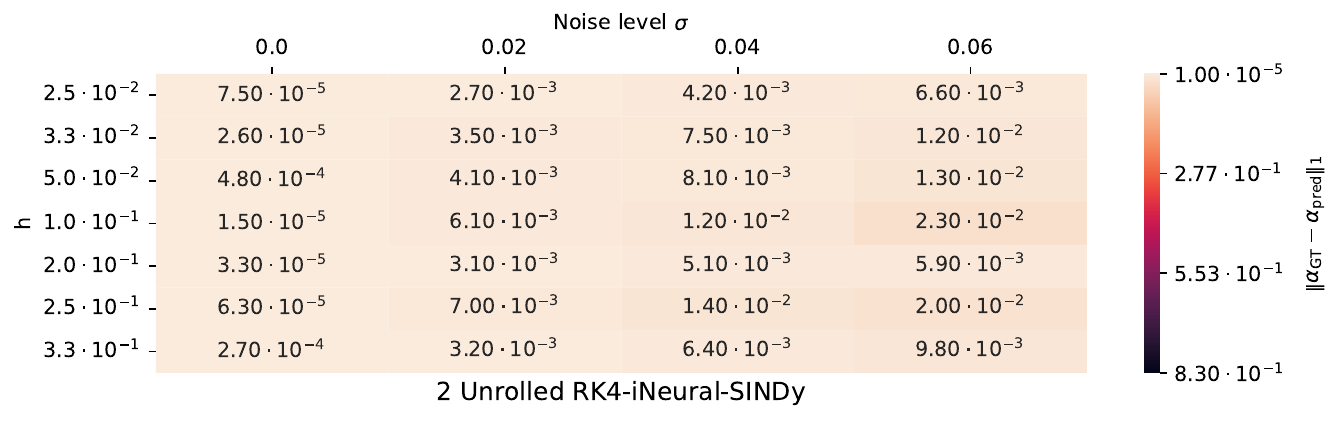}
    \caption{}
\end{subtable}

\caption{Robustness of iNeural-SINDy on the linear oscillator (Eq.~\ref{eq:linear_oscillator}), evaluated with increasing time step $h$ and  noise level $\sigma$ with a) Euler-iNeural-SINDy, b) 8 Unrolled Euler-iNeural-SINDy, c) RK4-iNeural-SINDy, and d) 2 Unrolled RK4-iNeural-SINDy.}
\label{tab:linear_oscillator_l1_loss_ineural}
\end{table*}

\subsection{Fitz Hugh Nagumo equation}

The system under consideration is the FitzHugh-Nagumo model, a two-dimensional nonlinear dynamical system originally introduced as a simplified version of the Hodgkin–Huxley equations for modeling the activation and deactivation dynamics of a spiking neuron. The governing equations are given by:
\begin{equation}
\left\{
\begin{aligned}
\dot{x}(t) &= 1.0\,x(t) - 1.0\,y(t) - \tfrac{1}{3}\,x^{3}(t) + 0.1, \\
\dot{y}(t) &= 0.1\,x(t) - 0.1\,y(t),
\end{aligned}
\right.
\label{eq:nagumo}
\end{equation}
where $\dot{x}(t)$ and $\dot{y}(t)$ denote the time derivatives of the state variables $x(t)$ and $y(t)$, respectively.

\paragraph{Experimental setup:}
We adopt the same setup as in the original iNeural-SINDy paper. Noisy observations are generated as follows: $\tilde{x}(t) = x(t) + \eta_x(t),$ and $\tilde{y}(t) = y(t) + \eta_y(t),$
where $\eta_x(t)$ and $\eta_y(t)$ are independent Gaussian noise with standard deviations $\sigma_x = \sigma \, \mathrm{std}(x(t))$ and $\sigma_y = \sigma \, \mathrm{std}(y(t))$ respectively, with $\sigma \in [0,0.06]$ controlling the relative noise amplitude. The experiments are conducted over a range of observation intervals $h \in [0.44, 1.3]$.

\paragraph{Results}
The results of iNeural-SINDy applied to FitzHugh-Nagumo equation are reported in Tab.~\ref{tab:nagumo_l1_loss_ineural}. The overall behavior is similar the linear oscillator equation presented previously.

\begin{table*}[t]
\centering

\begin{subtable}{0.9\textwidth}
    \includegraphics[width=\linewidth]{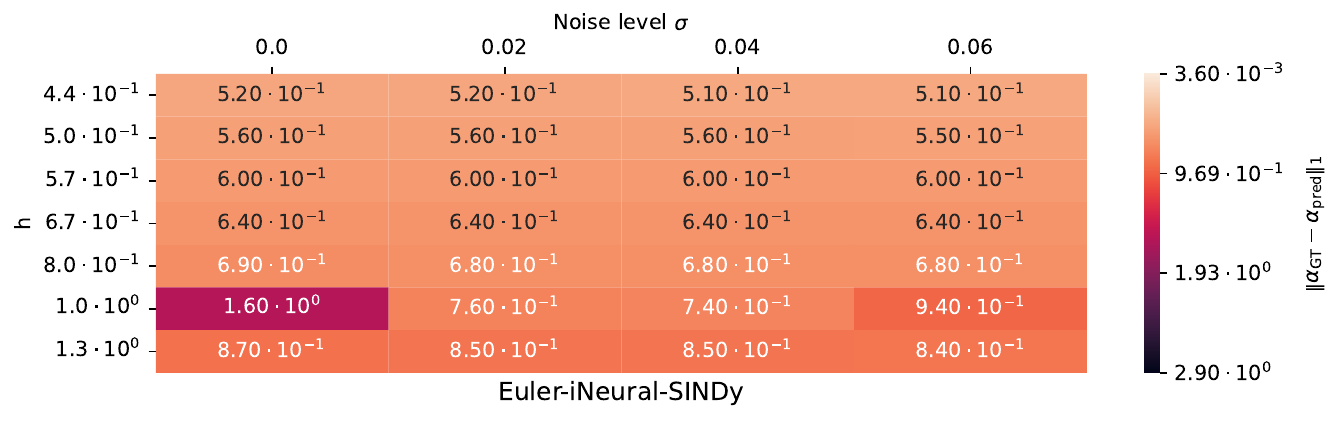}
    \caption{}
\end{subtable}

\begin{subtable}{0.9\textwidth}
    \includegraphics[width=\linewidth]{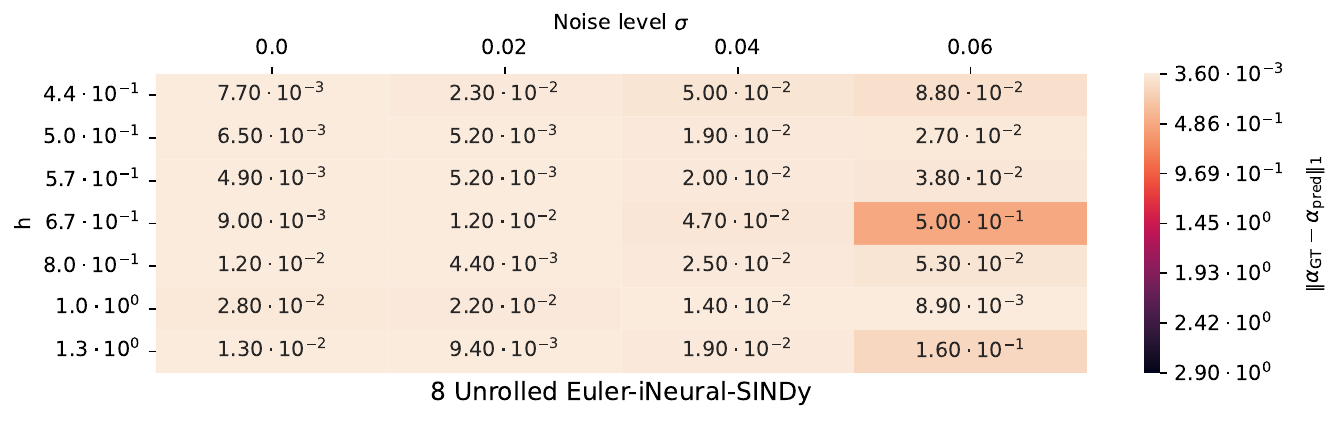}
    \caption{}
\end{subtable}

\begin{subtable}{0.9\textwidth}
    \includegraphics[width=\linewidth]{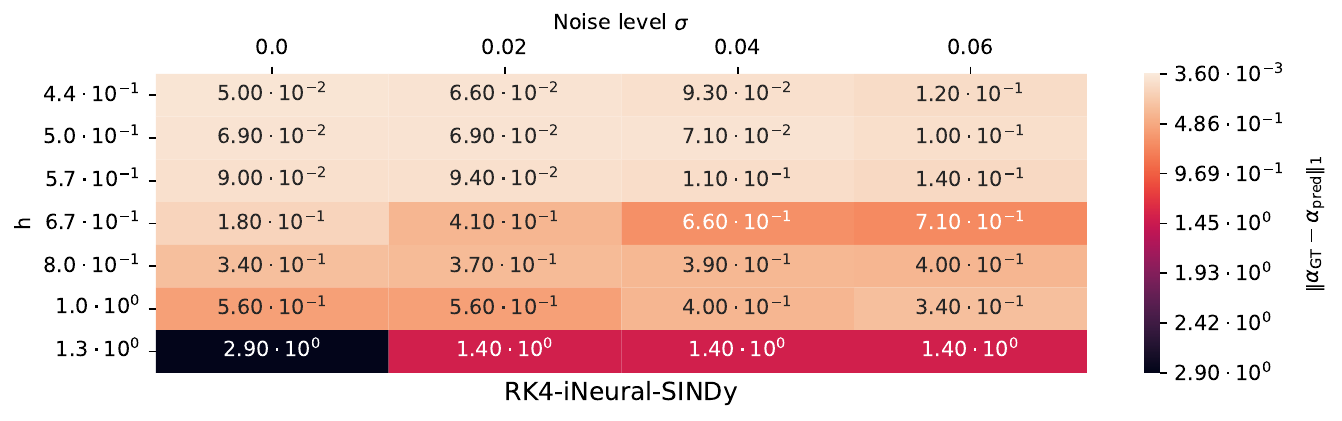}
    \caption{}
\end{subtable}

\begin{subtable}{0.9\textwidth}
    \includegraphics[width=\linewidth]{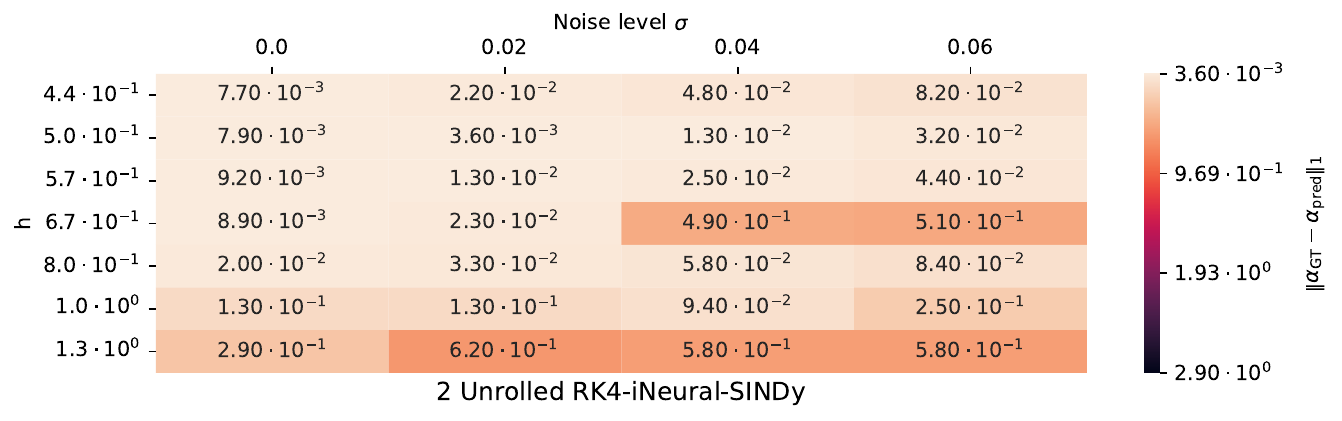}
    \caption{}
\end{subtable}

\caption{Robustness of iNeural-SINDy on the FitzHugh-Nagumo (Eq.~\ref{eq:nagumo}), evaluated with increasing time step $h$ and  noise level $\sigma$ with a) Euler-iNeural-SINDy, b) 8 Unrolled Euler-iNeural-SINDy, c) RK4-iNeural-SINDy, and d) 2 Unrolled RK4-iNeural-SINDy.}
\label{tab:nagumo_l1_loss_ineural}
\end{table*}

\clearpage


\begin{thebibliography}{10}
\providecommand{\url}[1]{\texttt{#1}}
\providecommand{\urlprefix}{URL }
\providecommand{\doi}[1]{https://doi.org/#1}
\bibitem[Boninsegna et~al.(2018)Boninsegna, N\"uske, and Clementi]{Boninsegna_2018}
L.~Boninsegna, F.~N\"uske, and C.~Clementi.
\newblock Sparse learning of stochastic dynamical equations.
\newblock \emph{The Journal of Chemical Physics}, 148\penalty0 (24), Mar. 2018.
\newblock ISSN 1089-7690.


\bibitem[Both et~al.(2021)Both, Choudhury, Sens, and Kusters]{Both_2021}
G.J. Both, S.~Choudhury, P.~Sens, and R.~Kusters.
\newblock Deepmod: Deep learning for model discovery in noisy data.
\newblock \emph{Journal of Computational Physics}, 428:\penalty0 109985, Mar. 2021.
\newblock ISSN 0021-9991.


\bibitem[{Brunton} et~al.(2016){Brunton}, {Proctor}, and {Kutz}]{Brunton2016}
S.L. {Brunton}, J.L. {Proctor}, and J.N. {Kutz}.
\newblock {Discovering governing equations from data by sparse identification of nonlinear dynamical systems}.
\newblock \emph{Proceedings of the National Academy of Science}, 113\penalty0 (15):\penalty0 3932--3937, Apr. 2016.


\bibitem[Brunton et~al.(2016)Brunton, Proctor, and Kutz]{brunton2016SINDYc}
S.L. Brunton, J.L. Proctor, and J.N. Kutz.
\newblock Sparse identification of nonlinear dynamics with control (sindyc), 2016.


\bibitem[Chen et~al.(2024)Chen, Li, and Jin]{ICNET}
C.~Chen, H.~Li, and X.~Jin.
\newblock An invariance constrained deep learning network for pde discovery, 2024.


\bibitem[Doum{\`e}che et~al.(2025)Doum{\`e}che, Biau, and Boyer]{Doumeche.BERN.2025}
N.~Doum{\`e}che, G.~Biau, and C.~Boyer.
\newblock {On the convergence of PINNs}.
\newblock \emph{Bernoulli}, 31\penalty0 (3):\penalty0 2127 -- 2151, 2025.


\bibitem[Fasel et~al.(2022)Fasel, Kutz, Brunton, and Brunton]{urban2022a}
U.~Fasel, J.N. Kutz, B.W. Brunton, and S.L. Brunton.
\newblock Ensemble-sindy: Robust sparse model discovery in the low-data, high-noise limit, with active learning and control.
\newblock \emph{Proceedings of The Royal Society A: Mathematical, Physical and Engineering Sciences}, 2022.


\bibitem[Forootani et~al.(2025)Forootani, Goyal, and Benner]{ForootaniGB25}
A.~Forootani, P.~Goyal, and P.~Benner.
\newblock A robust sparse identification of nonlinear dynamics approach by combining neural networks and an integral form.
\newblock \emph{Eng. Appl. Artif. Intell.}, 149:\penalty0 110360, 2025.

\bibitem[Goyal and Benner(2022)]{RK4-SINDy}
P.~Goyal and P.~Benner.
\newblock Discovery of nonlinear dynamical systems using a runge-kutta inspired dictionary-based sparse regression approach.
\newblock \emph{Proceedings. Mathematical, physical, and engineering sciences}, 478:\penalty0 20210883, 06 2022.


\bibitem[Hairer and Wanner(1996)]{Hairer-book}
E.~Hairer and G.~Wanner.
\newblock \emph{Solving Ordinary Differential Equations II. Stiff and Differential-Algebraic Problems}, volume~14.
\newblock 01 1996.


\bibitem[Hairer et~al.(2010)Hairer, N{\o}rsett, and Wanner]{Hairer}
E.~Hairer, S.P. N{\o}rsett, and G.~Wanner.
\newblock \emph{Solving ordinary differential equations. {I}: {Nonstiff} problems.}, volume~8 of \emph{Springer Ser. Comput. Math.}
\newblock Berlin: Springer, 2nd revised ed., 3rd corrected printing edition, 2010.
\newblock ISBN 978-3-642-05163-0.

\bibitem[Hoffmann et~al.(2018)Hoffmann, Fr\"ohner, and No\'e]{reactiveSINDy}
M.~Hoffmann, C.~Fr\"ohner, and F.~No\'e.
\newblock Reactive sindy: Discovering governing reactions from concentration data, 10 2018.

\bibitem[Kaheman et~al.(2020)Kaheman, Kutz, and Brunton]{Kaheman_2020}
K.~Kaheman, J.N. Kutz, and S.L. Brunton.
\newblock Sindy-pi: a robust algorithm for parallel implicit sparse identification of nonlinear dynamics.
\newblock \emph{Proceedings of the Royal Society A: Mathematical, Physical and Engineering Sciences}, 476\penalty0 (2242), Oct. 2020.
\newblock ISSN 1471-2946.


\bibitem[Karniadakis et~al.(2021)Karniadakis, Kevrekidis, Lu, Perdikaris, Wang, and Yang]{osti_1852843}
G.E. Karniadakis, I.G. Kevrekidis, L.~Lu, P.~Perdikaris, S.~Wang, and L.~Yang.
\newblock Physics-informed machine learning.
\newblock \emph{Nature Reviews Physics}, 3\penalty0 (6), 5 2021.
\newblock ISSN 2522-5820.

\bibitem[Li et~al.(2021)Li, Kovachki, Azizzadenesheli, Liu, Bhattacharya, Stuart, and Anandkumar]{LiKALBSA21}
Z.~Li, N.B. Kovachki, K.~Azizzadenesheli, B.~Liu, K.~Bhattacharya, A.M. Stuart, and A.~Anandkumar.
\newblock Fourier neural operator for parametric partial differential equations.
\newblock In \emph{{ICLR} 2021, Austria, May 3-7, 2021}. OpenReview.net, 2021.

\bibitem[Li et~al.(2023)Li, Zheng, Kovachki, Jin, Chen, Liu, Azizzadenesheli, and Anandkumar]{li2023physicsinformed}
Z.~Li, H.~Zheng, N.~Kovachki, D.~Jin, H.~Chen, B.~Liu, K.~Azizzadenesheli, and A.~Anandkumar.
\newblock Physics-informed neural operator for learning partial differential equations, 2023.

\bibitem[Long et~al.(2018)Long, Lu, Ma, and Dong]{long2018pdenetlearningpdesdata}
Z.~Long, Y.~Lu, X.~Ma, and B.~Dong.
\newblock Pde-net: Learning pdes from data, 2018.
\newblock URL \url{https://arxiv.org/abs/1710.09668}.

\bibitem[Messenger and Bortz(2021)]{Messenger_2021}
D.A. Messenger and D.M. Bortz.
\newblock Weak sindy for partial differential equations.
\newblock \emph{Journal of Computational Physics}, 443:\penalty0 110525, Oct. 2021.
\newblock ISSN 0021-9991.


\bibitem[Raissi et~al.(2019)Raissi, Perdikaris, and Karniadakis]{raissi2019physics}
M.~Raissi, P.~Perdikaris, and G.E. Karniadakis.
\newblock Physics-informed neural networks: A deep learning framework for solving forward and inverse problems involving nonlinear partial differential equations.
\newblock \emph{Journal of Computational physics}, 2019.

\bibitem[Rudy et~al.(2016)Rudy, Brunton, Proctor, and Kutz]{PDE-FIND}
S.H. Rudy, S.L. Brunton, J.L. Proctor, and J.N. Kutz.
\newblock Data-driven discovery of partial differential equations, 2016.
\newblock URL \url{https://arxiv.org/abs/1609.06401}.

\bibitem[Stephany and Earls(2022)]{PDE-READ}
R.~Stephany and C.~Earls.
\newblock Pde-read: Human-readable partial differential equation discovery using deep learning, 2022.
\newblock URL \url{https://arxiv.org/abs/2111.00998}.

\bibitem[Stephany and Earls(2024)]{STEPHANY2024106242}
R.~Stephany and C.~Earls.
\newblock Pde-learn: Using deep learning to discover partial differential equations from noisy, limited data.
\newblock \emph{Neural Networks}, 174:\penalty0 106242, 2024.
\newblock ISSN 0893-6080.


\bibitem[Yin et~al.(2021)Yin, Le~Guen, Dona, de~B√©zenac, Ayed, Thome, and Gallinari]{Yin_2021}
Y.~Yin, V.~Le~Guen, J.~Dona, E.~de~B\'ezenac, I.~Ayed, N.~Thome, and P.~Gallinari.
\newblock Augmenting physical models with deep networks for complex dynamics forecasting*.
\newblock \emph{Journal of Statistical Mechanics: Theory and Experiment}, 2021\penalty0 (12):\penalty0 124012, Dec. 2021.
\newblock ISSN 1742-5468.








\end{thebibliography}
\end{document}